\documentclass{article}

\usepackage[final]{neurips_2025}

\usepackage{answers}
\usepackage{setspace}
\usepackage{graphicx}
\usepackage{graphics}
\usepackage{subcaption}
\usepackage{enumitem}
\usepackage{multicol}
\usepackage{mathrsfs}
\usepackage{bbm}
\usepackage{amsmath,amsthm,amssymb}
\usepackage{rotating}
\usepackage{float}
\usepackage{url}
\usepackage{mathtools}
\usepackage{caption}
\usepackage{algorithm}
\usepackage{algorithmic}
\usepackage[toc,page]{appendix}
\usepackage{color}
\usepackage{tikz}
\usetikzlibrary{bayesnet}
\usepackage{tabularx}
\usepackage{array}
\usepackage{multirow}
\usepackage{balance}
\usepackage{xcolor}

\DeclarePairedDelimiterX{\infdivx}[2]{\Big[}{\Big]}{%
  #1\;\delimsize\|\;#2%
}

\newcommand{\vecx}{\mathbf{x}} 
\newcommand{\vecy}{\mathbf{y}}

\newcommand{\vecu}{\mathbf{u}}
\newcommand{\vecz}{\mathbf{z}}

\newcommand{\N}{\mathbbm{N}}

\newcommand{\R}{\mathbbm{R}}

\newtheorem{Theorem}{Theorem}
\newtheorem{Definition}{Definition}
\newtheorem{Lemma}{Lemma}

\newtheorem{Proposition}{Proposition}

\DeclareMathOperator*{\argmax}{argmax}

\usepackage[utf8]{inputenc} 
\usepackage[T1]{fontenc}    
\usepackage[hidelinks]{hyperref}       
\usepackage{url}            
\usepackage{booktabs}       
\usepackage{amsfonts}       
\usepackage{nicefrac}       
\usepackage{microtype}      
\usepackage{xcolor}         

\title{Neural Mutual Information Estimation \\ with Vector Copulas}

\author{
    Yanzhi Chen$^{1}$, Zijing Ou$^2$, Adrian Weller$^{1,3}$, Michael U. Gutmann$^{4}$\\
    \hspace{-0.535cm} $^1$University of Cambridge, $^2$Imperial College London, $^3$Alan Turing Institute, $^4$University of Edinburgh
}

\begin{document}

\maketitle

\begin{abstract}

Estimating mutual information (MI) is a fundamental task in data science and machine learning. Existing  estimators mainly rely on either highly flexible models (e.g., neural networks), which require large amounts of data, or overly simplified models (e.g., Gaussian copula), which fail to capture complex distributions. Drawing upon recent vector copula theory, we propose a principled interpolation between these two extremes to achieve a better trade-off between complexity and capacity. Experiments on state-of-the-art synthetic benchmarks and real-world data with diverse modalities demonstrate the advantages of the proposed estimator.
\end{abstract}

\section{Introduction}
Mutual information (MI) is a fundamental measure of the statistical dependence between random variables (RVs). Compared to other dependence measures, MI stands out due to its equitability and generality~\citep{kinney2014equitability, Cover2003Elements}: it can capture non-linear dependence of any form and can handle RVs with any dimensionalities, rendering it a powerful measure for quantifying statistical dependence. In data science, MI is widely used to analyze the relationships between protein sequences~\cite{gowri2024approximating} and gene profiles~\cite{reverter2008combining, song2012comparison}, as well as to assess feature importance and redundancy~\cite{peng2005feature}. In machine learning, MI broadly serves as a learning objective and  regularizer~\cite{hjelm2018learning, tschannen2019mutual, wu2020mutual, chen2016infogan, chen2023learning, alemi2016deep}, with diverse applications to representation learning~\cite{hjelm2018learning, tschannen2019mutual, wu2020mutual, chen2020neural,qiu2021unsupervised}, generative modeling~\cite{chen2016infogan}, fairness and privacy~\cite{chen2022scalable, noorbakhsh2024inf2guard}, etc.

A wide range of powerful, neural MI estimators have been developed~\cite{duong2023diffeomorphic, cheng2020club, franzeseminde2023, belghazi2018mutual, van2018representation, guo2022tight, letizia2024mutual}. Most of these estimators rely on a \emph{single, unconstrained} network to approximate certain quantities—such as the joint density $p(\vecx, \vecy)$ or the density ratios $p(\vecx, \vecy)/p(\vecx)p(\vecy)$—during MI estimation. While neural networks as universal functional approximators can, in theory,  approximate arbitrary functions given sufficient data~\cite{hornik1989multilayer, yarotsky2017error}, in practice we often only have a small set of data. Indeed, theoretical studies have shown that such \emph{distribution-free} treatment of MI estimation will inevitably suffer from requiring an exponential sample size~\cite{paninski2003estimation, wu2016minimax, barron2022convergence, mcallester2020formal}. A straightforward remedy is to restrict the model to simpler classes—for instance, assuming that the data  is approximately Gaussian. However, these assumptions are often overly simplistic to capture complex distributions in reality.

Recent advances in vector copula theory~\cite{fan2023vector} offer a promising avenue for addressing this dilemma. Vector copula theory extends classical copula theory~\cite{sklar1959fonctions} by generalizing it from \emph{univariate} to \emph{vector} marginals. It reveals that the multivariate marginals and the dependence structure (i.e., the vector copula) of a joint distribution are fully disentangled. This disentanglement motivates a more fine-grained way for making assumption in MI estimation, where we impose lightweight yet reasonable assumptions solely on the \emph{vector copula}  rather than on the \emph{entire distribution}. Crucially, the complexity of the vector copula can be adaptively adjusted through efficient vector copula selection, allowing for an optimal trade-off between  capacity and complexity. Experiments on state-of-the-art synthetic benchmarks and real-world data demonstrate the competitiveness of our estimator against state-of-the-art estimators. In summary, the main contributions of this work are three-fold:


\begin{itemize}[leftmargin=*]
    \item We develop a divide-and-conquer MI estimator based on recent vector copula theory, which explicitly disentangles marginal distributions and dependence structure in MI estimation; 
    \item We reinterpret existing estimators through the lens of vector copula, revealing that they correspond to varying parameterization and learning strategies of vector copula with various trade-offs;
    \item We provide consistency and error analysis of our estimator, along with extensive empirical evaluation on diverse test cases covering multiple modalities, marginal patterns and dependence structures. 
\end{itemize}

\begin{figure}[t]
\begin{subfigure}{1.0\textwidth}
\centering
\includegraphics[width=0.99\linewidth]{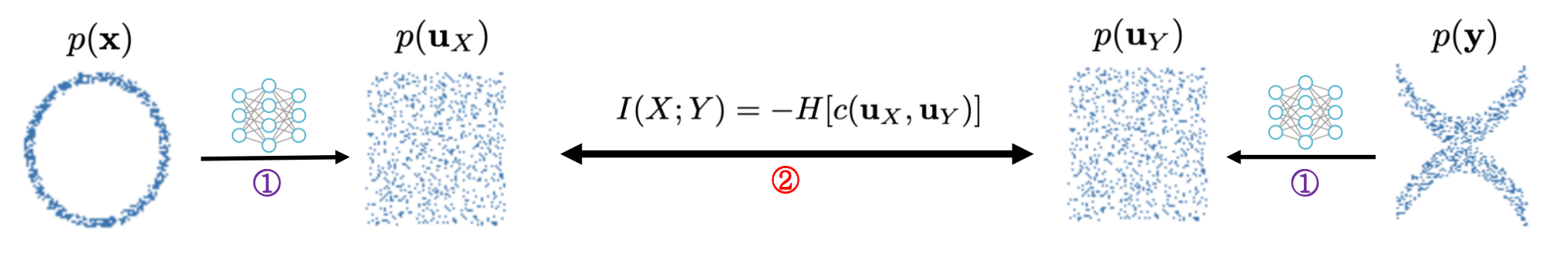}
\end{subfigure}
\vspace{0.1cm}
\caption[Overview of the proposed vector copula-based MI estimator]{\textbf{Overview of the proposed vector copula-based MI estimator} (VCE), which explicitly disentangles the modeling of marginal distribution and dependence structure (i.e., the vector copula). VCE first respectively computes the vector ranks $\vecu_X$ and $\vecu_Y$ corresponding to the two marginal variables $X$ and $Y$ with flow models ({\color{violet} \raisebox{.4pt}{\textcircled{\raisebox{-.9pt} {1}}}}). It then finds the vector copula $c \in \mathcal{C}$ from the vector copula pool $\mathcal{C}$ that best matches with the joint distribution $p(\vecu_X, \vecu_Y)$ of the estimated vector ranks ({\color{red} \raisebox{.4pt} {\textcircled{\raisebox{-.9pt} {2}}}}). Mutual information $I(X; Y)$ is computed as the negative differential entropy of the vector copula $c$, which itself is irrelevant to the two marginal distributions $p(\vecx)$ and $p(\vecy)$. }
\label{figure:vc}
\end{figure}

Code containing both our method and state-of-the-art neural estimators is available in  \href{https://github.com/cyz-ai/neural-MI-estimate-clean}{[{github repo}]}. 

\section{Preliminaries}
Throughout this work, we use upper case letters (e.g. $X$) to denote random variables and lower case letters (e.g. $\vecx$) to denote their instances.   We use $\mathcal{U}[0, 1]^d$ or $\mu$ to denote the uniform distribution on $[0, 1]^d$ and use $\mathcal{N}$ to denote Gaussian distribution on $\R^d$. $\nabla$ denotes the gradient and $J_{\vecx}\vecy$ denotes the Jacobian of $\vecy$ w.r.t $\vecx$. The symbol $\#$ denotes the push-forward operation.



\subsection{Mutual information and its estimation}
\label{sec:bg}

The mutual information (MI) between variables $X$ and $Y$ is defined as the Kullback-Leibler (KL) divergence between the joint distribution $p(\vecx, \vecy)$ and the product of marginal distributions $p(\vecx)p(\vecy)$:
\begin{equation}
      I(X; Y) = KL[p(\vecx, \vecy)\|p(\vecx)p(\vecy)] = \mathbb{E}\left[ \log \frac{p(\vecx,\vecy)}{p(\vecx)p(\vecy)} \right]
      \label{eq:mi}
\end{equation}

In this work, we consider estimating $I(X; Y)$ from an empirical dataset $\mathcal{D} = \{\vecx^{(i)}, \vecy^{(i)} \}^n_{i=1}$. Several neural network-based methods have been developed for MI estimation:


\textbf{Generative estimators}. These methods leverage generative models to approximate the various distributions in \eqref{eq:mi} or their equivalents, and use the learned generative models to construct an MI estimate~\citep{duong2023diffeomorphic, mcallester2020formal, cheng2020club, song2019understanding, franzeseminde2023}. The accuracy of generative estimators crucially depends on the quality of the learned generative models. Simpler models (e.g. Gaussian copula) are easy to learn but may fail to adequately capture the true data distribution~\cite{singh2017nonparanormal, chen2019adaptive}. In contrast, complex models (e.g. flow-based models~\cite{papamakarios2017masked, durkan2019neural, papamakarios2021normalizing, lipman2022flow} and diffusion models~\cite{franzeseminde2023}) offer greater expressiveness but can be challenging to optimize, in particular if the amount of data is  insufficient or the data dimensionality is high.

\textbf{Discriminative estimators}. These methods train a neural network $f$ with samples $\vecx, \vecy \sim p(\vecx, \vecy)$ and samples $\vecx, \vecy \sim p(\vecx)p(\vecy)$ to estimate the density ratio $p(\vecx, \vecy)/p(\vecx)p(\vecy)$~\cite{belghazi2018mutual, van2018representation, liao2020demi, rhodes2020telescoping, srivastava2023estimating, song2019understanding, letizia2024mutual}. Once trained, the learned density ratio  can either be used in \eqref{eq:mi} or in the Donsker-Varadhan (DV) representation~\citep{donsker1983asymptotic} to obtain an MI estimate. Discriminative methods avoid directly modeling  densities, however they are prone to the curse of \emph{high-discrepancy}~\cite{rhodes2020telescoping, srivastava2023estimating, mcallester2020formal, song2019understanding}, which occurs if $p(\vecx, \vecy)$ and $p(\vecx)p(\vecy)$ differ significantly --- for instances, cases with high MI or high-dimensional data. Several advanced methods were proposed to alleviate this issue, including clipping the network outputs~\cite{song2019understanding}, introducing reference distributions~\cite{srivastava2023estimating}, avoiding computing the partition function~\cite{letizia2024mutual}.

\subsection{Vector copula}

\textbf{Vector copula theory}. The recent vector copula theory~\cite{fan2023vector} provides a principled framework for modeling and analyzing the dependence between \emph{multivariate} random variables. It extends classical copula theory \cite{sklar1959fonctions} by considering `vector' marginals. We begin by the concept of vector ranks:
\begin{Definition}[Vector rank]
\label{def:vector-rank}
    Let $p$ be an absolutely continuous distribution on $\R^d$ with support in a convex set. Let $\mu$ be the uniform distribution on $[0, 1]^{d}$. There exists a convex function $\psi$ such that $\nabla \psi \# \mu = p$ and $\nabla \psi^{-1} \# p = \mu$. $\vecu \coloneqq \nabla \psi^{-1}$ is called the vector rank associated with $p$ ~\cite{chernozhukov2017monge}.
\end{Definition}
When $d=1$, vector rank reduces to standard scalar rank. Intuitively, vector rank transforms a multivariate distribution $p$ to a (multivariate) uniform distribution $\mu$, entirely removing its characteristics.


In the text below, we slightly overload this definition and use the term `vector rank' to refer to both the vector rank function $\vecu(\cdot)$ and also the corresponding random variable $\vecu$ induced by this function.


\begin{Definition}[Vector copula]
\label{def:vector-copula}
    Let $\vecu_X$ and $\vecu_Y$ be the vector ranks corresponding to $p(\vecx)$ and $p(\vecy)$ respectively. A vector copula $C(\vecu_X, \vecu_Y)$ is a cumulative distribution function on $[0, 1]
^{d_X + d_Y}$ with uniform marginals on $C(\vecu_X
) = \mathcal{U}[0, 1]^{d_X}$ and $C(\vecu_Y
) = \mathcal{U}[0, 1]^{d_Y}$. The probabilistic density function corresponding to $C$ is called \emph{vector copula density} and is denoted as $c(\vecu_X, \vecu_Y)$~\cite{fan2023vector}.
\end{Definition}


Given the above definition, we have the following result~\cite{fan2023vector} generalizing the Sklar theorem~\cite{sklar1959fonctions}.


\begin{Theorem}[Vector Sklar Theorem]
\label{thm:vector-sklar-theorem}
Let $X \in \R^{d_X}$ and $Y \in \R^{d_Y}$ be two random variables with joint distribution $p(\vecx, \vecy)$ on $\R^{d_X + d_Y}$. For any absolutely continuous distributions $p(\vecx, \vecy)$ with support in a convex set, there exist an unique function $c(\cdot, \cdot)$, such that
\begin{equation}
    p(\vecx, \vecy) = p(\vecx)p(\vecy)c(\vecu_X, \vecu_Y)
    \label{eq:vector-sklar-theorem}
\end{equation}
where $\vecu_X$ and $\vecu_Y$ are the vector ranks computed for $\vecx$ and $\vecy$ respectively. The function $c$ equals to the vector copula density associated with $\vecu_X$ and $\vecu_Y$~\cite{fan2023vector}.
\end{Theorem}

The vector Sklar theorem suggests that for a distribution $p(\vecx, \vecy)$, its marginal distributions and the joint dependence structure are entirely disentangled, with the latter fully characterized by the vector copula density $c$. Note that here we focus on the case of two RVs; we refer to~\cite{fan2023vector} for general cases.

\textbf{Instances of vector copula}. We discuss several instances of vector copula related to our work. One important instance is the \emph{vector Gaussian copula}~\cite{fan2023vector}.  This model assumes that the joint dependence structure admits a Gaussian structure, with its vector copula $C^{\mathcal{N}}$ being
\begin{equation}
    C^{\mathcal{N}}(\vecu_X, \vecu_Y) = \Phi(\phi^{-1}(\vecu_X), \phi^{-1}(\vecu_Y); \mathbf{0}, \Sigma)
\end{equation}
where $\Sigma = [[\mathbf{I}_X, \Sigma_{XY}] ,[\Sigma^{\top}_{XY}, \mathbf{I}_Y]]$ is a p.s.d matrix whose blocks $\mathbf{I}_X \in \R^{d_X \times d_X}$ and $\mathbf{I}_Y \in \R^{d_Y \times d_Y}$ are identity matrices. $\Phi(\cdot)$ is the cumulative distribution
function of multivariate normal distribution and $\phi(\cdot)$ is the (element-wise) cumulative distribution
function of univariate normal distribution. Equivalently, a vector Gaussian copula can be defined by its data generation process: $\epsilon \sim \mathcal{N}(\epsilon; \mathbf{0}, \Sigma)$, $\vecu_X = \phi(\epsilon_{\leq d_X})$, $\vecu_Y = \phi(\epsilon_{> d_X})$, with $\epsilon_{\leq d_X}$ and $\epsilon_{> d_X}$ being the first $d_X$ and the remaining dimensions of $\epsilon$ respectively. An analytic expression for $c^{\mathcal{N}}$ can be derived accordingly. 

Other useful instances of vector copula include $t$-vector copula, Archimedean vector copula and Kendall vector copula, which correspond to different inductive biases about the dependence structure.

\section{Methodology}
In this section, we propose a new mutual information (MI) estimator based on vector copula theory. The core of our method is Theorem 2, which establishes a connection between MI and vector copula:

\begin{Theorem}[MI is vector copula entropy]
\label{thm:main_theorem}
The mutual information $I(X; Y)$ is the negative differential entropy of the vector copula density: 
\begin{equation}
    I(X; Y) = -H[c(\vecu_X, \vecu_Y)] 
    \label{eq:MI_is_copula_entropy}
\end{equation}
where $\vecu_X$ and $\vecu_Y$ are the vector ranks corresponding to $p(\vecx)$ and $p(\vecy)$ respectively.
\end{Theorem}
\emph{Proof}: Please refer to Appendix A.  \qed

This theorem generalizes the results of~\cite{davy2003copulas, ma2011mutual} from univariate to vector marginals\footnote{Building upon classic copula, the theory in~\cite{davy2003copulas, ma2011mutual} only holds for bivariate cases, and generalizing their results to high-dimensional cases require non-trivial formulation and derivation—precisely our key contribution.}. It establishes that MI depends solely on the vector copula, which itself is invariant to marginal distributions. Notably, the theorem also reveals that the pointwise mutual information (PMI) i.e. $p(\vecx, \vecy)/p(\vecx)p(\vecy)$ can equivalently be viewed as a \emph{density} $c(\vecu_X, \vecu_Y)$ in its own right, in contrast to the vast majority of existing works~\cite{belghazi2018mine, huinfonet, butakov2024mutual, czyz2023mixtures} which continue to treat PMI as a \emph{density ratio}. This shift in perspectives opens us new possibility in the parameterization and learning of the PMI, including directly modeling it as a normalized density learned via MLE, as will be discussed later.

Theorem \ref{thm:main_theorem} immediately suggests a new \emph{divide-and-conquer} approach for MI estimation: we can first estimate the vector ranks $\vecu_X$ and $\vecu_Y$, followed by subsequent learning of the vector copula $c$\footnote{Alternatively, one may also  learn the marginals $\hat{p}(\vecx), \hat{p}(\vecy)$ and the vector copula $\hat{c}$ jointly. However, joint learning can be ill-posed~\cite{ashok2023tactis}. Our ablation study in Appendix B2  suggests that separate learning is more robust.  }: 
\begin{equation}
    I(X; Y) \approx \hat{I}(X; Y) \coloneqq \frac{1}{n} \sum^n_{i=1} \log \hat{c}(\hat{\vecu}_X^{(i)}, \hat{\vecu}_Y^{(i)})
    \label{eq:MC_VCE}
\end{equation}
where $\hat{\vecu}_X$, $\hat{\vecu}_Y$ and $\hat{c}$ are the empirical estimates to $\vecu_X$, $\vecu_Y$ and $c$ respectively.

We discuss below several potential advantages of the above divide-and-conquer estimation strategy:

\begin{itemize}[leftmargin=*]
    \item By disentangling the \emph{modeling} of marginals distribution and copula, we can use differently-sized models in their parameterization, avoiding using a single overly flexible or overly simplified model for the entire distribution. This leads to a better trade-off between model complexity and capacity;
    \item By disentangling the \emph{learning} of marginals and copula, we can reuse the pre-trained marginals across multiple copula choices with varying complexities, allowing model selection to be performed solely in the copula space in a computational efficient way. It also reduces overall learning difficulty.
\end{itemize}

In the following, we elaborate methods to estimate the vector ranks and the vector copula respectively.

\begin{figure*}[!t]
\begin{minipage}[!t]{0.497\linewidth}
\centering
\vspace{-0.01cm}
\begin{algorithm}[H]
  \caption{Vector copula MI estimate (VCE)}
  \label{alg:snl+}
\begin{algorithmic}
  \vspace{0.07cm}
  \STATE {\bfseries Input:} data $\mathcal{D} = \{\vecx^{(i)}, \vecy^{(i)}\}^n_{i=1}$
  \STATE {\bfseries Output:} estimated $\hat{I}(X; Y)$
  \STATE {\bfseries Parameters:} flows $f_X,f_Y$, copulas $\{c_1, ..c_M\}$ 
  \STATE {\bfseries Initialization:} $\mathcal{D} = \mathcal{D}_{train} \cup \mathcal{D}_{val}$, $K=1$,
  \vspace{0.50cm}
  \STATE $\triangleright \enskip$ \emph{Marginal distributions learning} \\
  \STATE learn $f_X$ with $\mathcal{D}_X = \{\vecx^{(i)}\}^{n}_{i=1}$ by FM;
  \STATE learn $f_Y$ with $\mathcal{D}_Y = \{\vecy^{(i)}\}^{n}_{i=1}$ by FM;
  \FOR{$i$ in $1$ to $n$ } 
     \STATE compute $\hat{\vecu}_X^{(i)} = {\tt{rank}}(f_X(\vecx^{(i)}))$;
     \STATE compute $\hat{\vecu}_Y^{(i)} = {\tt{rank}}(f_Y(\vecy^{(i)}))$;
  \ENDFOR 
  \vspace{0.50cm}
  \STATE $\triangleright \enskip$ \emph{Vector copula density estimation} \\
  \REPEAT
        \STATE set $c(\vecu_X, \vecu_Y) = \frac{1}{K} \sum^{K}_{k=1} p_k c_k(\vecu_X, \vecu_Y)$;
        \vspace{0.06cm}
        \STATE $\hat{c} = \arg\max_{c} \mathbb{E}_{\hat{\vecu}_X, \hat{\vecu}_Y \sim \mathcal{D}_{train}}[\log c(\hat{\vecu}_X, \hat{\vecu}_Y) ]$;
        \STATE $\mathcal{L}_{\text{val}} \leftarrow \mathbb{E}_{\hat{\vecu}_X, \hat{\vecu}_Y \sim \mathcal{D}_{val}}[\log c(\hat{\vecu}_X, \hat{\vecu}_Y) ]$;
        \vspace{0.06cm}
        \STATE $K \leftarrow 2K$;
  \UNTIL{no improvement on $\mathcal{L}_{\text{val}}$}
  \STATE \textbf{return} $\hat{I}(X; Y) = \frac{1}{n} \sum^n_{i=1} \log \hat{c}(\hat{\vecu}_X^{(i)}, \hat{\vecu}_Y^{(i)})$
\end{algorithmic}
\end{algorithm}
\end{minipage}
\begin{minipage}[!t]{0.497\linewidth}
\centering
\begin{algorithm}[H]
  \caption{Vector copula MI estimate' (VCE')}
\begin{algorithmic}
  \vspace{0.07cm}
  \STATE {\bfseries Input:} data $\mathcal{D} = \{\vecx^{(i)}, \vecy^{(i)}\}^n_{i=1}$
  \STATE {\bfseries Output:} estimated $\hat{I}(X; Y)$
  \STATE {\bfseries Parameters:} flows $f_X$, $f_Y$,  ratio estimator $r$
  \STATE {\bfseries Initialization:} reference copula $c'$, $\mathcal{D}' = \emptyset$
  \vspace{0.50cm}
  \STATE $\triangleright \enskip$ \emph{Marginal distributions learning} \\
  \STATE learn $f_X$ with $\mathcal{D}_X = \{\vecx^{(i)}\}^{n}_{i=1}$ by FM;
  \STATE learn $f_Y$ with $\mathcal{D}_Y = \{\vecy^{(i)}\}^{n}_{i=1}$ by FM;
  \FOR{$i$ in $1$ to $n$ } 
     \STATE compute $\hat{\vecu}_X^{(i)} = {\tt{rank}}(f_X(\vecx^{(i)}))$;
     \STATE compute $\hat{\vecu}_Y^{(i)} = {\tt{rank}}(f_Y(\vecy^{(i)}))$;
  \ENDFOR 
  \vspace{0.50cm}
  \STATE $\triangleright \enskip$ \emph{Vector copula density estimation} \\
  \REPEAT
    \STATE sample $\vecu^{(j)}_X, \vecu^{(j)}_Y \sim c'(\vecu_X, \vecu_Y)$;
    \STATE $\mathcal{D}' \leftarrow \mathcal{D}' \cup \{ \vecu^{(j)}_X, \vecu^{(j)}_Y \}$;
  \UNTIL{$|\mathcal{D}'| = n$}
  \STATE train $r$ to classify samples from $\mathcal{D}$ and $\mathcal{D}'$;
  \STATE set $\hat{c}(\vecu_X, \vecu_Y) = r(\vecu_X, \vecu_Y)\cdot c'(\vecu_X, \vecu_Y)$;
  \STATE \textbf{return} $\hat{I}(X; Y) = \frac{1}{n} \sum^n_{i=1} \log \hat{c}(\hat{\vecu}_X^{(i)}, \hat{\vecu}_Y^{(i)})$
\end{algorithmic}
\end{algorithm}
\end{minipage}

\end{figure*}

\clearpage

\subsection{Marginal distribution learning}
In this step, we learn the two marginal distributions $p(\vecx)$ and $p(\vecy)$ with flexible flow-based models~\cite{papamakarios2017masked, durkan2019neural, papamakarios2021normalizing, lipman2022flow} and use them to compute the vector ranks $\vecu_X$ and $\vecu_Y$. 

\textbf{Flow-based modeling of marginals}. Let $f_X: \R^{d_X} \to \R^{d_X}$ and $f_Y: \R^{d_Y} \to \R^{d_Y}$ be two flow-based models  and let $p_{f_X}(\vecx)$ and $p_{f_Y}(\vecy)$ be the  densities induced by $f_X$ and $f_Y$ respectively. We respectively learn $f_X$ and $f_Y$ with data $\vecx \sim p(\vecx)$ and data $\vecy \sim p(\vecy)$ by flow matching~\cite{lipman2022flow}:
\begin{equation}
    \min_{f_X} \mathbb{E}[\mathcal{L}_{\text{FM}}(\vecx; f_X)], \qquad \min_{f_Y} \mathbb{E}[\mathcal{L}_{\text{FM}}(\vecy; f_Y)]
\end{equation}
where $\mathcal{L}_{\text{FM}}$ denotes the flow-matching loss~\cite{lipman2022flow}. Upon convergence, $f_X$ and $f_Y$ respectively transform the two marginals to a standard normal distribution: $\mathcal{N}(\mathbf{0}, \mathbf{I}) \approx f_X \# p(\vecx)$ and $\mathcal{N}(\mathbf{0}, \mathbf{I}) \approx f_Y \# p(\vecy)$.

\textbf{Vector ranks computation}. With the learned flows $f_X$ and $f_Y$, we compute the vector ranks as:
\begin{equation} \label{eq:vrs_flows}
    \hat{\vecu}_X^{(i)} = {\tt{rank}}(f_X(\vecx^{(i)})),  \qquad \hat{\vecu}_Y^{(i)} = {\tt{rank}}(f_Y(\vecy^{(i)}))
\end{equation}
where ${\tt{rank}}_d(\pmb{\epsilon}) =  \frac{1}{n+1} \sum^{n}_{j=1} \mathbf{1}[\epsilon_d \geq \epsilon^{(j)}_d]$ is the element-wise ranking function that computes the scalar ranks for each of the dimension in $\epsilon$. Given universal density approximators $f_X$, $f_Y$, $\hat{\vecu}_X$ and $\hat{\vecu}_Y$ serve as consistent estimates of the true vector ranks $\vecu_X$ and $\vecu_Y$.

While the joint density $p(\vecx, \vecy)$ is often challenging to estimate, the marginal distributions $p(\vecx)$ and $p(\vecy)$ are typically far easier to learn due to their lower dimensionality. It is thus reasonable to expect that  $\hat{\vecu}_X$ and $\hat{\vecu}_Y$ are close approximations to $\vecu_X$ and $\vecu_Y$ in moderate dimensionality settings.

\emph{Remark}. The above process of estimating $\vecu_X$ and $\vecu_Y$ can be viewed as a generalization of classic copula transformation in MI estimation, where we compute vector ranks rather than scalar ranks.

\subsection{Vector copula estimation}
In this step, we learn the vector copula $c$ with the previously estimated vector ranks $\hat{\vecu}_X$ and $\hat{\vecu}_Y$, leveraging a model-based parameterization and a careful model selection strategy.

\textbf{Model-based parameterization of copula}. As noted earlier, any parametric model can be used to represent the vector copula  $c$, regardless of whether an analytical PMI is available. In this work, we parameterize $c$ as a mixture of existing parametric vector copulas~\cite{fan2023vector} from the copula pool, whose model complexity can be well controlled by tuning the number of mixture components:
\begin{equation}
    c(\vecu_X, \vecu_Y) =  \sum^K_{k=1} p_k c_k(\vecu_X, \vecu_Y), 
    \label{eq:vector-copula-design}
\end{equation}
where $\sum^K_{k=1} p_k = 1$ and each $c_k \in \mathcal{C}$ is selected from the predefined pool $\mathcal{C}$ of vector copulas. Any inductive bias about the dependence structure can be used to guide copula selection. Here, we simply implement each $c_k$ as a vector Gaussian copula and learn $c$ by maximum likelihood estimate (MLE):
\begin{equation}
\begin{split}
    \max_c \mathbb{E} [\log c(\vecu_X&, \vecu_Y)]
    \label{eq:learn-copula-density}
\end{split}
\end{equation}
In theoretical analysis, we analyze why this copula design is a cheap yet reasonable modeling of $c$.

\textbf{Efficient model selection}. A key design in our method is the explicit exploration of the capacity–complexity trade-off in copula modeling, which is governed by the number of mixture components $K$. Here, we determine $K$ by cross validation, using negative log-likelihood (NLL) as the criterion. This process is computationally cheap: each copula is already lightweight, involving no neural networks; furthermore, different copulas can be trained in parallel using one single loss.



Algorithm 1 summarizes the main pipeline of the proposed vector copula-based estimator (VCE).

\emph{Remark}. As an alternative to the above model-based parameterization, one may also adopt a reference-based parameterization for the vector copula, inspired by the design in~\cite{huk2025your}. Specifically, let $c'$ be a reference vector copula that is easy to sample (e.g. a vector Gaussian copula). We can learn $c$ by first estimating the  density ratio $r = c/c'$ using samples from $c$ and $c'$~\cite{gutmann2010noise, srivastava2023estimating, liao2020demi}, then recover the vector copula $c$ as $c = r \cdot c'$; see Algorithm 2. By parameterizing $r$ as a deep neural network, this method allows for a more flexible modeling of $c$, at the cost of a less fine-grained control over its complexity.

\clearpage



\section{Theoretical analysis}
In this section, we analyze several important theoretical properties of the proposed VCE estimator. 

\begin{Proposition}[Consistency of VCE]
Assuming that (a) $f_X$ and $f_Y$ are universal PDF approximators with continuous support and (b) the number of mixture components $K$ in~\eqref{eq:vector-copula-design} is sufficiently large. Define $\hat{I}_n(X; Y) \coloneqq \frac{1}{n} \sum_{i=1}^n \log \hat{c}(\hat{\vecu}_X^{(i)}, \hat{\vecu}_Y^{(i)})$. For every $\epsilon > 0$, there exists $n(\varepsilon) \in \mathbb N$, such that 
\[
     \left|\hat{I}_n(X; Y) - I(X; Y)\right| < \varepsilon, \quad \forall n \geq n(\varepsilon), a.s.
\]
\end{Proposition}

\emph{Proof}. Please refer to Appendix A \qed

Additionally, we have the following result analyzing the estimation error w.r.t the quality of the learned marginals $p_{f_X}(\vecx), p_{f_Y}(\vecy)$ and the estimated vector copula density $\hat{c}$.

\begin{Proposition}[Error of vector copula-based MI estimate]
\label{thm:error-all}
Let $\hat{\vecu}_X$ and $\hat{\vecu}_Y$ be the estimated vector ranks. Let $c(\hat{\vecu}_X, \hat{\vecu}_Y)$ be the true joint distribution of $\hat{\vecu}_X$ and $\hat{\vecu}_Y$, and $\hat{c}(\hat{\vecu}_X, \hat{\vecu}_Y)$ its estimate. Assuming that  sufficient Monte Carlo samples are used to compute $\hat{I}(X; Y)$ in \eqref{eq:MC_VCE}, we have
\[
    \Big | I(X; Y) - \hat{I}(X; Y) \Big| \leq \Big| H(\hat{\vecu}_X) + H(\hat{\vecu}_Y) \Big | +  KL [c(\hat{\vecu}_X, \hat{\vecu}_Y)) \| \hat{c}(\hat{\vecu}_X, \hat{\vecu}_Y) ]  
\]
where the first term on the RHS vanishes as $p_{f_X}(\vecx) \to p(\vecx)$ and $p_{f_Y}(\vecy) \to p(\vecy)$. In the limit of perfectly learned marginals, the error simplifies to
\[
    | I(X; Y) - \hat{I}(X; Y)| = KL[c \| \hat{c}],
\]
with $c$ and $\hat{c}$ being the true vector copula density and estimated vector copula density, respectively.
\end{Proposition}
\emph{Proof}. Please refer to Appendix A.  \qed

Proposition \ref{thm:error-all} decomposes the estimation error of the proposed VCE estimator into two components:

\begin{itemize}[leftmargin=*]
    \item \emph{Marginal estimation error}.  Imperfect marginal estimations introduce a bias given by $|H(\hat{\vecu}_X) + H(\hat{\vecu}_Y)| > 0$, which diminishes as both marginals are learned more accurately (recall that ideally, we have $\hat{\vecu}_X \sim \mathcal{U}[0, 1]^{d_X}$ and $\hat{\vecu}_Y \sim \mathcal{U}[0, 1]^{d_Y}$). For data with moderate dimensionality, we expect  this bias to be small, as the two marginals are with low-dimensionality, being easy to estimate.
    \item \emph{Dependence structure modeling error}. This error arises from the discrepancy between the estimated copula $\hat{c}$ and the true copula $c$. It depends on two factors: (a) \emph{capacity} - whether the parameterization of $\hat{c}$ is sufficiently expressive to approximate $c$; and (b) \emph{complexity} - how easy $\hat{c}$ can be learned from the limited data. These factors highlight the importance of model selection for the copula $c$.
\end{itemize}


\begin{Proposition}[Vector Gaussian copula as second-order approximation]
\label{thm:taylor-expansion-copula}
A vector Gaussian copula $c^{\mathcal{N}}$ corresponds to the second-order Taylor expansion of the true vector copula $c^*$ up to variable transformation. 
\end{Proposition}
\emph{Proof}. Please refer to Appendix A. \qed

This result explains our choice of using a mixture of Gaussian copulas as a cheap yet principled approximation to the true vector copula. A single vector Gaussian copula already offers a reasonable approximation of the true copula by capturing dependencies up to second order; higher-order interactions, if necessary, can be modeled by adding mixture components in a fully controllable way.

Finally, we have the following result regarding cases with weakly dependent random variables (RVs).

\begin{Proposition}[Vector copula of independent RVs]
\label{thm:prod-marginal}
The vector copula corresponding to the product of marginals $p'(\vecx, \vecy) = p(\vecx)p(\vecy)$ is a vector Gaussian copula if $p'(\vecx, \vecy)$ is absolutely continuous.
\end{Proposition}
\emph{Proof}. Please refer to Appendix A. \qed

Proposition \ref{thm:prod-marginal} suggests that if the two RVs $X$ and $Y$ are nearly independent, our estimator is \emph{likely} to provide an accurate estimation of $I(X; Y)$ as the true vector copula is Gaussian-like, being close to the family of our copula design~\eqref{eq:vector-copula-design}. For weakly dependent RVs, it is reasonable to expect that  $p(\vecx, \vecy)$ resembles a vector Gaussian copula, with the difference captured by the additional components in~\eqref{eq:vector-copula-design}.


\section{Reinterpreting existing MI estimators}
In this section, we reinterpret existing MI estimators through the lens of vector copula theory, showing that they correspond to different parameterizations and learning strategies of the vector copula.

\textbf{Reinterpreting discriminative estimators}. Existing critic-based approach to MI estimation~\cite{belghazi2018mutual, mukherjee2020ccmi, liao2020demi, srivastava2023estimating, song2019understanding, letizia2024mutual} can be interpreted as parameterizing the vector copula $c(\vecu_X, \vecu_Y)$ using a feedforwarding neural network $f$:
\begin{equation}
    c(\vecu_X, \vecu_Y) \propto e^{f(\vecx, \vecy)}
\end{equation}
which is learned by discerning samples from the joint $p(\vecx, \vecy)$ and the product of marginals $p(\vecx)p(\vecy)$ (via e.g contrastive learning). Specifically, recall that the optimal critic $f$ in these methods corresponds to the log density ratio up to an additive constant $C$~\cite{poole2019variational}: $f(\vecx, \vecy) = \log p(\vecx, \vecy)/p(\vecx)p(\vecy) + C$, with the PMI itself equal to the vector copula density, as established by the vector Skalar theorem.


Compared to our model-based parameterization of the vector copula density in \eqref{eq:vector-copula-design}, this neural network parameterization is more flexible and can potentially capture more complex dependence structures. However, as discussed earlier, such distribution-free parameterizations lack complexity control, which may lead to a poor bias–variance trade-off. Furthermore, discriminative methods learns the vector copula by comparing distributions, which can be challenging if they differ significantly (e.g., in high-MI cases, see~\cite{rhodes2020telescoping, srivastava2023estimating, mcallester2020formal, song2019understanding}.  In contrast, our main method learns the vector copula by maximum likelihood estimate (MLE), which is the most efficient consistent estimator for the copula.

\textbf{Reinterpreting generative estimators}. \noindent Many generative estimators for MI~\citep{duong2023diffeomorphic, mcallester2020formal, cheng2020club, butakov2024mutual, song2019understanding} require either learning the joint distribution $p(\vecx, \vecy) = p(\vecx)p(\vecy)c(\vecu_X, \vecu_Y)$ or the conditional distribution $p(\vecy|\vecx) = p(\vecy)c(\vecu_X, \vecu_Y)$ using a single model. This process can be interpreted as learning the marginal distribution(s) and the vector copula  simultaneously, with the two components parameterized \emph{jointly} via a \emph{single} generative model. Our method, on the contrary, explicitly separates the modeling and the learning of the marginal distributions $p(\vecx), p(\vecy)$ from that of the vector copula $c(\vecu_X, \vecu_Y)$. This strategy not only enables a more fine-grained control over model complexity, but also mitigates the challenge of jointly learning the marginal distribution and the dependence structure—a strategy aligned with the spirit of classical copula transformations~\cite{huinfonet, safaai2018information, zeng2018jackknife, purkayastha2024fastmi} to simplify MI estimation.  

We further discuss two recent works~\cite{butakov2024mutual, duong2023diffeomorphic} closely related to our work. These methods operate by respectively transforming the two RVs $X$ and $Y$ by two flow-based models, such that the joint distribution of the transformed data can be approximated by a distribution with an easy-to-compute MI (for instance, a Gaussian distribution). Their practical methods, $\mathcal{N}$-MIENF and DINE-Gaussian, can be reinterpreted as assuming the dependence structure as a vector Gaussian copula (see Lemma 3 in Appendix A4 for a detailed derivation):
\begin{equation}
    c(\vecu_X, \vecu_Y) \approx c^{\mathcal{N}}(\vecu_X, \vecu_Y; \Sigma)
\end{equation}
which corresponds to the case $K=1$ in the VCE estimator and is accurate (only) if the true dependence is Gaussian-like. The possibility of using non-Gaussian base distribution is also discussed in~\cite{butakov2024mutual}, albeit without practical implementation. Additionally, the marginals and the vector copula in their method are learned jointly rather than separately as in our method, and they continue to treat PMI as a density ratio $p(\vecx, \vecy)/p(\vecx)p(\vecy)$, unlike our method which treats it as a density $c(\vecu_X, \vecu_Y)$.





\section{Experiments}

\begin{figure*}[t!]
            \hspace{-0.02\linewidth}
            \begin{subfigure}{.205\textwidth}
                    \centering
                    \begin{minipage}[t]{1.0\linewidth}
                    \centering
                     \includegraphics[width=1.0\linewidth]{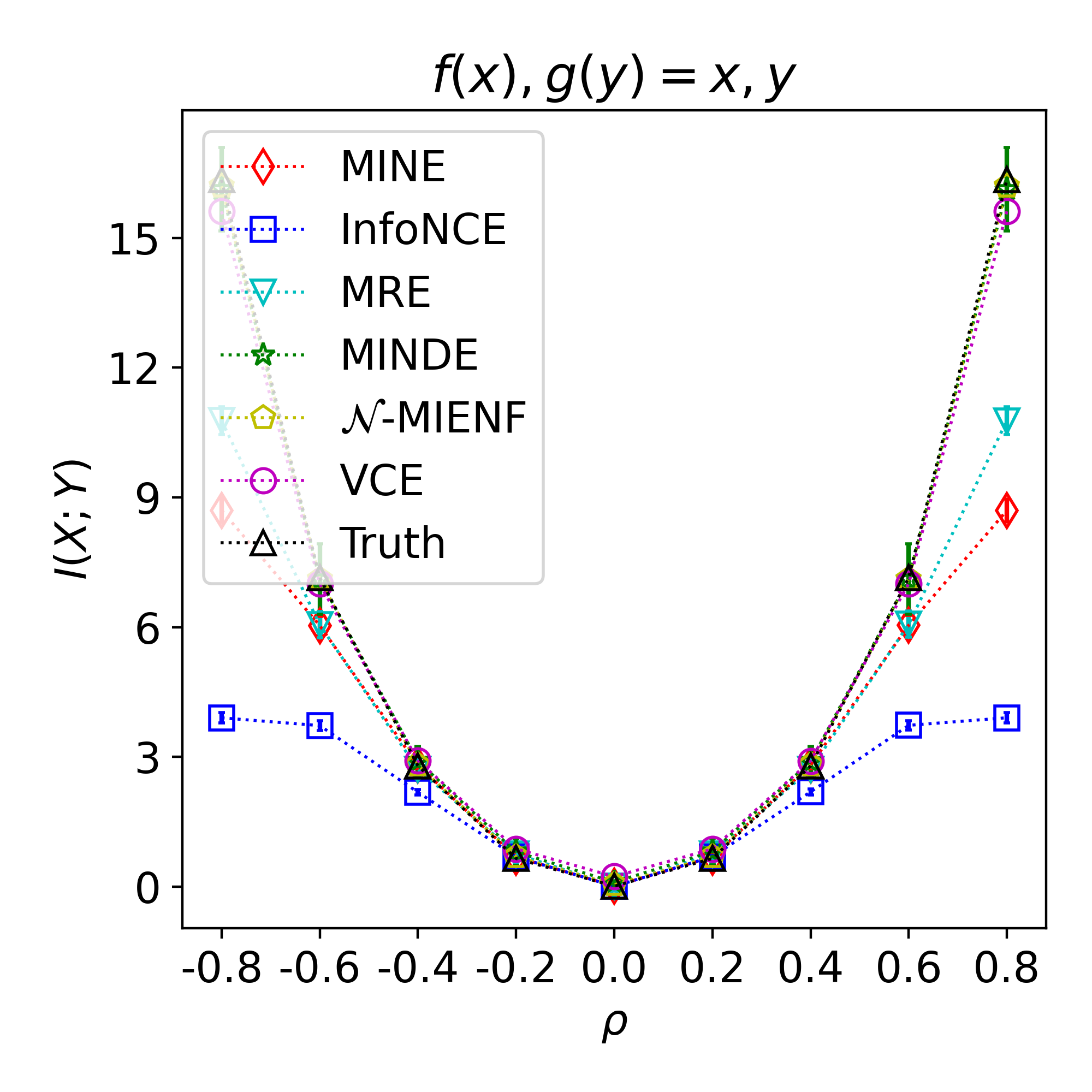}
                    \end{minipage}
            \subcaption{\centering $X, Y \sim \mathcal{N}$}
            \end{subfigure}
            \hspace{-0.015\linewidth}
            \begin{subfigure}{.205\textwidth}
                    \centering
                    \begin{minipage}[t]{1.0\linewidth}
                    \centering
                     \includegraphics[width=0.99\linewidth]{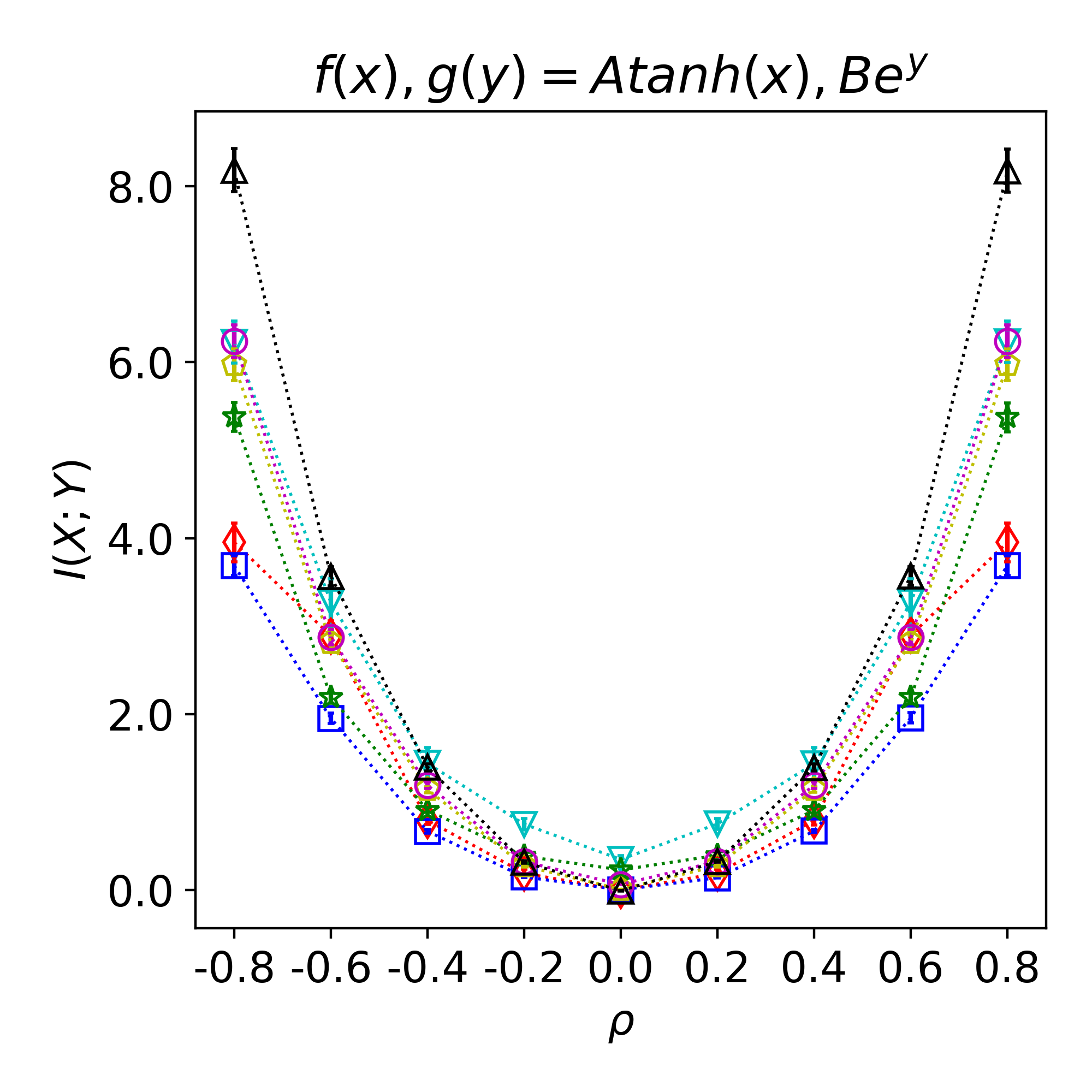}
                    \end{minipage}
            \subcaption{\centering $\mathbf{A}\text{tanh}(X), \mathbf{B}e^Y$}
            \end{subfigure}
            \hspace{-0.015\linewidth}
            \begin{subfigure}{.205\textwidth}
                    \centering
                    \label{fig:op}
                    \begin{minipage}[t]{1.0\linewidth}
                    \centering
                     \includegraphics[width=1.0\linewidth]{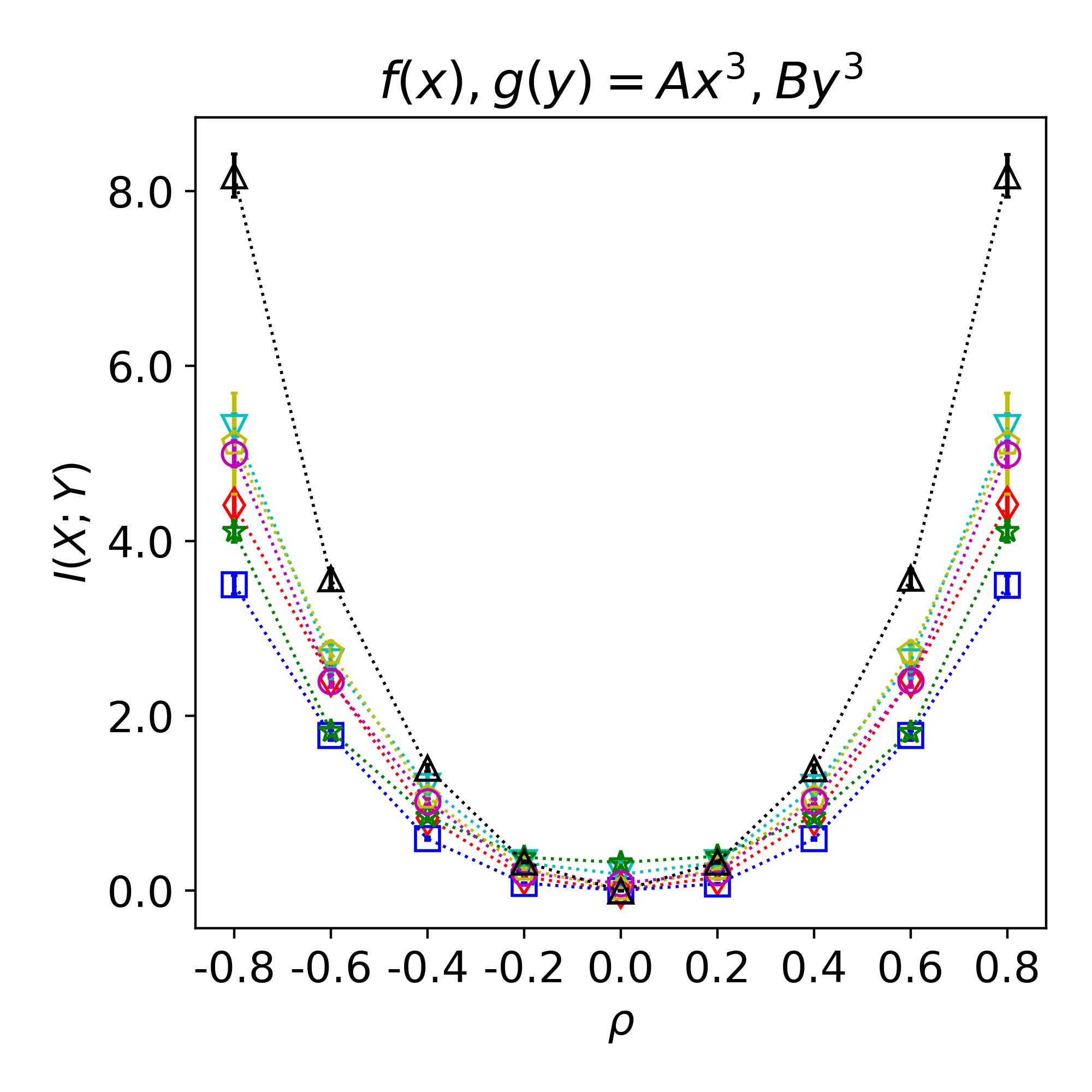}
                    \end{minipage}
            \subcaption{\centering $\mathbf{A}X^3, \mathbf{B}Y^3$}
            \end{subfigure}
            \hspace{-0.015\linewidth}
            \begin{subfigure}{.205\textwidth}
                    \centering
                    \begin{minipage}[t]{1.0\linewidth}
                    \centering
                     \includegraphics[width=1.0\linewidth]{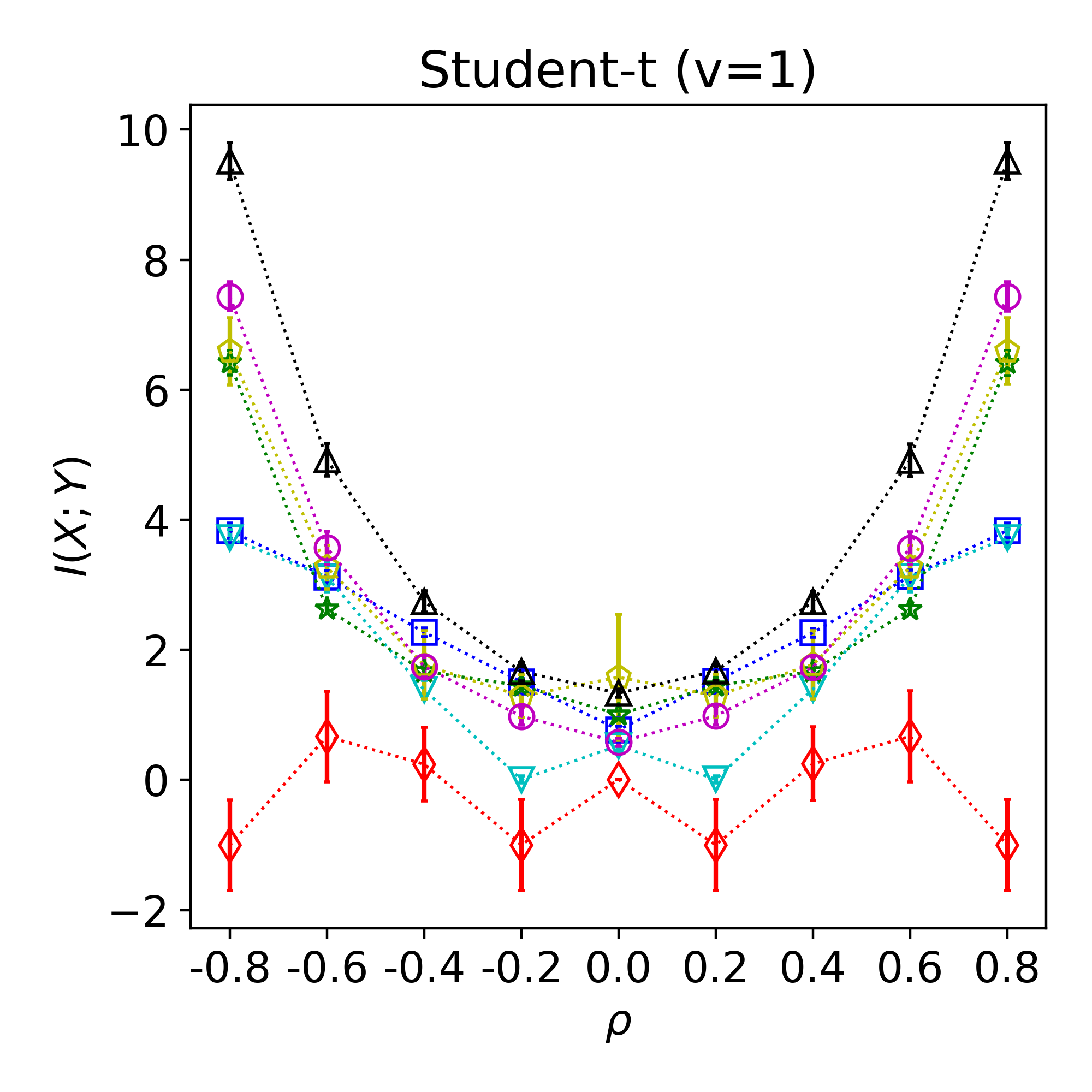}
                    \end{minipage}
            \subcaption{\centering Student-$t$}
            \end{subfigure}
            \hspace{-0.015\linewidth}
            \begin{subfigure}{.205\textwidth}
                    \centering
                    \begin{minipage}[t]{1.0\linewidth}
                    \centering
                     \includegraphics[width=1.0\linewidth]{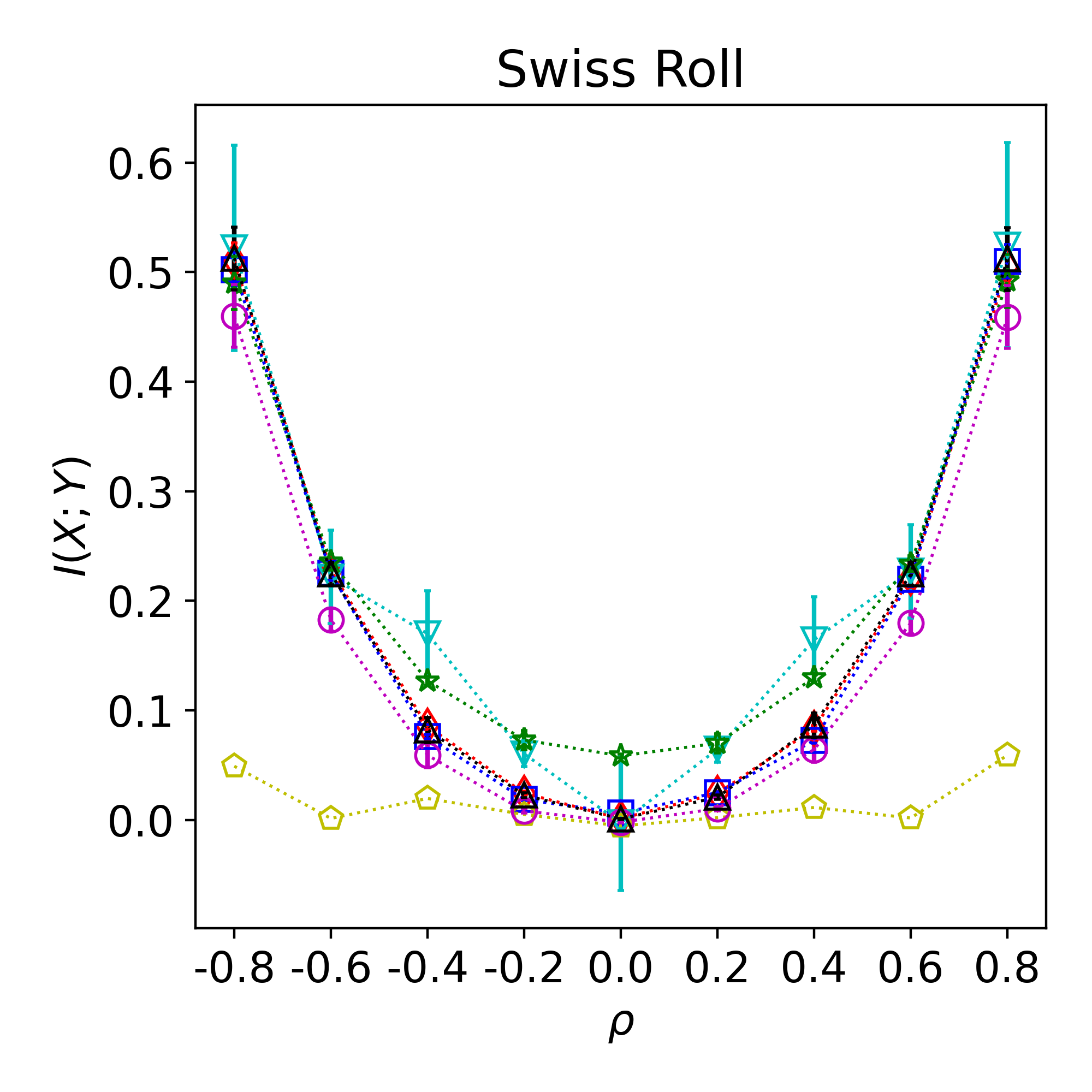}
                    \end{minipage}
            \subcaption{\centering Swiss Roll}
            \end{subfigure}
\vspace{-0.10cm}
\caption[Comparing different MI estimators under various dependence strengths. ]{Comparing MI estimators under various dependence strengths $\rho$. Data in cases (b)(c) are generated by first sampling $X, Y \sim \mathcal{N}$ as in case (a), then transforming them with the shown transformations. The  dimensionalities of the data in the five cases are 64, 32, 32, 32, 2 respectively. }
\label{fig:synthetic:rho}
\end{figure*}

\begin{figure*}[t!]
            \hspace{-0.02\linewidth}
            \begin{subfigure}{.205\textwidth}
                    \centering
                    \label{fig:g-and-k}
                    \begin{minipage}[t]{1.0\linewidth}
                    \centering
                     \includegraphics[width=1.0\linewidth]{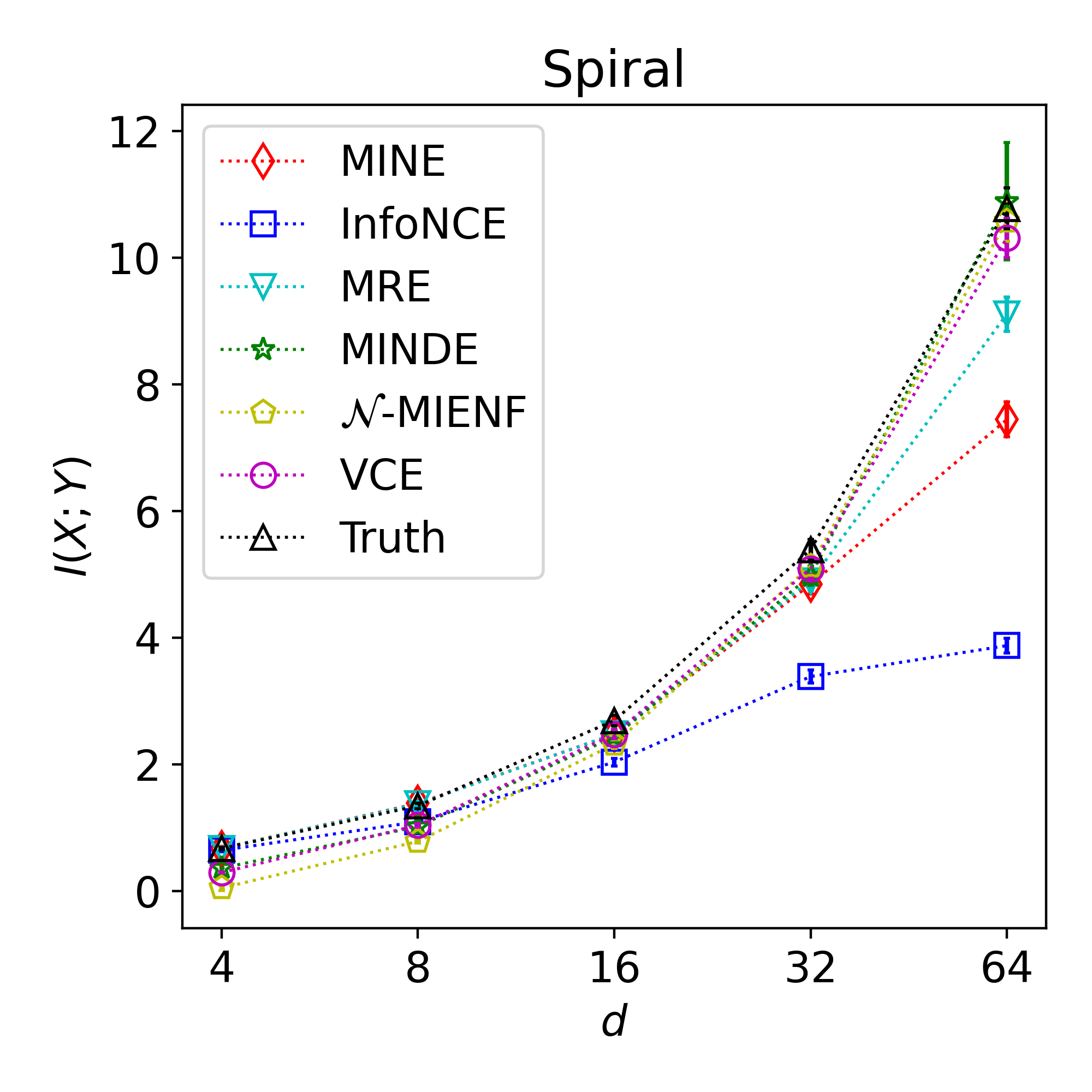}
                    \end{minipage}
            \subcaption{\centering Spiral}
            \end{subfigure}
            \hspace{-0.015\linewidth}
            \begin{subfigure}{.205\textwidth}
                    \centering
                    \label{fig:lr}
                    \begin{minipage}[t]{1.0\linewidth}
                    \centering
                     \includegraphics[width=1.0\linewidth]{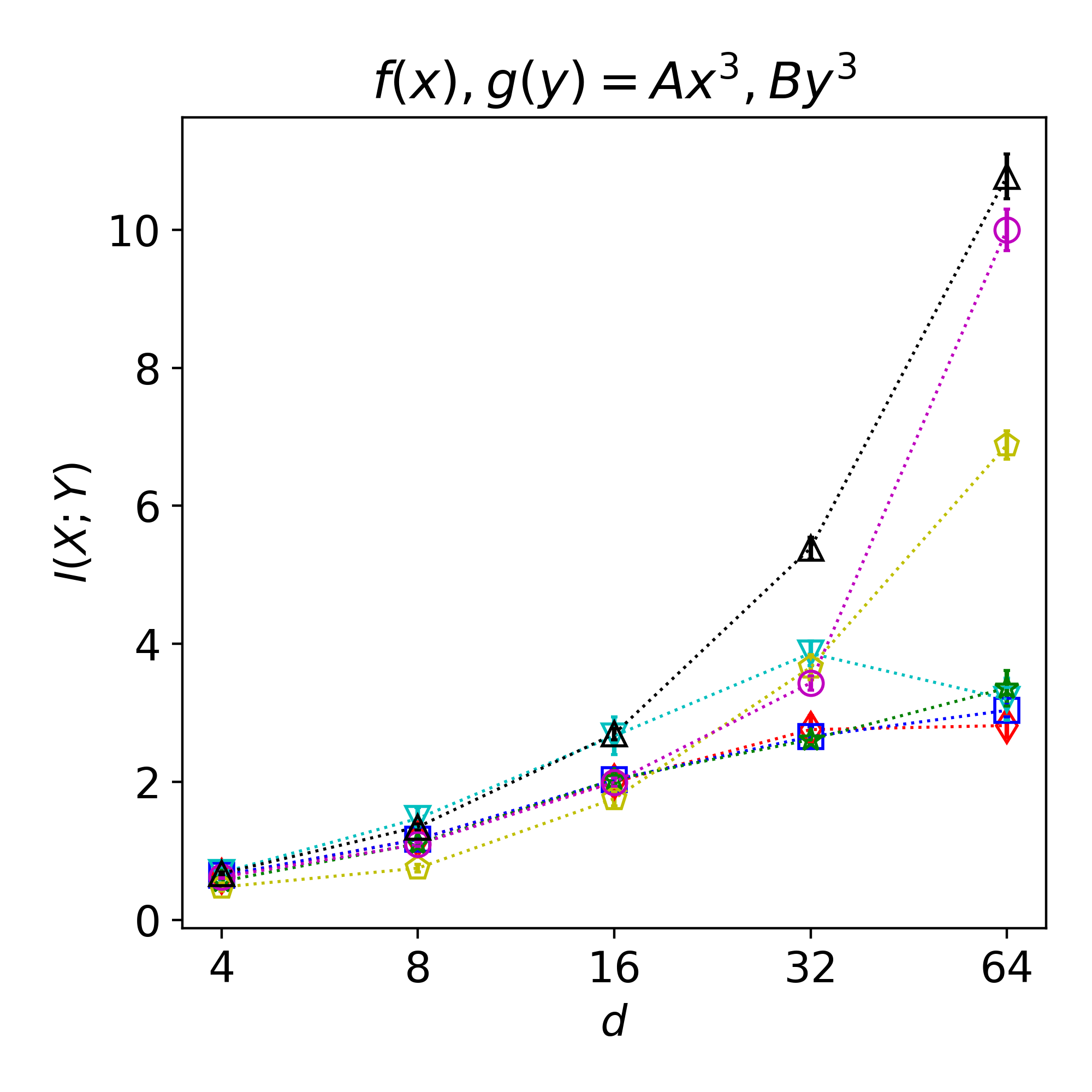}
                    \end{minipage}
            \subcaption{\centering $\textbf{A}X^3, \textbf{B}Y^3$}
            \end{subfigure}
            \hspace{-0.015\linewidth}
            \begin{subfigure}{.205\textwidth}
                    \centering
                    \begin{minipage}[t]{1.0\linewidth}
                    \centering
                     \includegraphics[width=1.0\linewidth]{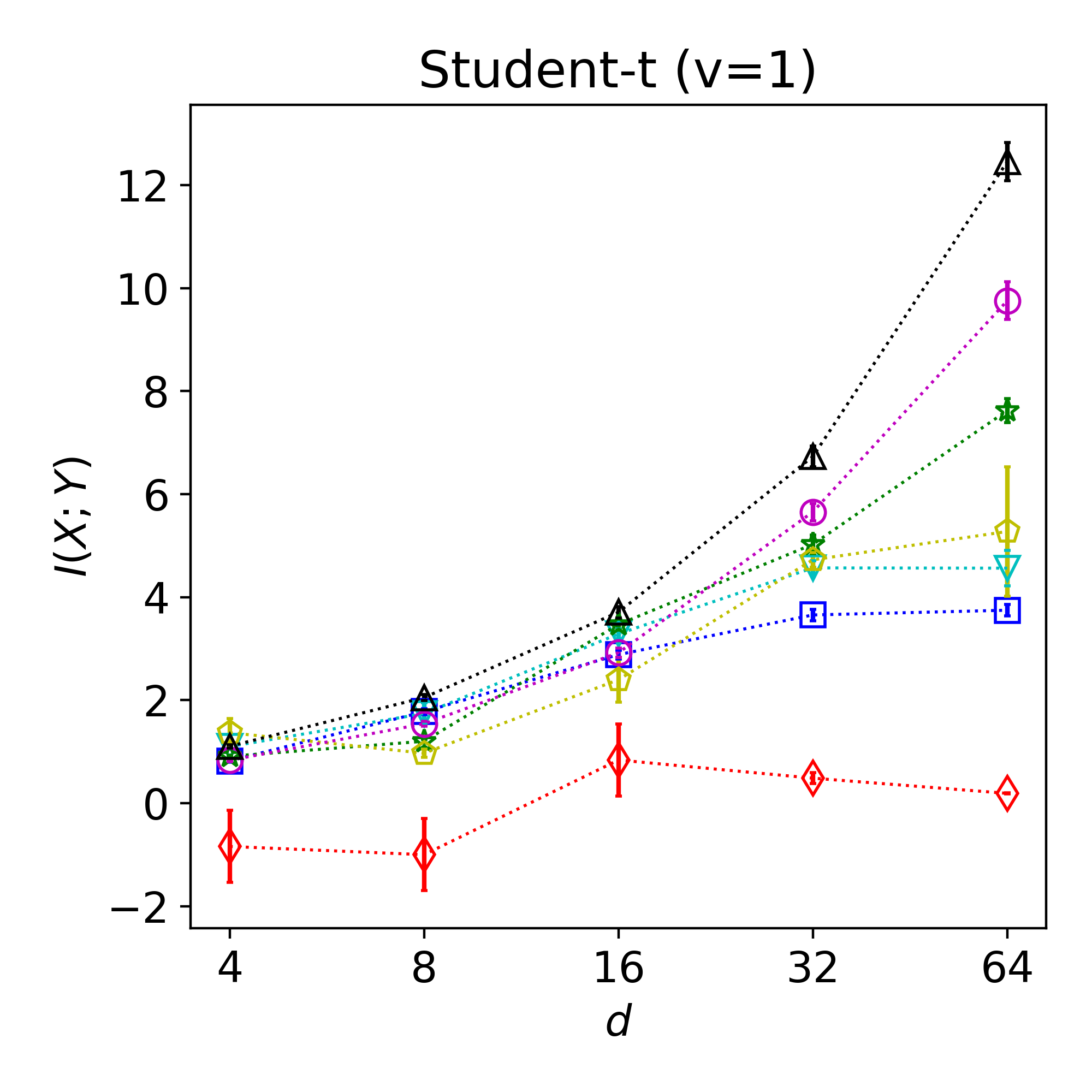}
                    \end{minipage}
            \subcaption{\centering Student-$t$}
            \end{subfigure}
            \hspace{-0.015\linewidth}
            \begin{subfigure}{.205\textwidth}
                    \centering
                    \label{fig:levy}
                    \begin{minipage}[t]{1.0\linewidth}
                    \centering
                     \includegraphics[width=1.0\linewidth]{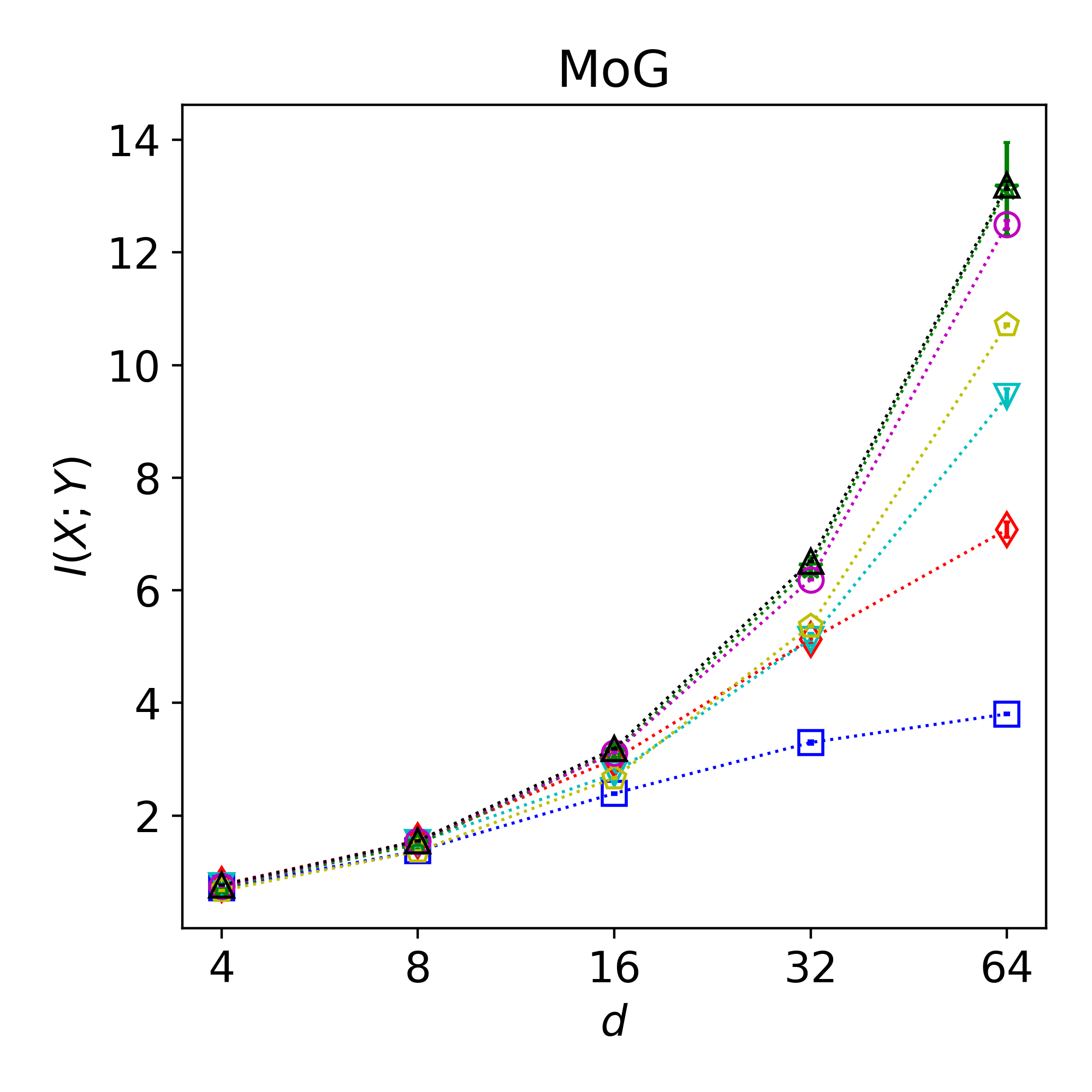}
                    \end{minipage}
            \subcaption{\centering MoG 1}
            \end{subfigure}
            \hspace{-0.015\linewidth}
            \begin{subfigure}{.205\textwidth}
                    \centering
                    \label{fig:ising}
                    \begin{minipage}[t]{1.0\linewidth}
                    \centering
                     \includegraphics[width=1.0\linewidth]{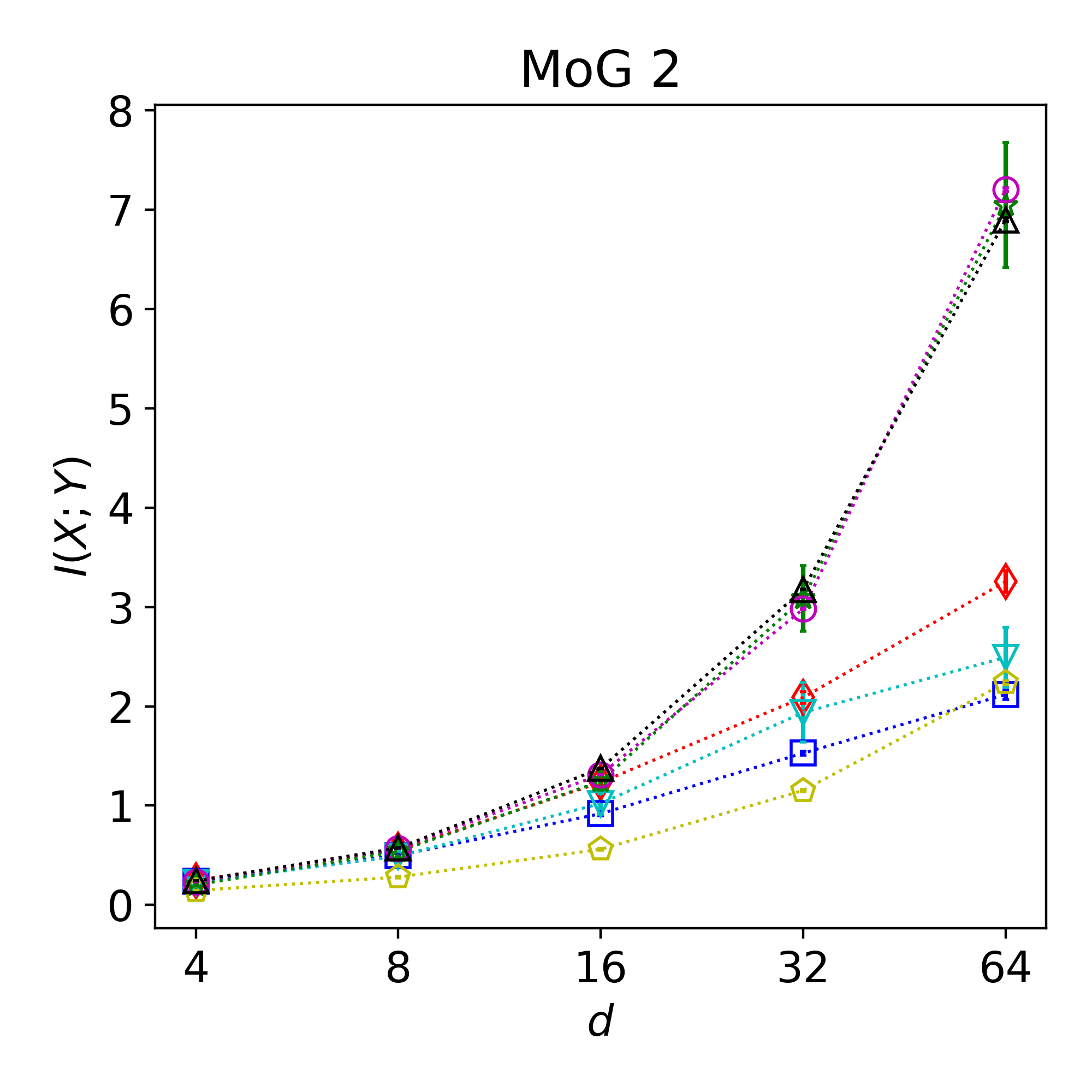}
                    \end{minipage}
            \subcaption{\centering MoG 2}
            \end{subfigure}
\vspace{-0.10cm}
\caption[Comparing different MI estimators under various data dimensionality $d$. ]{Comparing different MI estimators under various data dimensionality $d$ and fixed dependence levels. MoG corresponds to mixture of Gaussians. Spiral corresponds to spiral transformation. }
\label{fig:synthetic:dim}
\end{figure*}

\textbf{Baselines}. We consider five representative neural estimators in the field: MINE \cite{belghazi2018mutual}, InfoNCE \cite{van2018representation}, MRE \cite{srivastava2023estimating}, MINDE \cite{franzeseminde2023} and $\mathcal{N}$-MIENF \cite{butakov2024mutual}. The first three methods are critic-based whereas the latter three are generative model-based. MRE is chosen as the representative of state-of-the-art discriminative methods, which is specifically designed to address the high-discrepancy issue in these methods. MINDE is chosen to represent the state-of-the-art generative methods, which leverages powerful diffusion model in MI estimation. Further baselines are considered in Appendix B2.

\textbf{Hyperparams}. For the vector copula in VCE, we consider mixtures with $1, 4, 8, 16, 32$ components.

\textbf{Neural architecture, optimizer and training details}. Please refer to appendix B1 for more details.

In the following evaluation, we primarily focus on evaluating the VCE estimator (Algorithm 1), and present the results of the alternative VCE' estimator (Algorithm 2) in the appendix. All results are collected through 8 independent runs. Error bars reported are the standard deviations (std) of the runs.

\subsection{Synthetic distributions}

\textbf{Setups}. In~\cite{czyz2023beyond}, a diverse set of models with known MI are developed to comprehensively evaluate MI estimators. We consider representative cases from this benchmark, further extending it by (a) considering varying dependence strengths for each chosen case; (b) employing mixing matrices $\mathbf{A}, \mathbf{B}$ to couple the dimensions in $X$ and $Y$ respectively. We also include the mixture models in~\cite{czyz2023mixtures} to enrich our tests. Together, our test cases cover non-Gaussianity, skewness, heterogeneous marginals, long tails, low-dimensional
manifold structure, coupling dimensions, high-dimensionality, varying dependence strengths and non-Gaussian dependence structure. Each test case contains $n=10^4$ data.

\textbf{Results}. Figure~\ref{fig:synthetic:rho} and Figure~\ref{fig:synthetic:dim} compare the performance of different MI estimators\footnote{Comparison to classic copula-based MI estimators and further discriminative estimators is in Appendix B2. }. Overall, VCE provides good estimates in \emph{all} scenarios, consistently ranking among the top performers.

Compared to discriminative methods e.g., MINE and InfoNCE, VCE demonstrates significant advantages, particularly in high MI settings (e.g. strong dependence level $\rho$ or high dimensionality $d$). This advantage may be because our method avoids directly comparing two highly distinct distributions as in these methods, which is challenging. The advantage may also attribute to the better complexity-capacity trade-off in our method, which avoids an overly powerful model for the copula.  

Compared to the generative method $\mathcal{N}$-MIENF, VCE demonstrates advantages in scenarios involving non-Gaussian dependence structures (see e.g. the MoG cases and 64D $t$-distribution). In such cases, $\mathcal{N}$-MIENF's assumption of a Gaussian dependence structure falls short in capturing the true dependence structure. This underscores the pitfalls of using a overly simplified model for the copula.

We specifically discuss two challenging cases highlighted in prior works~\cite{czyz2023beyond, franzeseminde2023}: (a) Spiral transformation, which highly transforms the original data; and (b) multivariate $t$-distribution with degree of freedom $\nu = 1$, which exhibit heavy-tailed dependence. For these two highly challenging scenarios, VCE and MINDE are the only two methods that can simultaneously provide reasonable estimates in \emph{both} cases, with VCE outperforming MINDE in  other settings (see e.g., Figure~\ref{fig:synthetic:rho}.c and Figure~\ref{fig:synthetic:dim}.b).

\begin{figure*}[t!]
            \hspace{-0.005\linewidth}
            \begin{subfigure}{.49\textwidth}
                    \centering
                    \label{fig:gaussians}
                    \begin{minipage}[t]{1.0\linewidth}
                    \centering
                     \includegraphics[width=1.0\linewidth]{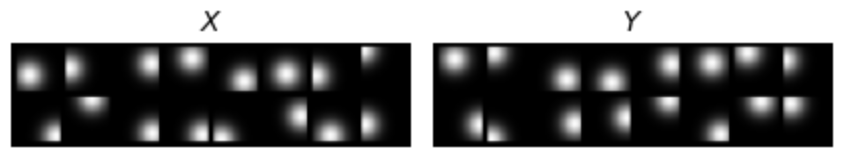}
                    \end{minipage}
            \subcaption{\textbf{Gaussian plates}}
            \end{subfigure}
            \hspace{0.010\linewidth}
            \begin{subfigure}{.495\textwidth}
                    \centering
                    \label{fig:rectangles}
                    \begin{minipage}[t]{1.0\linewidth}
                    \centering
                     \includegraphics[width=1.0\linewidth]{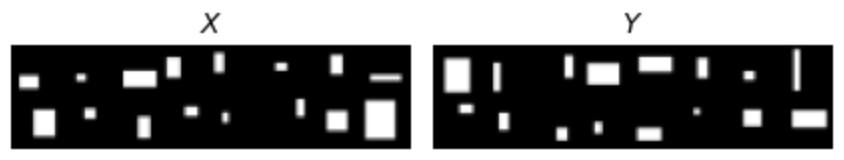}
                    \end{minipage}
            \subcaption{\textbf{Rectangles}}
            \end{subfigure}
\caption[The image benchmark]{The image dataset ~\cite{butakov2024information}, which contains images of rectangles  and  Gaussian plates. }
\label{fig:iamge-dataset-example}
\end{figure*}

\begin{table*}[t]
\centering
\hspace{-0.1cm}
\resizebox{\textwidth}{!}{
\begin{tabular}{lccc p{15pt} ccc}
    \toprule
    \textbf{Method} & \multicolumn{3}{c}{\textbf{Gaussian Plates}} & & \multicolumn{3}{c}{\textbf{Rectangles}} \\
    \cmidrule(lr){2-4} \cmidrule(lr){6-8}
    & $I(X; Y)=1$ & $I(X; Y)=3$ & $I(X; Y)=7$ & & $I(X; Y)=1$ & $I(X; Y)=3$ & $I(X; Y)=7$ \\
    \midrule
    MINE & $0.89 \pm 0.07$ & $2.86 \pm 0.24$ & $5.46 \pm 0.27$ & & $0.81 \pm 0.13$ & $\mathbf{2.57} \pm \mathbf{0.26}$ & $5.39 \pm 0.23$ \\
    InfoNCE & $0.86 \pm 0.14$ & $2.63 \pm 0.13$ & $3.83 \pm 0.12$ & & $0.78 \pm 0.17$ & $2.49 \pm 0.28$ & $3.86 \pm 0.15$ \\
    MRE & $1.23 \pm 0.16$ & $2.85 \pm 0.21$ & $5.91 \pm 0.28$ & & $0.82 \pm 0.24$ & $2.56 \pm 0.48$ & $\mathbf{5.45} \pm \mathbf{0.31}$ \\
    $\mathcal{N}$-MIENF & $0.74 \pm 0.12$ & $2.42 \pm 0.16$ & $3.85 \pm 0.22$ & & $0.54 \pm 0.13$ & $0.76 \pm 0.14$ & $1.54 \pm 0.11$ \\
    \midrule
    VCE & $\mathbf{0.92} \pm \mathbf{0.04}$ & $\mathbf{2.93} \pm \mathbf{0.12}$ & $\mathbf{6.53} \pm \mathbf{0.36}$ & & $\mathbf{0.83} \pm \mathbf{0.12}$ & $2.27 \pm 0.23$ & $5.02 \pm 0.14$ \\
    \bottomrule
\end{tabular}
}
\caption[Comparing different MI estimators on the image dataset. ]{Comparing different MI estimators on the image benchmark proposed in~\cite{butakov2024information}. }
\label{tab:image-dataset-results}
\end{table*}

\subsection{Image dataset with known MI}

\vspace{-0.02cm}

\textbf{Setups}. We next consider the benchmark~\cite{butakov2024information}, which contains correlated images $X$ and $Y$; see Figure \ref{fig:iamge-dataset-example}. Here $X \in \R^{16 \times 16}$ and $Y \in \R^{16 \times 16}$, and the ground truth $I(X; Y)$ is known for this dataset. Following recent works~\cite{butakov2024information, gowri2024approximating}, we preprocess these high-dimensional image data by an autoencoder $e: \R^{16 \times 16} \to \R^{d'}$, which proves effective in  reducing data dimensionality while preserving key information. The quality of such compression w.r.t $d'$ is analyzed theoretically and empirically in Appendix A5 and B2, based on which we set $d'=16$. A total number of $10,000$ data is used. Note that while the dependence between $X$ and $Y$ are Gaussian for this dataset~\cite{butakov2024information}, the dependence structure for the compressed data can be non-Gaussian even if the compression is near-lossless. 


\textbf{Results}. Table~\ref{tab:image-dataset-results} compares the performance of different MI estimators on this task. Our estimator consistently outperforms the recent $\mathcal{N}$-MIENF estimator on this dataset, and it shows highly competitive performance against discriminative methods. However, our method performs slightly worse than discriminative methods in the Rectangles case. One reason why our approach loses to discriminative approaches in the Rectangles case may be that the underlying dependence structure of the preprocessed data is highly complex in this case, which is difficult to model effectively with a single vector Gaussian copula or even a reasonable mixture of such copulas. Discriminative methods, on the contrary, adopt a neural network-based parameterization of the vector copula, being inherently more flexible. These results highlight the limitation of model-based parameterization of the vector copula density in certain cases. Nonetheless, our estimator still provides a highly reliable estimate.

\subsection{Embeddings of language models}

\textbf{Setups}. We further consider a real-world dataset in natural language processing. It consists of pairs of embeddings from a language model (LM)~\cite{devlin2018bert, dubey2024llama} computed on the IMDB dataset~\cite{maas-EtAl:2011:ACL-HLT2011}, which contains negative or positive movie comments; see Table 2. The ground truth MI of this dataset is unknown, but it can be computed numerically accurately; see Appendix B1. A total number of $n=4 \times 10^3$ data are used. Similar to the previous task, we preprocess data by an autoencoder $e: \R^{d_{\text{LM}}} \to \R^{16}$, with $d_{\text{LM}}$ being the dimensionality of the LM's embeddings. The quality of such compression is empirically studied in Appendix B2, which is near-lossless.


\textbf{Results}. Table~\ref{tab:nlp-dataset-results} summarizes the results for this dataset. In this scenario, where the underlying mutual information (MI) is relatively low, our method does not show a significant advantage over discriminative methods. This is likely because for this dataset, the high-discrepancy issue~\cite{rhodes2020telescoping, srivastava2023estimating, mcallester2020formal, song2019understanding} is not significant, and discriminative methods offer a more flexible parameterization of the vector copula density $c$ than our method (see Section 5). Nonetheless, our method still provides an estimate close to discriminative methods, and it significantly outperforms the generative method $\mathcal{N}$-MIENF.


\begin{table*}
  \centering
\label{tab:nlp-dataset-examples}
\begin{tabularx}{\textwidth}{lXX}
    \hline
     & \qquad\qquad\qquad\qquad\qquad \textbf{$X$} & \qquad\qquad\qquad\qquad\qquad \textbf{$Y$} \\
      \hline
\vspace{-0.25cm} \tt \small 1  &
\begin{spacing}{0.44}
\tt \small ({\color{green} positive}) I thought this was a wonderful way to spend time on ...
\end{spacing}
&
\begin{spacing}{0.44}
\tt \small ({\color{green} positive}) If you like original gut wrenching laughter you will like ...
\end{spacing}
\\
    \hline
\vspace{-0.25cm} \tt \small 2  &
\begin{spacing}{0.44}
\tt \small ({\color{red} negative}) So im not a big fan of Boll's work but then ...
\end{spacing}
&
\begin{spacing}{0.44}
\tt \small ({\color{green} positive}) This a fantastic movie of three prisoners who become famous...
\end{spacing}
\\
\hline
  \end{tabularx}
\caption{The text benchmark, which contains reviews of positive or negative movie comments. }
\vspace{-2mm}
\end{table*}

\begin{table*}[t]
\centering
\begin{minipage}{0.48\textwidth}
\centering
\resizebox{\textwidth}{!}{
\begin{tabular}{l p{8pt} c p{6pt} c}
\toprule
\textbf{Method} & & $I(X;Y)\approx2.1$ & & $I(X;Y)\approx0.9$ \\
\midrule
MINE & & 1.83 $\pm$ 0.04 & & {0.71} $\pm$ {0.05} \\
InfoNCE & & 1.64 $\pm$ 0.09 & & {0.70} $\pm$ {0.06} \\
MRE & & 1.72 $\pm$ 0.07 & & 1.23 $\pm$ 0.02 \\
$\mathcal{N}$-MIENF & & 0.91 $\pm$ 0.05 & & 0.43 $\pm$ 0.03 \\
\midrule
VCE & & \textbf{2.01} $\pm$ \textbf{0.04} & & \textbf{0.83} $\pm$ \textbf{0.01} \\
\bottomrule
\end{tabular}
}
\caption*{\textbf{(a) Llama-3 13B}}
\end{minipage}
\hfill
\begin{minipage}{0.48\textwidth}
\centering
\resizebox{\textwidth}{!}{
\begin{tabular}{l p{8pt} c p{6pt} c}
\toprule
\textbf{Method} & & $I(X;Y)\approx1.5$ & & $I(X;Y)\approx0.2$ \\
\midrule
MINE & & \textbf{1.42} $\pm$ \textbf{0.04} & & {0.18} $\pm$ {0.02} \\
InfoNCE & & 1.41 $\pm$ 0.03 & & {0.19} $\pm$ {0.04} \\
MRE & & 1.23 $\pm$ 0.09 & & 0.31 $\pm$ 0.09 \\
$\mathcal{N}$-MIENF & & 0.73 $\pm$ 0.03 & & 0.11 $\pm$ 0.02 \\
\midrule
VCE & & 1.22 $\pm$ 0.02 & & \textbf{0.19} $\pm$ \textbf{0.02} \\
\bottomrule
\end{tabular}
}
\caption*{\textbf{(b) BERT}}
\end{minipage}
\caption[Comparing different MI estimators on the text dataset. ]{Comparing different MI estimators on the text dataset. Left: evaluation on the embeddings of Llama-3 13B model~\cite{dubey2024llama}. Right: evaluation on the embeddings of a BERT model~\cite{devlin2018bert}.  }
\label{tab:nlp-dataset-results}
\vspace{-2mm}
\end{table*}

\subsection{Further analysis and ablation studies}
We conduct further analysis on the effect of \emph{model selection} and \emph{separate learning} in Appendix B2.

\section{Conclusion}


In this work, we introduced a new mutual information (MI) estimator grounded in recent vector copula theory. A fundamental difference to existing approaches is the explicit disentanglement of marginal distributions and dependence structure in our method. This separation enables more flexible and fine-grained modeling, avoiding the pitfalls of both overly simplistic or excessively complex approaches, and reducing overall learning difficulty via strategic factorization of the original estimation problem. Extensive experiments demonstrate our method's effectiveness and robustness.


  

Beyond the development of practical estimator, our research also offers fresh perspectives on MI estimation. By viewing PMI as a density rather than a density ratio, we open new avenues for modeling. Additionally, our approach to vector rank computation generalizes the classical copula transformation and holds promise as a versatile preprocessing step for a broad range of MI estimators. Finally, by reinterpreting existing estimators through the lens of vector copula theory, we obtain new insights into the parameterization and learning of different estimators and the underlying trade-offs.

Copulas have been widely used for MI estimate~\cite{samo2021inductive, safaai2018information, keziou2016semiparametric, zeng2018jackknife, letizia2025copula, singh2017nonparanormal, huk2025your, ma2011mutual, geenens2023towards}. Existing methods primarily focus on \emph{classic copulas}, where the copula transformation is applied independently to each univariate marginal to better account for the marginal-invariant property of MI. This strategy has been shown to improve accuracy and reduce variance~\cite{zeng2018jackknife, letizia2025copula}. We go one step further by using \emph{vector copulas}, where the transformation jointly considers all dimensions of the multivariate marginals. This can be seen as a generalization of classic copula transformation, where we not only consider MI's invariance to \emph{element-wise} bijections but also to \emph{any}  diffeomorphisms. Another key difference lies in that these works still treat PMI as a density ratio, whereas our work treats PMI as a density.

We note that, while powerful, our estimator is not a panacea. One limitation of our method is that it relies on the two marginal distributions to be reasonably modeled. While marginal distributions are far easier to learn than the joint distribution, they can still be challenging to learn for high-dimensional data e.g., images. Fortunately, dimensionality reduction techniques~\cite{gowri2024approximating, chen2020neural} help to mitigate this issue. Another limitation lies in the flexibility of our model-based parameterization of vector copula, which can be less flexible than neural network methods. However, as our method strikes a good trade-off between complexity and capacity across diverse cases, we consider it as a highly competitive method.


\section*{Acknowledgments}
AW acknowledges supports from a Turing AI Fellowship under grant EP/V025279/1, the Alan Turing Institute, and the Leverhulme Trust via the Leverhulme Centre for the Future of Intelligence. MUG was supported in part by Gen AI - the AI Hub for Generative Models,
funded by EPSRC (EP/Y028805/1). YC acknowledge  supports from the Cambridge Trust and the Qualcomm Innovation Fellowship. 
ZO acknowledges supports from Lee Family Scholarship via Imperial College London.

\clearpage

\bibliography{reference}
\bibliographystyle{unsrt}

\clearpage

\section*{A. Theoretical derivations}

\subsection*{A0. Proof of Theorem 2}
\setcounter{Theorem}{1}
\begin{Theorem}[MI is vector copula entropy]
The mutual information $I(X; Y)$ is the negative differential entropy of the vector copula density: 
\begin{equation}
    I(X; Y) = -H[c(\vecu_X, \vecu_Y)] 
\end{equation}
where $\vecu_X$ and $\vecu_Y$ are the vector ranks corresponding to $p(\vecx)$ and $p(\vecy)$ respectively.
\end{Theorem}
\noindent \emph{Proof}: 
The proof itself relies on the following lemma.

\begin{Lemma}[Equivalence between $p(\vecu_X, \vecu_Y)$ and $c(\vecu_X, \vecu_Y)$]
The vector copula density $c(\vecu_X, \vecu_Y)$ equals to the probabilistic density function $p(\vecu_X, \vecu_Y)$ of the vector ranks $\vecu_X$, $\vecu_Y$.
\label{lemma:p-c-equivalence}
\end{Lemma}
\noindent \emph{Proof of lemma}. According to the definition of vector ranks, we have the following two identities:
\[
    p(\vecu_X) = |J_{\vecx} \vecu_X|^{-1} p(\vecx) = 1, \qquad p(\vecu_Y) = |J_{\vecy} \vecu_Y|^{-1} p(\vecy) = 1
\]
where the first equality comes from the law of variable transformation and the second equality comes from the fact that $p(\vecu_X) = \mathcal{U}(0, 1)^{d_X}$ and $p(\vecu_Y) = \mathcal{U}(0, 1)^{d_Y}$ i.e. they are both factorized uniform distributions. Applying the the law of variable transformation again  and rearranging terms, we have
\[
    p(\vecu_X, \vecu_Y) = |J_{\vecx} \vecu_X|^{-1} |J_{\vecy} \vecu_Y|^{-1} p(\vecx, \vecy) = |J_{\vecx} \vecu_X|^{-1} |J_{\vecy} \vecu_Y|^{-1} p(\vecx)p(\vecy) c(\vecu_X, \vecu_Y) = c(\vecu_X, \vecu_Y)
    \label{eq:thm1-B}
\]
which completes the proof. \qed

Now let us turn to the proof of the theorem itself. Due to the bijectivity of vector rank functions (see Definition 1 in the main text), we have
\begin{equation}
    I(X; Y) =  I(\vecu_X; \vecu_Y) = H(\vecu_X) + H(\vecu_Y) - H(\vecu_X, \vecu_Y)
\label{eq:MI-latent}
\end{equation}
where $H(\vecu_X, \vecu_Y) = H[p(\vecu_X, \vecu_Y)]$ is the entropy of the joint distribution $p(\vecu_X, \vecu_Y)$ of the vector ranks $\vecu_X, \vecu_Y$. The first equality comes from the fact that MI is preserved under diffeomorphic maps $f, g$ i.e. $I(X; Y) = I(f(X); g(Y))$, so that $I(X; Y) = I(\vecu_X; \vecu_Y)$. 

Consider the terms in \eqref{eq:MI-latent}:

\begin{itemize}[leftmargin=*]
    \item For $H(\vecu_X)$ and $H(\vecu_Y)$, we have $H(\vecu_X) = H(\vecu_Y) = 0$ since $p(\vecu_X) = \mathcal{U}(0, 1)^{d_X}$ and $p(\vecu_Y) = \mathcal{U}(0, 1)^{d_Y}$;
    \item For $H(\vecu_X, \vecu_Y)$, we have $(H[p(\vecu_X, \vecu_Y)] = H[c(\vecu_X, \vecu_Y)]$ due to Lemma \ref{lemma:p-c-equivalence}.
\end{itemize}

Combined, we have $I(X; Y) = H(\vecu_X) + H(\vecu_Y) - H(\vecu_X, \vecu_Y) = 0 + 0 - H[c(\vecu_X, \vecu_Y)]$, which completes the proof. \qed

\setcounter{Proposition}{0}

\subsection*{A1. Proof of Proposition 1}
\label{subsection:ch3-consistency}

\begin{Proposition}[Consistency of VCE]
Assuming that (a) the flows $f_X$ and $f_Y$ are universal PDF approximator with continuous support and (b) the number of mixture components $K$ is sufficiently large. Define $\hat{I}_n(X; Y) \coloneqq \frac{1}{n} \sum_{i=1}^n \log \hat{c}(\hat{\vecu}_X^{(i)}, \hat{\vecu}_Y^{(i)})$. For every $\epsilon > 0$, there exists $n(\varepsilon) \in \mathbb N$, such that 
\[
     \left|\hat{I}_n(X; Y) - I(X; Y)\right| < \varepsilon, \quad \forall n \geq n(\varepsilon), a.s.
\]
\end{Proposition}

\noindent \emph{Proof}. The proof relies on the following lemma. \\

\begin{Lemma}[Consistency of Nested Argmax Estimators] \label{lemma:consistent-argmax}
    Let $\hat{\theta}_1$ be a consistent estimator of $\theta_1^*$, and $\hat{\theta}_2$ is a consistent estimator of $\argmax_{\theta_2} f(\theta_2, \hat{\theta}_1)$. Assume that $f(\theta_1, \theta_2)$ is continuous in both $\theta_2$ and $\theta_1$, and that the maximizer $\argmax_{\theta_2} f(\theta_2, \theta_1)$ is unique for any $\theta_1$. Then $\hat{\theta}_2$ is also a consistent estimator of $\argmax_{\theta_2} f(\theta_2, \theta_1^*)$.
\end{Lemma}

\noindent \emph{Proof of lemma}. 
Given the consistency of $\hat{\theta}_1$, we have $\hat{\theta}_1 \xrightarrow{P} \theta^*$. By the continuous mapping theorem \cite{mann1943stochastic} and the continuity of $f$, it follows that 
\begin{align}
    f(\theta_2, \hat{\theta}_1) \xrightarrow{P} f(\theta_2, {\theta}^*) \quad \text{for any fixed}~\theta_2, \nonumber
\end{align}
which implies that the function $f(\theta_2, \hat{\theta}_1)$ converges pointwise to $f(\theta_2, {\theta}^*)$. Then, by the uniform convergence theorem for maximizers~\cite{rudin1974functional}, we have
\begin{align}
    \hat{\theta}_2 = \argmax_{\theta_2} f(\theta_2, \hat{\theta}_1) \xrightarrow{P} \argmax_{\theta_2} f(\theta_2, {\theta}^*) = \theta_2^*, \nonumber
\end{align}
which completes the proof.
\qed \\

Given the above lemma, we now prove the proposition itself. The complete proof of the proposition consists of four steps: \\

\noindent \emph{(a). Estimation of $\vecu_X$, $\vecu_Y$ is consistent}. Under the assumption that $f_x$ and $f_y$ are universal PDF approximator with continuous supports, they converge to the true marginal distributions in the limit of infinite data. Consequently, the estimated vector ranks $\hat{\vecu}_X$ and $\hat{\vecu}_Y$ converge in probability to the true vector ranks ${\vecu}_X$ and ${\vecu}_Y$, respectively. That is, $\hat{\vecu}_X \xrightarrow{P} {\vecu}_X$ and $\hat{\vecu}_Y \xrightarrow{P} {\vecu}_Y$. \\

\noindent \emph{(b). Estimation of $c$ is consistent given ground truth $\vecu_X, \vecu_Y$}. By the universal approximation theorem of mixtures \cite{hangelbroek2010nonlinear} and the consistency of maximum likelihood estimator \cite{van2000asymptotic}, the estimator 
\begin{align}
    \argmax_{c} \frac{1}{m} \sum_{j=1}^m \log c({\vecu}_X, {\vecu}_Y), \qquad {\vecu}_X, {\vecu}_Y \sim p(\vecu_X, {\vecu}_Y), \nonumber
\end{align}
is a consistent estimator of the true copula density $c^*$. Here $p(\vecu_X, {\vecu}_Y)$ is the true distribution of vector ranks. \\


\noindent \emph{(c). Estimation of $c$ is consistent in two-phrase learning}. Combining the results (a)(b), above, by Lemma \ref{lemma:consistent-argmax}, the estimator 
\begin{align}
    \hat{c} = \argmax_{c} \frac{1}{m} \sum_{j=1}^m \log c(\hat{\vecu}_X, \hat{\vecu}_Y),  \hat{\vecu}_X, \hat{\vecu}_Y \sim \hat{p}(\hat{\vecu}_X, \hat{\vecu}_Y), \nonumber
\end{align}
is also consistent. Here $\hat{p}$ is the distribution induced by the learned flows. \\

\noindent \emph{(d). Estimation of MI is consistent.} Given the above results, we now show that our estimator is consistent. We begin by defining the following terms:
\[
     \hat{I}_n(X; Y) \coloneqq \frac{1}{n} \sum_{i=1}^n \log \hat{c}(\hat{\vecu}_X^{(i)}, \hat{\vecu}_Y^{(i)}),
\]
\[
    I'_n(X; Y) \coloneqq \frac{1}{n} \sum_{i=1}^n \log {c}^*(\hat{\vecu}_X^{(i)}, \hat{\vecu}_Y^{(i)}),
\]
\[
    I''_n(X;Y) \coloneqq \frac{1}{n} \sum_{i=1}^n \log {c}^*(\vecu_X^{(i)}, \vecu_Y^{(i)}),
\]
where $c^*$ is the true vector copula and $\vecu_X, \vecu_Y$ are the true vector ranks. Note that $I(X; Y) = \mathbb{E}[\log c^*(\vecu_X, \vecu_Y)]$, which is the limit of $I''_n(X; Y)$ as $n \to \infty$. 

By triangle inequality,
\begin{equation}
    \left|I(X; Y) - \hat{I}_n(X; Y) \right| \leq \underbrace{\left|\hat{I}_n(X; Y) - I'_n(X; Y) \right|}_{\triangle} + \underbrace{\Big|I'_n(X; Y) - I''_n(X; Y) \Big|}_{\nabla} + \Big|I''_n(X; Y) - I(X; Y) \Big|
    \label{proof:triangle-inequality}
\end{equation}
(i) Since the estimator $\hat{c}$ is consistent, we know that  for every $ \varepsilon > 0$, there exists a sufficiently large $n \in \mathbb{N}$, such that $|\log \hat{c}(\hat{\vecu}_X^{(i)}, \hat{\vecu}_Y^{(i)}) - \log c^*(\hat{\vecu}_X^{(i)}, \hat{\vecu}_Y^{(i)})| < \epsilon, \forall \hat{\vecu}_X^{(i)}, \hat{\vecu}_Y^{(i)}, a.s.$ \ . Then for the first term in the RHS of \eqref{proof:triangle-inequality}, we have
\begin{align}
    \triangle = \frac{1}{n} \left|  \sum_{i=1}^n \log \hat{c}(\hat{\vecu}_X^{(i)}, \hat{\vecu}_Y^{(i)}) \!-\!  \sum_{i=1}^n \log {c}^*(\hat{\vecu}_X^{(i)}, \hat{\vecu}_Y^{(i)}) \right| 
    &\leq \frac{1}{n}  \sum^n_{i=1} \left|  \log \hat{c}(\hat{\vecu}_X^{(i)}, \hat{\vecu}_Y^{(i)}) - \log {c}^*(\hat{\vecu}_X^{(i)}, \hat{\vecu}_Y^{(i)}) \right| \\ \nonumber 
    &= \epsilon 
\end{align}
(ii) Since the estimators $\hat{\vecu}_X, \hat{\vecu}_Y$ are consistent, we know that for every $ \varepsilon > 0$, there exists a sufficiently large $n \in \mathbb{N}$, such that $|\log {c}^*(\hat{\vecu}_X^{(i)}, \hat{\vecu}_Y^{(i)}) - \log c^*(\vecu_X^{(i)}, \vecu_Y^{(i)})| < \epsilon, \forall i, a.s.$ \ . Then for the second term in the RHS of \eqref{proof:triangle-inequality}, we have
\begin{align}
    \nabla = \frac{1}{n}  \left| \sum_{i=1}^n \log {c}^*(\hat{\vecu}_X^{(i)}, \hat{\vecu}_Y^{(i)}) \!-\!  \sum_{i=1}^n \log {c}^*({\vecu}_X^{(i)}, {\vecu}_Y^{(i)}) \right| 
    &\leq \frac{1}{n}  \sum^n_{i=1} \left|  \log {c}^*(\hat{\vecu}_X^{(i)}, \hat{\vecu}_Y^{(i)}) - \log {c}^*({\vecu}_X^{(i)}, {\vecu}_Y^{(i)}) \right| \\ \nonumber
    &= \epsilon  
\end{align}
(iii) For the third term, it vanishes given large $n$ due to the normal strong law of large numbers under mild conditions. \\

Given (i)(ii)(iii) and \eqref{proof:triangle-inequality}, it follows that for every $\epsilon > 0$, there exist $n(\varepsilon) \in \N$, such that $\left|\hat{I}_n(X; Y) - I(X; Y)\right| < \varepsilon, \forall n \geq n(\varepsilon), a.s.$ \qed \\

\subsection*{A2. Proof of Proposition 2}
\label{subsection:ch3-error-analysis}

\begin{Proposition}[Error of vector copula-based MI estimate]
Let $\hat{\vecu}_X$ and $\hat{\vecu}_Y$ be the estimated vector ranks. Let $p(\hat{\vecu}_X, \hat{\vecu}_Y)$ and $\hat{p}(\hat{\vecu}_X, \hat{\vecu}_Y)$ be the true and the estimated joint distributions of $\hat{\vecu}_X$ and $\hat{\vecu}_Y$ respectively\footnote{Note that in this case, $p(\hat{\vecu}_X, \hat{\vecu}_Y)$ is not the true vector copula density unless $\hat{\vecu}_X = \vecu_X$ and $\hat{\vecu}_Y = \vecu_Y$. }. Assuming that sufficient Monte Carlo samples are used to compute $\hat{I}(X; Y)$ in eq. (5) in the main text, we have
\begin{equation}
    \Big | I(X; Y) - \hat{I}(X; Y) \Big| \leq \Big| H(\hat{\vecu}_X) + H(\hat{\vecu}_Y) \Big | +  KL [p(\hat{\vecu}_X, \hat{\vecu}_Y) \| \hat{p}(\hat{\vecu}_X, \hat{\vecu}_Y) ]  
    \label{formula:eq:error-main}
\end{equation}
where the first term on the RHS vanishes as $\hat{p}(\vecx) \to p(\vecx)$ and $\hat{p}(\vecy) \to p(\vecy)$. In the limit of perfectly learned marginals, we have
\begin{equation}
\Big | I(X; Y) - \hat{I}(X; Y) \Big| = KL[c \| \hat{c}]
    \label{formula:eq:error-special}
\end{equation}
where $c$ and $\hat{c}$ are the true and estimated vector copula densities respectively.
\end{Proposition}

\noindent \emph{Proof}. 
The proof begins with the following two facts:
\begin{itemize}[leftmargin=*]
    \item On one hand, due to the bijectivity of flow-based models, we have $I(X; Y) = I(\hat{\vecu}_X; \hat{\vecu}_Y)$. Then
    \[
    I(X; Y) = H(\hat{\vecu}_X) + H(\hat{\vecu}_Y) - H(\hat{\vecu}_X, \hat{\vecu}_Y) = H(\hat{\vecu}_X) + H(\hat{\vecu}_Y) - \mathbb{E}_{p}[-\log p(\hat{\vecu}_X, \hat{\vecu}_Y)].
    \]
    \item On the other hand, as $n \to \infty$, we have that by construction, \[
    \hat{I}(X; Y) = \mathbb{E}_{p}[-\log \hat{p}(\hat{\vecu}_X, \hat{\vecu}_Y) ].
    \]
\end{itemize}
These combined results lead to the following identify:
\begin{equation}
    I(X; Y) - \hat{I}(X; Y) = H(\hat{\vecu}_X) + H(\hat{\vecu}_Y) - \Big(\mathbb{E}_p[-\log p(\hat{\vecu}_X, \hat{\vecu}_Y)]- \mathbb{E}_p[-\log \hat{p}(\hat{\vecu}_X, \hat{\vecu}_Y) ]  \Big)
\end{equation}
which can be rewritten as
\begin{equation}
    I(X; Y) - \hat{I}(X; Y) =  H(\hat{\vecu}_X) + H(\hat{\vecu}_Y)  + KL [p(\hat{\vecu}_X, \hat{\vecu}_Y)) \| \hat{p}(\hat{\vecu}_X, \hat{\vecu}_Y)]  
\end{equation}
By applying triangular inequality, we have
\begin{equation}
    \Big | I(X; Y) - \hat{I}(X; Y) \Big| \leq \Big| H(\hat{\vecu}_X) + H(\hat{\vecu}_Y) \Big | + KL [p(\hat{\vecu}_X, \hat{\vecu}_Y)) \| \hat{p}(\hat{\vecu}_X, \hat{\vecu}_Y)]
    \label{formula:error-full}
\end{equation}
which completes the first part of the proof.

Now we turn to the second part of the proof. In the limit of perfectly learned marginals, we have $\hat{p}(\vecx) = p(\vecx)$ and $\hat{p}(\vecy) = p(\vecy)$. This yields
\[
\hat{\vecu}_X = \hat{P}(\vecx) = P(\vecx) = \vecu_X, \qquad \hat{\vecu}_Y = \hat{P}(\vecy) = P(\vecy) = \vecu_Y
\]
Since $\vecu_X \sim \mathcal{U}[0, 1]^{d_X}$ and $\vecu_Y \sim \mathcal{U}[0, 1]^{d_Y}$, we have
\[
    H(\vecu_X) = H(\vecu_Y) = 0.
\]
Therefore the first term on the RHS in \eqref{formula:error-full} vanishes. 

For the second term on the RHS in \eqref{formula:error-full}, since $\hat{\vecu}_X = \vecu_X$ and $\hat{\vecu_Y} = \vecu_Y$, we have
\vspace{0.1cm}
\[
     KL [p({\hat{\vecu}}_X, {\hat{\vecu}}_Y)) \| \hat{p}(\hat{\vecu}_X, \hat{\vecu}_Y)] =  KL [p({\vecu}_X, {\vecu}_Y)) \| \hat{p}(\vecu_X, \vecu_Y)] = KL [c({\vecu}_X, {\vecu}_Y)) \| \hat{c}(\vecu_X, \vecu_Y)]
\]
where the last equality comes from Lemma~\ref{lemma:p-c-equivalence}, which states that $p(\vecu_X, \vecu_Y) = c(\vecu_X, \vecu_Y)$.

Substituting both terms to \eqref{formula:error-full}, we have $\Big | I(X; Y) - \hat{I}(X; Y) \Big| = 0 + 0 - KL[c\|\hat{c}] = KL[c\|\hat{c}]$. \qed

\subsection*{A3. Proof of Proposition 3}
\label{subsection:ch3-power-of-vgc}

\begin{Proposition}[Vector Gaussian copula as second-order approximation]
A vector Gaussian copula $c^{\mathcal{N}}$ corresponds to the second-order Taylor expansion of the true vector copula $c^*$ up to variable transformation. 
\end{Proposition}

\noindent \emph{Proof}.
Denote $\vecu  = [\vecu_X, \vecu_Y]$ and $\vecz = \phi^{-1}(\vecu)$ where $\phi(\cdot)$ is the element-wise CDF of Gaussian distribution. Let $p(\vecz)$ be the distribution of $\vecz$ and let $\mu$ be the mode of this distribution. We have
\begin{align}
    \log c^*(\vecu) &= \log | J_{\vecz}\vecu|^{-1} + \log p(\vecz) 
\end{align}
Applying a second-order Taylor expansion of $\log p(\vecz)$ around the mode $\mu$, we get
\[
     \log c^*(\vecu) \approx \log | J_{\vecz}\vecu|^{-1} + \log p(\mu) + \mathbf{g}^{\top}(\vecz - \mu) + \frac{1}{2}(\vecz - \mu)^{\top}\mathbf{H}(\vecz - \mu) 
\]
where $\mathbf{g}$ and  $\mathbf{H}$ is the gradient and the Hessian of $p(\vecz)$ at $\mu$. Since $\mu$ is the mode, we have $\mathbf{g} = \mathbf{0}$. Therefore
\[
    \log c^*(\vecu) \approx \log | J_{\vecz}\vecu|^{-1} + \underbrace{\log p(\mu) + \frac{1}{2}(\vecz - \mu)^{\top}\mathbf{H}(\vecz - \mu)}_{h(\vecz)}  
\]
Now consider normalizing this unnormalized (log) density by defining a proper density $q(\vecz)  = h(\vecz)/\int h(\vecz) d\vecz$. Given the quadratic form of $h(\vecz)$, its corresponding normalized density $q(\vecz)$ must be a Gaussian distribution with certain mean $\mu$ and covariance $\Sigma$. Then
\begin{align}
    \log c^*(\vecu) 
    \approx \log | J_{\vecz}\vecu|^{-1} + \log \mathcal{N}(\vecz; \mu, \Sigma) = \log c^{\mathcal{N}}(\vecu; \mu, \Sigma)
\end{align}
Note that RHS itself is a valid probabilistic density function. This shows that the vector Gaussian copula corresponds to the second-order Taylor approximation of the true  vector copula in a transformed space induced by CDF of (univariate) standard normal distribution: $\phi: \R \to (0, 1)$. \qed \\

\subsection*{A4. Proof of Proposition 4}

\begin{Proposition}[Vector copula of the product of marginals]
The copula of the distribution $p'(\vecx, \vecy) = p(\vecx)p(\vecy)$ is a vector Gaussian copula if $p'(\vecx, \vecy)$ is absolutely continuous.
\end{Proposition}
\noindent \emph{Proof}. The proof of the proposition relies on the following lemma. \\

\begin{Lemma}[Equivalent representation of vector Gaussian copula]
Let $f, g$ be two bijective functions. Consider the following data generation process for random variables $X \in \R^{d_X}$ and $Y \in \R^{d_Y}$:
\[
    \vecx = f(\epsilon_{\leq d_X}), \qquad \vecy = g(\epsilon_{> d_X}),
\]
\[
    \epsilon \sim \mathcal{N}(\epsilon; 0, \Sigma)
\]
where $\epsilon \in \R^{d_X + d_Y}$. $\epsilon_{\leq d_X}$ denotes the first $d_X$ dimensions of $\epsilon$ and $\epsilon_{> d_X}$ denotes the last $d_Y$ dimensions of $\epsilon$, and $\mathcal{N}(\epsilon; 0, \Sigma)$ is a Gaussian distribution with zero mean and covariance $\Sigma$.
Then the vector copula of the distribution $p(\vecx, \vecy)$ corresponding to the above generation process is a vector Gaussian copula.
\label{lemma:vgc}
\end{Lemma}
\noindent \emph{Proof of lemma}: Let $f', g'$ be certain bijective functions. The above data generating process can be equivalently expressed as follows:
\[
    \vecx = f'(\epsilon'_{\leq d_X}), \qquad \vecy = g'(\epsilon'_{> d_X}),
\]
\[
    \epsilon' \sim \mathcal{N}(\epsilon'; 0, \Sigma')
\]
where $\Sigma' = \begin{bmatrix}
\mathbf{I}_X & \Sigma'_{XY} \\
 \Sigma'^{\top}_{XY} & \mathbf{I}_Y 
\end{bmatrix}$ is a p.s.d matrix whose blocks $\mathbf{I}_X \in \R^{d_X \times d_X}$ and $\mathbf{I}_Y \in \R^{d_Y \times d_Y}$ are two identity matrices. 

Consider $\vecu_X = \phi(\epsilon'_{\leq d_X})$ and $\vecu_Y = \phi(\epsilon'_{>d_X})$, where $\phi$ is the element-wise cumulative distribution function (CDF) of univariate normal distribution. Since different dimensions $\vecu_X$ are  independent (as dimensions in $\epsilon'_{\leq d_X}$ are independent), and that each dimension in $\vecu_X \sim \mathcal{U}[0, 1]$, $\vecu_X \sim \mathcal{U}[0, 1]^{d_X}$ and thereby is the vector rank corresponding to $p(\vecx)$. Similarly, $\vecu_Y$ is also the vector rank corresponding to $p(\vecy)$. In summary,  $\vecu_X$ and $\vecu_Y$ are the vector ranks corresponding to  $p(\vecx)$ and  $p(\vecy)$ respectively. 

Now consider the joint CDF $P(\vecu_X, \vecu_Y)$ of the random variables $\vecu_X$ and $\vecu_Y$:
\[
    P(\vecu_X, \vecu_Y) = P(\epsilon'_{\leq d_X}, \epsilon'_{> d_X}) = \Phi(\epsilon'_{\leq d_X}, \epsilon'_{> d_X}, \Sigma') = \Phi(\phi^{-1}(\vecu_X), \phi^{-1}(\vecu_Y), \Sigma')
\]
where $\Phi$ is the CDF of multivariate normal distribution. Comparing the RHS of the equation and the definition of vector Gaussian copula, one can see that $P(\vecu_X, \vecu_Y)$ satisfies the definition of vector Gaussian copula. \qed \\

Given the above lemma, we now turn to the proof of the proposition itself. Literature~\cite{papamakarios2021normalizing} shows that for any absolutely continuous distribution $p(\vecx)$, there exists a diffeomorphism that turns a Gaussian distribution into $p(\vecx)$. Then there exist two diffeomorphisms $f, g$ such that
\[
    \vecx \sim p(\vecx) \Leftrightarrow \vecx = f(\epsilon_X), \enskip \epsilon_X \sim \mathcal{N}(\epsilon_X; 0, \mathbf{I}), \qquad     \vecy \sim p(\vecy) \Leftrightarrow \vecy = g(\epsilon_Y), \enskip \epsilon_Y \sim \mathcal{N}(\epsilon_Y; 0, \mathbf{I})
\]
Since $\vecx \perp \vecy$, we have that 
\[
    I(X; Y) = 0 \Rightarrow I(\epsilon_X; \epsilon_Y) = 0
\]
Therefore $\epsilon_X \perp \epsilon_Y$. Then
\[
    p(\epsilon_X, \epsilon_Y) = p(\epsilon_X) p(\epsilon_Y) = \mathcal{N}(\epsilon_X; 0, \mathbf{I})\mathcal{N}(\epsilon_Y; 0, \mathbf{I}) = \mathcal{N}(\epsilon; 0, \mathbf{I})
\]
where $\epsilon = [\epsilon_X, \epsilon_Y]$ is a random variable whose first $d_X$ dimensions is $\epsilon_X$ and the last $d_Y$ dimensions is $\epsilon_Y$. 

This implies that data $\vecx, \vecy \sim p(\vecx)p(\vecy)$  can be equivalently expressed by the following data generation process:
\[
    \vecx = f(\epsilon_X), \qquad \vecy = g(\epsilon_Y),
\]
\[
    \epsilon \sim \mathcal{N}(\epsilon; 0, \mathbf{I})
\]
whose vector copula, according to Lemma~\ref{lemma:vgc}, is a vector Gaussian copula. \qed

\subsection*{A5. Error bound in MI estimation with lossy compression}

\begin{Proposition}[Error bound in MI estimation with lossy compression]
\label{prop:lossy-compression}
Let $X, Y \in \mathbb{R}^D$ be random variables with a joint distribution $p(\vecx, \vecy)$ that is absolutely continuous with respect to the Lebesgue measure. Let $e: \mathbb{R}^D \to \mathbb{R}^d$ be an encoder and $h: \mathbb{R}^d \to \mathbb{R}^D$ be a decoder, both deterministic mappings. Suppose that the conditional log-densities $\log p(\vecy \mid \vecx)$ and $\log p(\vecy \mid \vecx)$ are differentiable w.r.t $\vecx$ and $\vecy$ respectively, and their gradient are uniformly bounded:
\[
\| \nabla_{\vecx} \log p(\vecy \mid \vecx) \| \| \leq L, \quad \text{and} \quad \| \nabla_{\vecy} \log p(\vecx \mid \vecy) \| \leq L \quad \forall \vecx, \vecy.
\]
Assume the reconstruction error is uniformly bounded:
\[
\| h(e(\vecx)) - \vecx \|_2 \leq \xi \quad \text{and} \quad \| h(e(\vecy)) - \vecy \|_2 \leq \xi \quad \forall \vecx, \vecy.
\]
Then, as $\xi \to 0$, the mutual information under compression satisfies:
\[
\left| I(e(X); e(Y)) - I(X; Y) \right| = O(L \xi).
\]
\end{Proposition}

\noindent \emph{Proof}. We begin with the following lemma. \\

\begin{Lemma}[Local KL Stability under Uniformly Bounded Score]
\label{lemma:kl_stability_bounded_score}
Let $p(y \mid z)$ be a conditional probability density defined over $\mathcal{Y} \times \mathcal{Z} \subseteq \mathbb{R}^m \times \mathbb{R}^D$, and suppose:
\begin{itemize}[leftmargin=*]
    \item For all $(y, z) \in \mathcal{Y} \times \mathcal{Z}$, the mapping $z \mapsto \log p(y \mid z)$ is differentiable;
    \item The score function is uniformly bounded such that
    $\|\nabla_z \log p(y \mid z)\| \leq L, \forall y \in \mathcal{Y},\, z \in \mathcal{Z}$.
\end{itemize}
Then for any $z \in \mathcal{Z}$ and any perturbation vector $\varepsilon \in \mathbb{R}^d$ with $\|\varepsilon\| \to 0$, the KL divergence between nearby conditionals satisfies:
\[
\mathrm{KL}\left[ p(y \mid z) \,\|\, p(y \mid z + \varepsilon) \right] = O(L\|\varepsilon\|).
\]
\end{Lemma}

\noindent \emph{Proof of lemma}. We begin by the Taylor expansion of $\log p(y|z + \varepsilon)$ around $z$:
\[
\log p(y|z + \varepsilon)
= \log p(y \mid z) + \nabla_z \log p(y \mid z)^\top \varepsilon
+ \underbrace{r(y, \varepsilon)}_{o(\|\varepsilon\|^2)}
\]
where $r(y, \varepsilon)$ is the remainder. Since $\|\nabla_z \log p(y \mid z)\| \leq L$, we have
\[
    \Big |\log p(y|z + \varepsilon) - \log p(y|z) \Big | = L \|\varepsilon\| +  o(\|\epsilon\|)
\]

Now consider the KL divergence between the two conditional densities:
\[
\mathrm{KL}\left[ p(y \mid z) \,\|\, p(y \mid z + \varepsilon) \right]
= \mathbb{E}_{p(y \mid z)} \left[ \log \frac{p(y \mid z)}{p(y \mid z + \varepsilon)} \right] \leq \mathbb{E}\left[\Big |\log p(y|z) - \log p(y|z + \varepsilon) \Big |\right].
\]

Substituting the above Taylor expansion term into the KL divergence, we have
\[
    \mathrm{KL}\left[ p(y \mid z) \,\|\, p(y \mid z + \varepsilon) \right] \leq \mathbb{E}\left[\Big |\log p(y|z) - \log p(y|z + \varepsilon) \Big |\right] = L\|\varepsilon\| + o(\|\varepsilon \|) = O(L\|\varepsilon \|)
\]
which completes the proof of the lemma. \qed

To prove the theorem, we need another lemma.

\begin{Lemma}[One-side error bound in MI estimation with lossy compression]
\label{lemma:one-side-bounded-MI-compressed-estimate}
Let $X, Y \in \mathbb{R}^D$ be random variables with a joint distribution $p(\vecx, \vecy)$ that is absolutely continuous with respect to the Lebesgue measure. Let $e: \mathbb{R}^D \to \mathbb{R}^d$ be an encoder and $h: \mathbb{R}^d \to \mathbb{R}^D$ be a decoder, both deterministic mappings. Supposing that all conditions mentioned in Lemma A3 are met. Assume the reconstruction error is uniformly bounded:
\[
\| h(e(\vecx)) - \vecx \|_2 \leq \xi, \quad \forall \vecx
\]
Then, as $\xi \to 0$, the mutual information under compression satisfies:
\[
\left| I(e(X); Y) - I(X; Y) \right| = O(L \xi).
\]
\end{Lemma}

\noindent \emph{Proof of lemma}. Denote $F := h \circ e$ be the reconstruction map, and define the reconstruction residual $\varepsilon := F(\vecx) - \vecx$. By assumption, $\|\varepsilon\| \leq \xi$ for all $\vecx$.

We have 
\begin{align*}
        I(F(X); Y) - I(X; Y) &= \int p(\vecx, \vecy) \log \frac{p(\vecy|\vecx)}{p(\vecy)}d\vecx d\vecy - \int p(F(\vecx), \vecy) \log \frac{p(\vecy|F(\vecx))}{p(\vecy)} dF(\vecx) d\vecy \\
    &= \int p(\vecx, \vecy) \log p(\vecy|\vecx) d\vecx d\vecy - \int p(F(\vecx), \vecy) \log p(\vecy|F(\vecx)) dF(\vecx) d\vecy \\
    &= \int p(\vecx, \vecy) \log p(\vecy|\vecx) d\vecx d\vecy - \int p(\vecx, \vecy) \log p(\vecy|F(\vecx)) d\vecx d\vecy \\
    &= \int p(\vecx) \text{KL} \Big [ p(\vecy|\vecx) \| p(\vecy|F(\vecx)) \Big ] d\vecx \\
    &\leq \sup_{\vecx} \text{KL} \Big [ p(\vecy|\vecx) \| p(\vecy|F(\vecx)) \Big ] 
\end{align*}
By the KL stability lemma, under the assumption that $|\nabla_{\vecx} \log p(\vecy \mid \vecx)| \leq L, \forall \vecx$, we have
\[
    \Big | I(F(X); Y) - I(X; Y) \Big| = O(L \|\varepsilon \|) = O(L\xi)
\]
where in the last step we substitute $\|\varepsilon\| \leq \xi$. \qed \\

\noindent Given the above lemmas, we now turn to the proof of the proposition itself. 

By data process inequality, we have
\[
I(X; Y) \geq I(e(X); Y) \geq I(F(X); Y)
\]
Therefore
\[
 0 \leq I(X; Y) - I(e(X); Y)  \leq I(X; Y) - I(F(X); Y) 
\]
hence
\[
 \Big |I(X; Y) - I(e(X); Y) \Big|  \leq \Big |I(X; Y) - I(F(X); Y) \Big| = O(L\xi)
\]
A similar argument applies to the deviation $|I(e(X); e(Y)) - I(e(X); Y)|$, yielding
\[
     \Big |I(e(X); Y) - I(e(X); e(Y)) \Big| = O(L\xi)
\]
By triangular inequality, we have
\[
\Big | I(X; Y) - I(e(X); e(Y)) \Big| \leq \Big |I(X; Y) - I(e(X); Y) \Big| + \Big |I(e(X); Y) - I(e(X); e(Y)) \Big| = O(L\xi)
\]
which completes the proof. \qed

\clearpage
\section*{B. Experiment details and further results}

\subsection*{B1. Experiment details}

\paragraph{Neural network settings} 
For controlled experiment, we use the same generative model in $\mathcal{N}$-MIENF and our method, and use the same critic network for MINE, InfoNCE and MRE; see below for the details of the networks. All networks are trained by Adam~\cite{kingma2014adam} with its default settings, where the learning rate is set to be $5\times10^{-4}$ and the batch size is set to be $512$. Early stopping are applied to avoid overfitting in all network training. We use 80\% of the data for training and 20\% for validation. The detailed architectures of the neural networks used are as follows: 

\begin{itemize}[leftmargin=*]
    \item \emph{Flow models}. We implement the two flow models $f_X, f_Y$ in our method and $\mathcal{N}$-MIENF by a continuous flow model trained by flow matching~\cite{lipman2022flow}. This flow model is implemented as a 4-layer MLP with 1024 hidden units per each layer and softplus non-linearity. 
    \item \emph{Critic networks}. We implement the critic network $f$ in discrminative methods (MINE, MRE and InfoNCE) a MLP with 3 hidden layers,  each of which has 500 neurons. A densenet architecture \cite{huang2017densely} is used for the network, where we concatenate the input of the first layer (i.e., $x$ and $y$) and the representation of the penultimate layer before feeding them to the last layer. Leaky ReLU \cite{maas2013rectifier} is used as the activation function for all hidden layers. 
    \item \emph{Autoencoders}. For the autoencoder used in part of the experiments, we implement it as a 7-layer MLP with skip connection with architecture $d_{\text{input}} \to 512 \to 512 \to d_{\text{hidden}} \to 512 \to 512 \to d_{\text{input}}$, where $d_{\text{input}}$ and $d_{\text{hidden}}$ are dimensionalities of the input and the representation respectively.  
\end{itemize}

\paragraph{Resampling real-world dataset to generate dataset with known MI} We use a technique inspired by that in~\cite{gowri2024approximating} to turn a real-world dataset $\mathcal{D}$ with data $Z \in \R^{d}$ and ground truth labels $L \in \{1, 2, ..., K\}$ into a dataset $\mathcal{D}'$ with data $X \in \R^{d}$ and $Y \in \R^d$, where $I(X; Y)$ is known. The method is based on the assumption $H[L|Z] \approx 0$ i.e. given the data, there is no ambiguity about its label. This condition is well satisfied for the IMDB dataset~\cite{maas-EtAl:2011:ACL-HLT2011}, where positive and negative comments are well-distinguished~\cite{maas-EtAl:2011:ACL-HLT2011}. 

Specifically, to generate data, we first sample $X, Y, L_X, L_Y \sim p(X|L_X)p(Y|L_Y)p(L_X, L_Y)$ where $p(L_X, L_Y)$ is a user-defined joint distribution for the discrete random variables $L_X, L_Y$ and $p(X|L)$ and $p(Y|L)$ are the distributions of data within class $L$, respectively. It is shown in~\cite{gowri2024approximating} that under the assumption $H[L|X] \approx 0$ and $H[L|Y] \approx 0$, we have $I(X; Y) \approx I(L_X; L_Y)$. The latter is analytically known due to the availability of the discrete distributions $p(L_X, L_Y)$ and $p(L_X)p(L_Y)$.

\subsection*{B2. Further results and ablation studies}

\begin{figure*}[t!]
            \hspace{-0.02\linewidth}
            \begin{subfigure}{.205\textwidth}
                    \centering
                    \begin{minipage}[t]{1.0\linewidth}
                    \centering
                     \includegraphics[width=1.0\linewidth]{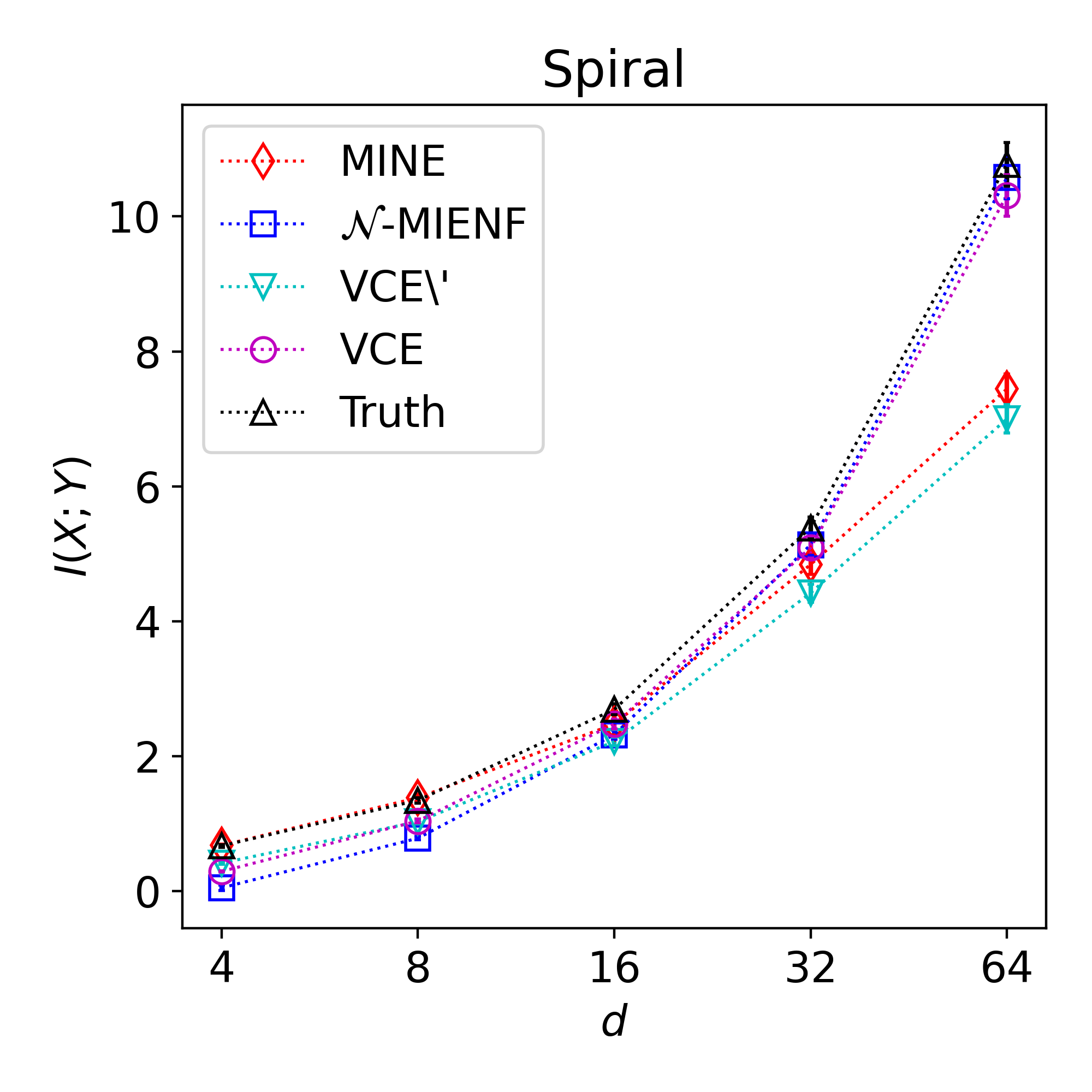}
                    \end{minipage}
            \subcaption{\centering Spiral}
            \end{subfigure}
            \hspace{-0.015\linewidth}
            \begin{subfigure}{.205\textwidth}
                    \centering
                    \begin{minipage}[t]{1.0\linewidth}
                    \centering
                     \includegraphics[width=1.0\linewidth]{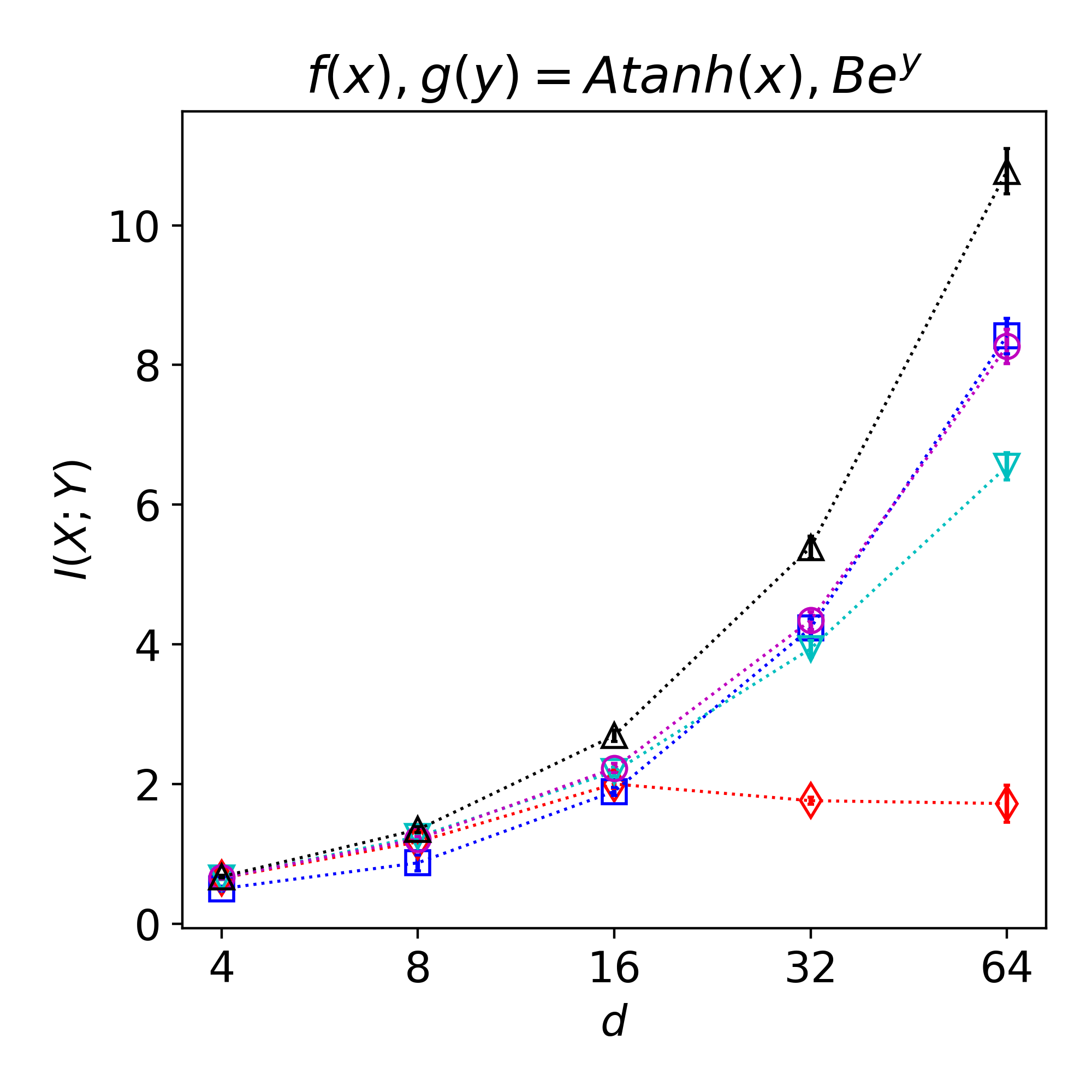}
                    \end{minipage}
            \subcaption{\centering $\textbf{A}\text{tanh}(X), \textbf{B}e^Y$}
            \end{subfigure}
            \hspace{-0.015\linewidth}
            \begin{subfigure}{.205\textwidth}
                    \centering
                    \begin{minipage}[t]{1.0\linewidth}
                    \centering
                     \includegraphics[width=1.0\linewidth]{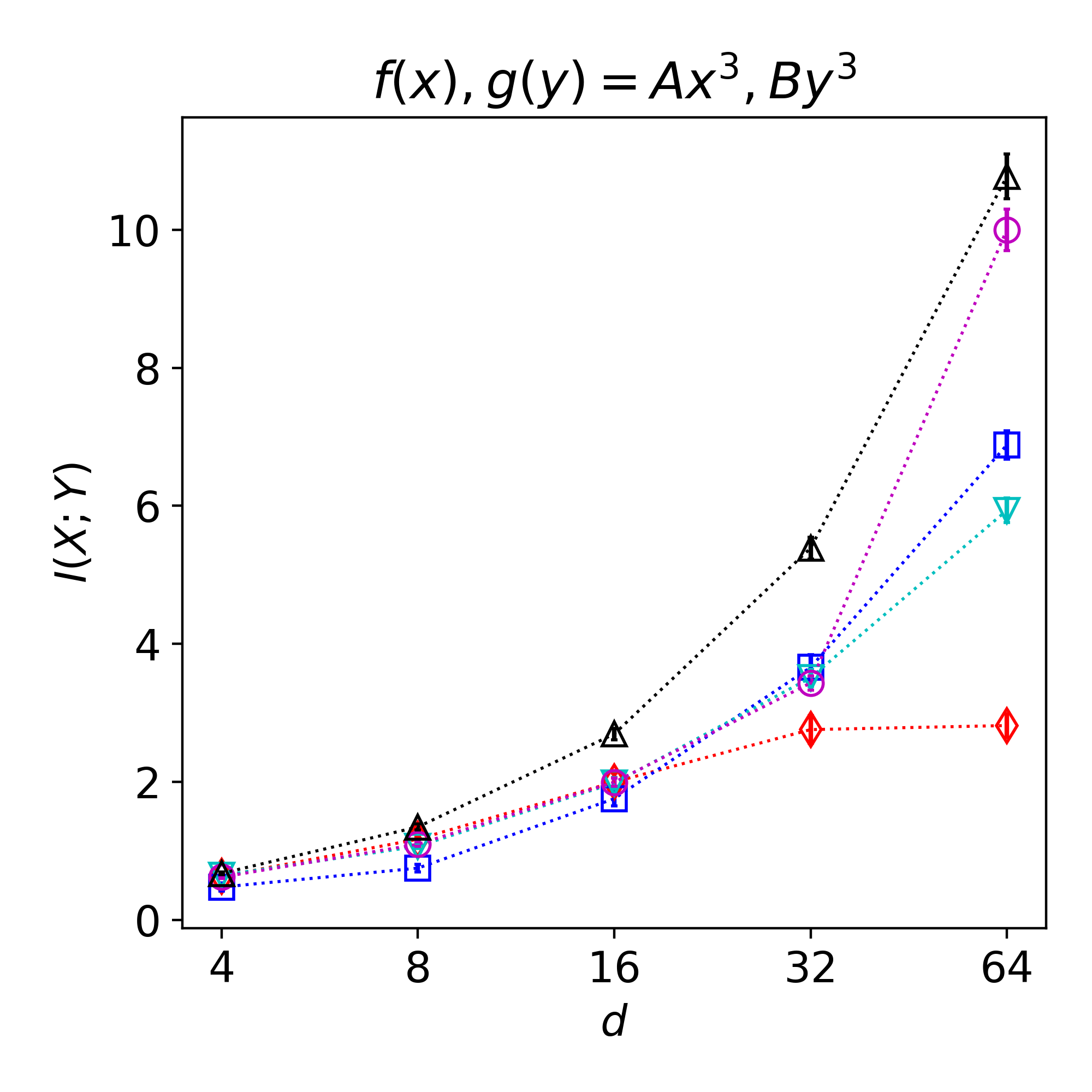}
                    \end{minipage}
            \subcaption{\centering $\textbf{A}X^3, \textbf{B}Y^3$}
            \end{subfigure}
            \hspace{-0.015\linewidth}
            \begin{subfigure}{.205\textwidth}
                    \centering
                    \begin{minipage}[t]{1.0\linewidth}
                    \centering
                     \includegraphics[width=1.0\linewidth]{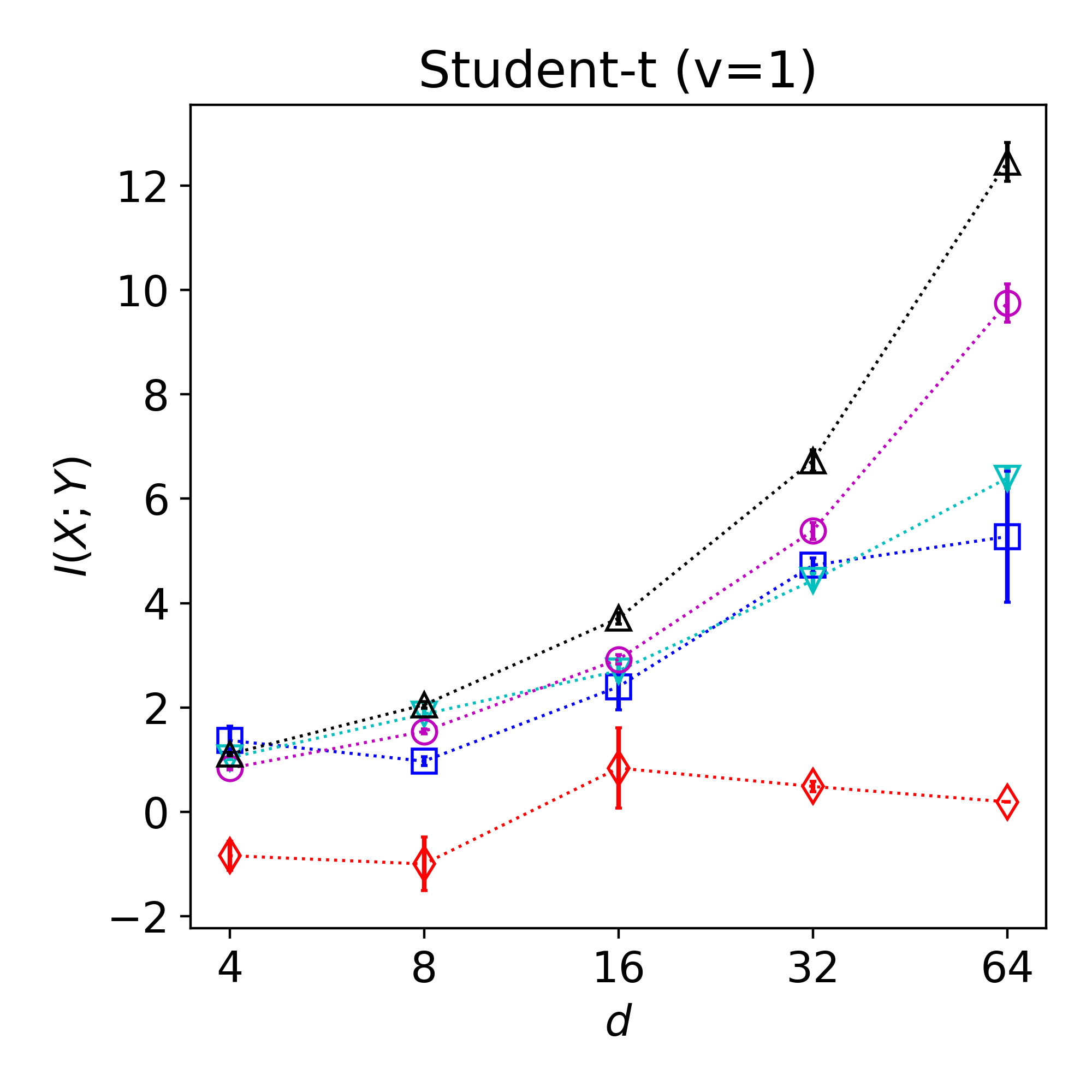}
                    \end{minipage}
            \subcaption{\centering Student-$t$}
            \end{subfigure}
            \hspace{-0.015\linewidth}
            \begin{subfigure}{.205\textwidth}
                    \centering
                    \begin{minipage}[t]{1.0\linewidth}
                    \centering
                     \includegraphics[width=1.0\linewidth]{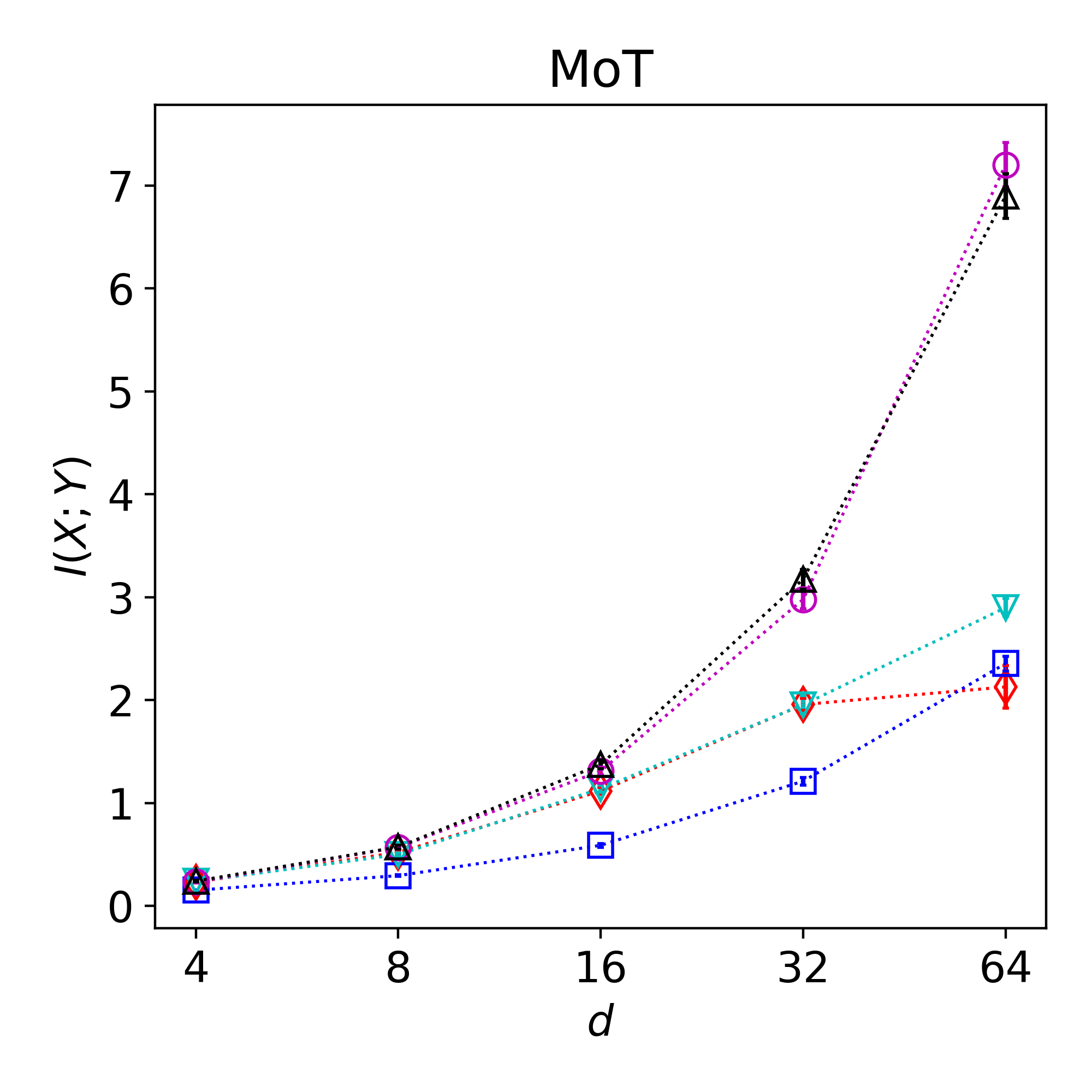}
                    \end{minipage}
            \subcaption{\centering MoT}
            \end{subfigure}
\vspace{-0.10cm}
\caption[Additional results for the VCE' estimator.]{Additional results for the VCE' estimator. In this work, we implement VCE' by taking the reference copula as the independent copula $c'$, so that VCE' is equivalent to performing MINE in the vector copula space. Here MoT stands for `mixture of triangles'. }
\label{fig:synthetic:dim-vce2}
\end{figure*}

\paragraph{VCE' performance}
In the main text, we introduce an alternative estimator, VCE', which models the copula density $c$ using a reference copula $c'$ rather than a mixture of learned vector copulas. In our implementation, $c'$ is chosen to be the independent copula, and we use the MINE loss to estimate the density ratio $r = c / c'$, thereby recovering the target copula as $c = r \cdot c'$. As shown in Figure~\ref{fig:synthetic:dim-vce2}, VCE' serves as a useful and reasonable estimator: it significantly outperforms MINE or closely matches its performance across various settings, although it underperforms compared to our main estimator VCE.

\paragraph{Vector rank computation as data preprocessing} 
In the main text, we discuss the potential of our vector ranks computation method as a versatile data preprocessing for MI estimation. This is evidenced by the comparison between VCE' and MINE in Figure~\ref{fig:synthetic:dim-vce2}: although both use the same loss function, VCE'—which operates in the vector rank space instead of the original data space—consistently outperforms MINE across various settings. The advantage is especially pronounced in scenarios involving heterogeneous marginals (case.b in Figure~\ref{fig:synthetic:dim-vce2}) and heavy-tailed distributions (case.d in Figure~\ref{fig:synthetic:dim-vce2}). These results demonstrate the effectiveness of vector rank computation as a principled data preprocessing technique for enhancing MI estimation.

\paragraph{Capacity-complexity trade-off of the copula}

A core design in our method is an explicit exploration of the complexity-capacity trade-off of the vector copula. We delve into this process to provide further insights into its impact on the estimation accuracy.  

\begin{figure*}[t!]
            \hspace{-0.005\linewidth}
            \begin{subfigure}{.245\textwidth}
                    \centering
                    \begin{minipage}[t]{1.0\linewidth}
                    \centering
                     \includegraphics[width=1.0\linewidth]{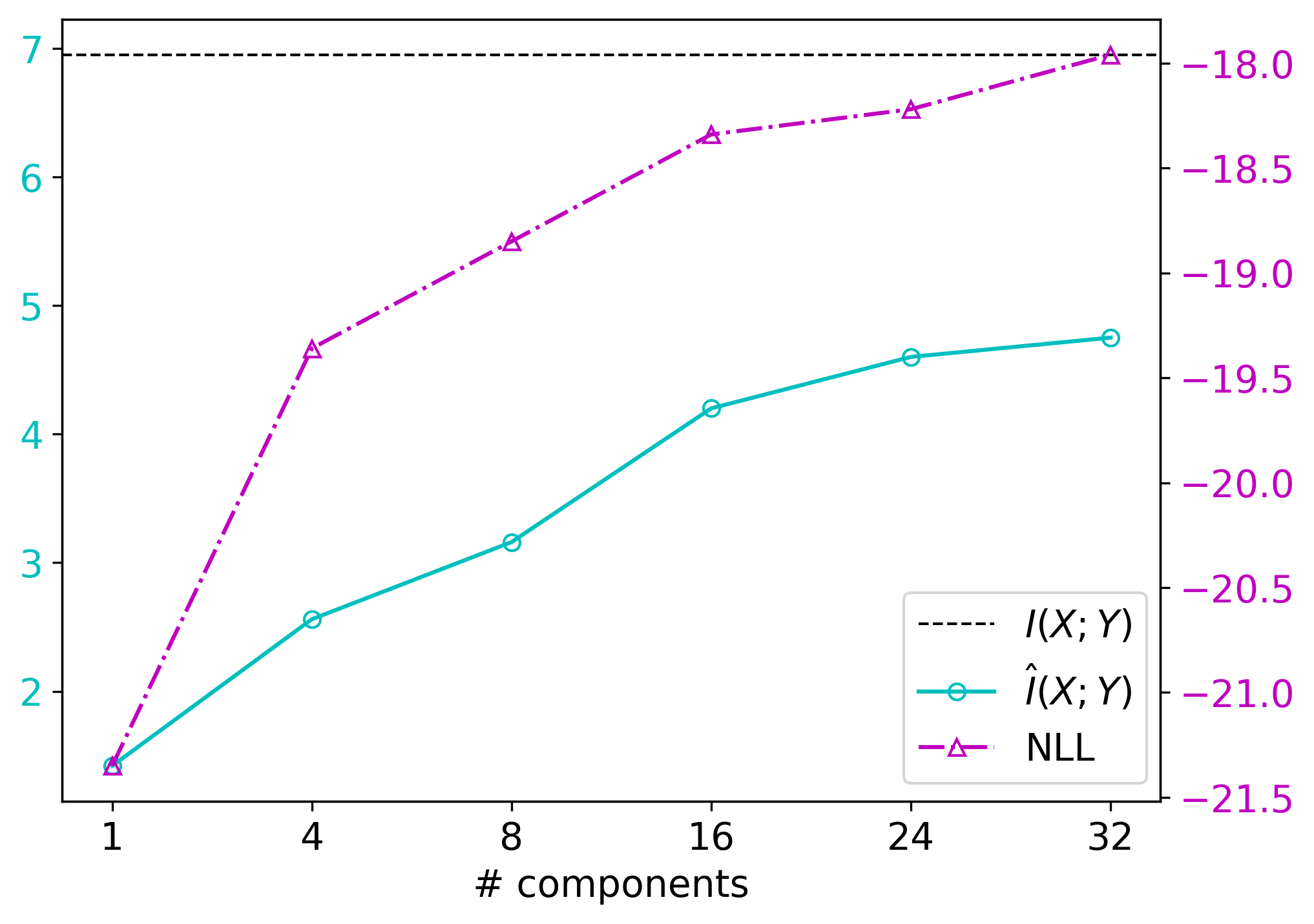}
                    \end{minipage}
            \subcaption{\textbf{Rectangles}}
            \end{subfigure}
            \hspace{-0.0125\linewidth}
            \begin{subfigure}{.245\textwidth}
                    \centering
                    \begin{minipage}[t]{1.0\linewidth}
                    \centering
                     \includegraphics[width=1.0\linewidth]{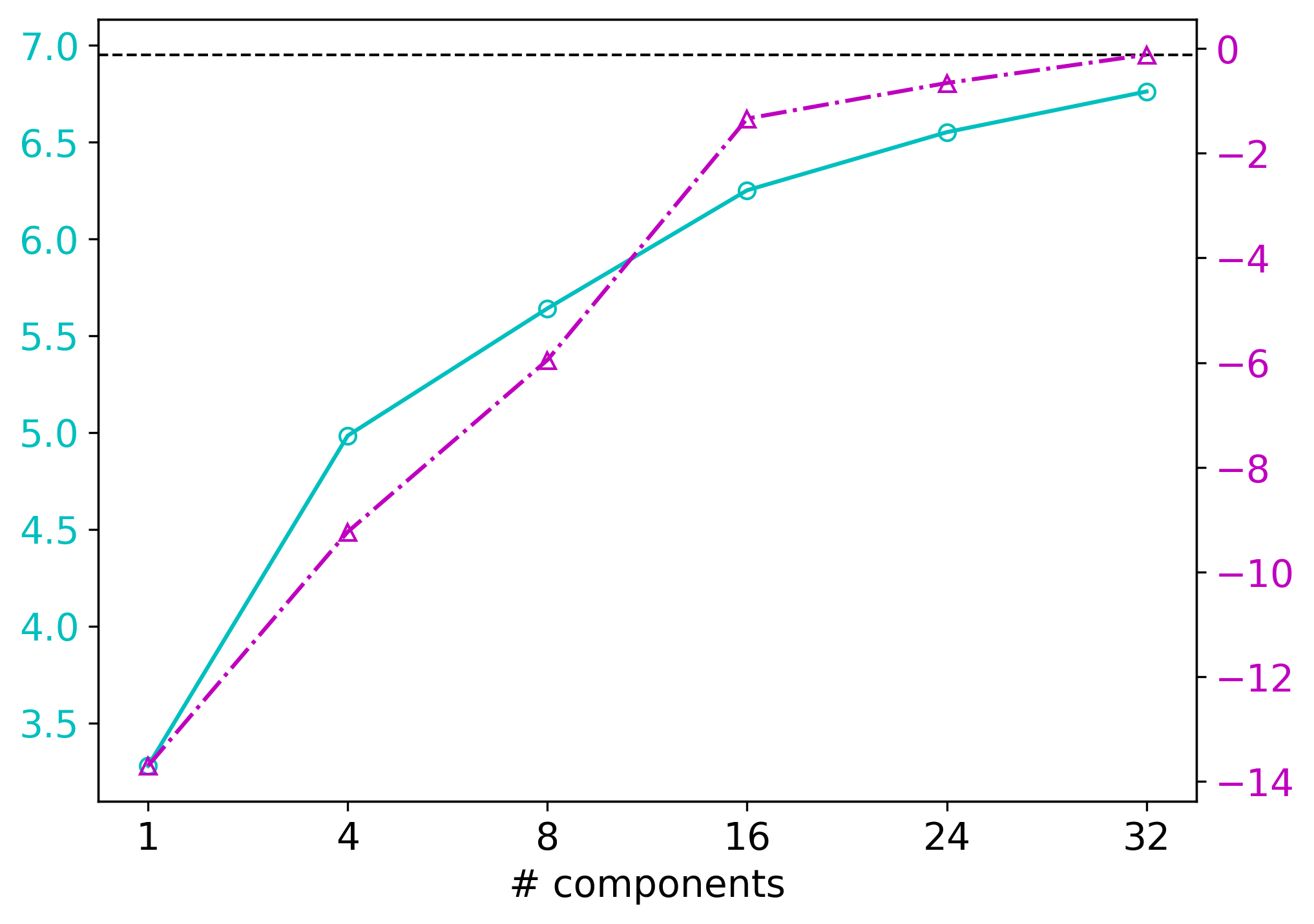}
                    \end{minipage}
            \subcaption{\textbf{Gaussian plates}}
            \end{subfigure}
            \hspace{-0.0125\linewidth}
            \begin{subfigure}{.245\textwidth}
                    \centering
                    \begin{minipage}[t]{1.0\linewidth}
                    \centering
                     \includegraphics[width=1.0\linewidth]{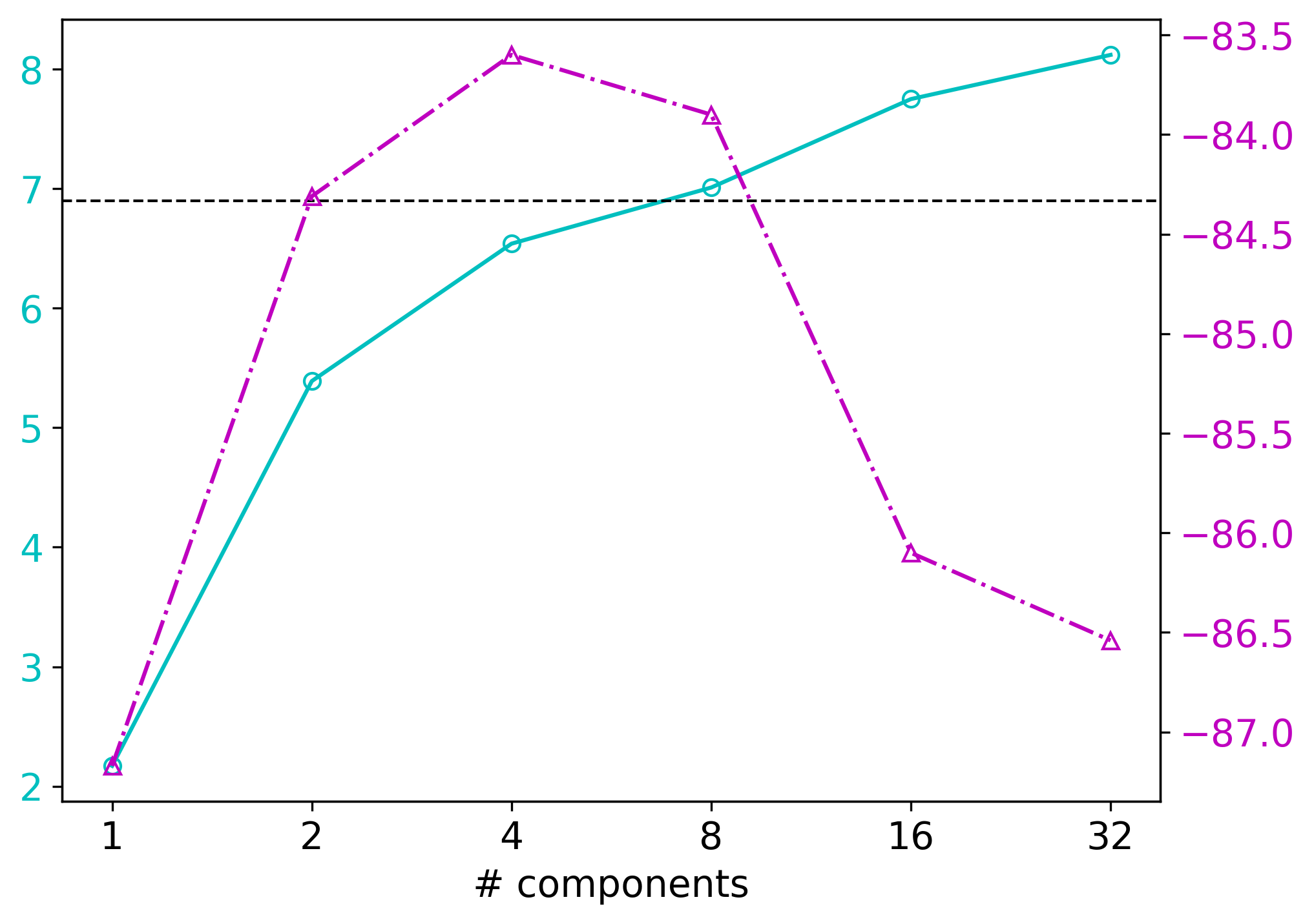}
                    \end{minipage}
            \subcaption{\textbf{MoG}}
            \end{subfigure}
            \hspace{-0.0125\linewidth}
            \begin{subfigure}{.245\textwidth}
                    \centering
                    \begin{minipage}[t]{1.0\linewidth}
                    \centering
                     \includegraphics[width=1.0\linewidth]{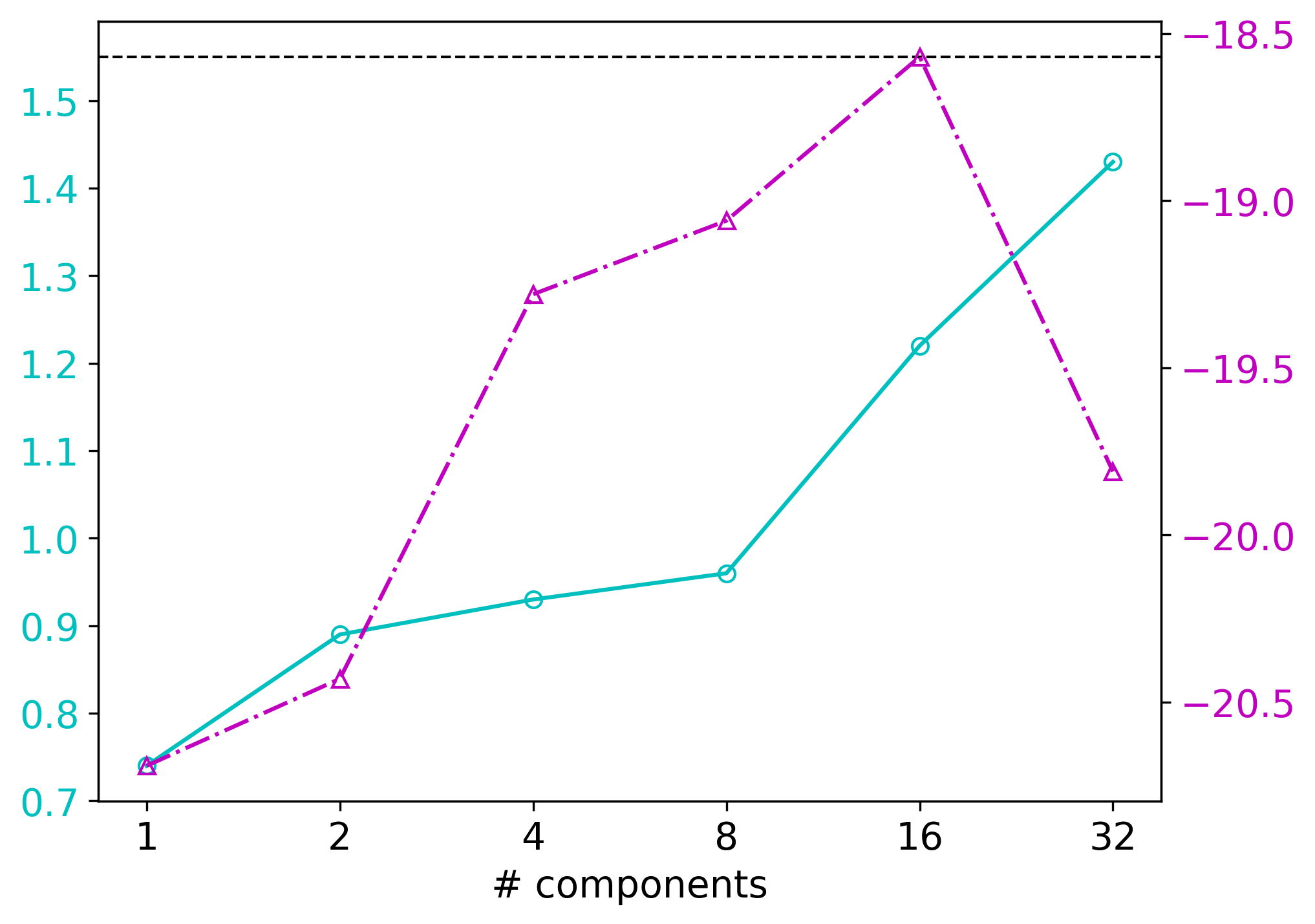}
                    \end{minipage}
            \subcaption{\textbf{BERT}}
            \end{subfigure}
\caption[Exploring the number of components in the vector copula density. ]{Exploring the effect of number of components $K$ in the vector copula density $c$ in the proposed VCE method. The figures shown corresponds to one typical run of the estimator. }
\label{fig:number-of-component-K}
\end{figure*}

Figure~\ref{fig:number-of-component-K} visualizes the model selection procedure described in A. Overall, the negative log-likelihood (NLL) of the vector copula on the validation set generally aligns well with the quality of MI estimate: a higher NLL generally leads to a closer gap between $I(X; Y)$ and $\hat{I}(X; Y)$. Taking the MoG case as example (see Figure~\ref{fig:number-of-component-K}.c), as the capacity of the vector copula density increases, we observe improvements in both the negative log-likelihood (NLL) and the estimated MI. However, when the copula becomes overly complex, both the NLL and MI estimate worsen. A sweet spot is found at $K \approx 6$ mixture components in the copula. The results underscore the importance of the complexity-capacity trade-off of the vector copula\footnote{The trends in NLL and MI are not always perfectly aligned. This is reasonable, as the  NLL is only calculated on a validation set whereas MI is calculated on the full dataset. This leads to an occasional mismatch between the two values, especially when the validation set is not fully representative of the overall data distribution. Nonetheless, the validation NLL remains a reliable proxy for guiding model selection within our framework.}.

In summary, selecting copula with the best complexity-capacity trade-off is important. The NLL on the validation set serves as an effective criterion in this selection process.

\paragraph{Joint learning vs separate learning} 

In addition to the separate \emph{modeling} of marginal distributions and vector copula, an important design of our method is the explicit separation of the \emph{learning} of marginal and copula. We provide  empirical evidence to highlight the advantange of this design. 

\begin{table}[t]
\centering
\hspace{0.00\textwidth}
\begin{subfigure}{0.495\textwidth}
\begin{minipage}{1.0\textwidth}
 \begin{tabular}{lcccc}
    \toprule
    \cmidrule(r){1-5}
	& med	& std	& fail*	& $I(X; Y)$ \\
\midrule
Student-$t$	 & 7.81 & 5.55 & 2/10 & 12.4	 \\
   \midrule
$\mathbf{A}X^3, \mathbf{B}e^Y$	& 6.02 & 0.98 & 1/10 & 10.8		 \\
    \bottomrule \\
  \end{tabular}
\end{minipage}
\subcaption{\textbf{Joint learning}}
\end{subfigure}
\hspace{-0.01\textwidth}
\begin{subfigure}{0.495\textwidth}
\begin{minipage}{1.0\textwidth}
 \begin{tabular}{lcccc}
    \toprule
    \cmidrule(r){1-5}
	& med	& std	& fail*	& $I(X; Y)$ \\
\midrule
Student-$t$ & 9.50 & 0.35 & 0/10 & 12.4	 \\
   \midrule
$\mathbf{A}X^3, \mathbf{B}e^Y$	& 8.12 & 0.12 & 0/10 & 10.8		 \\
    \bottomrule \\
  \end{tabular}
\end{minipage}
\subcaption{\textbf{Separate learning}}
\end{subfigure}
\caption[Comparing the accuracy of joint learning and separate learning. ]{Joint learning vs separate learning. Results are collected from 10 independent runs. Data dimensionality is 64. *Fail: fraction of runs where $|\hat{I}(X; Y) - I(X; Y)| > \frac{1}{2}I(X; Y)$. The student-$t$ distribution is with degree of freedom $\nu = 1$. The case `$\mathbf{A}X^3, \mathbf{B}e^Y$' corresponds to applying the shown transformation to $X, Y \sim \mathcal{N}$, where $\mathbf{A}$ and $\mathbf{B}$ are invertible matrices. }
\label{tab:joint-vs-separate-1}
\end{table}

In Table~\ref{tab:joint-vs-separate-1}, we compare the estimations obtain via joint learning and separate learning on two challenging cases: a 64-dimensional $t$-distribution with degree of freedom $\nu = 1$, and a distribution with heterogeneous marginal characteristics. As expected, separate learning produces not only more accurate and but also more robust estimation in both cases, as indicated by lower bias and reduced standard deviation. Importantly, we observe that for these two challenging cases, jointly learning the marginal and copula occasionally fails, returning highly biased MI in approximately 2 out of 10 independent runs. This issue does not occur with separate learning. The result highlights the advantage of separate learning in certain cases, which avoids directly learning the marginal distribution and the vector copula altogether --- a task that could be otherwise overly challenging.

Beyond accuracy and robustness, separate learning also improves computational efficiency, particularly in the context of model selection. In practice, we observe that separate learning achieves a 2.1$\sim$3.7 times acceleration over joint learning. This gain attributes to the fact that we only need to train multiple lightweight models in the copula space, rather than multiple full joint models.

\paragraph{Quality of autoencoder-based compression}
As noted in the main text, we preprocess the image and text datasets using an autoencoder. The quality of this compression is crucial, as highly lossy compression will lead to inaccurate assessment of the performance of different estimators. We investigate the quality of this compression.

\begin{figure*}[t!]
            \hspace{-0.008\linewidth}
            \begin{subfigure}{.495\textwidth}
                    \centering
                    \begin{minipage}[t]{1.0\linewidth}
                    \centering
                     \includegraphics[width=1.0\linewidth]{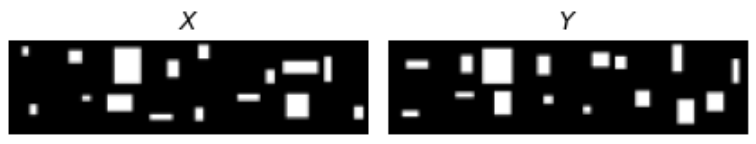}
                    \end{minipage}
            \subcaption{\textbf{Rectangles, original}}
            \end{subfigure}
            \hspace{0.010\linewidth}
            \begin{subfigure}{.49\textwidth}
                    \centering
                    \begin{minipage}[t]{1.0\linewidth}
                    \centering
                     \includegraphics[width=1.0\linewidth]{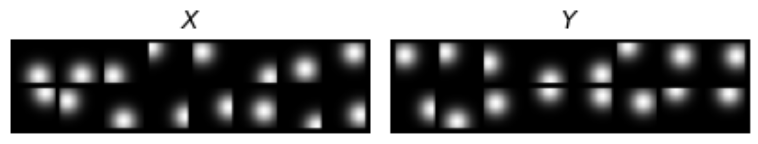}
                    \end{minipage}
            \subcaption{\textbf{Gaussian plates, original}}
            \end{subfigure}
            \hspace{-0.016\linewidth}
            \begin{subfigure}{.495\textwidth}
                    \centering
                    \begin{minipage}[t]{1.0\linewidth}
                    \centering
                     \includegraphics[width=1.0\linewidth]{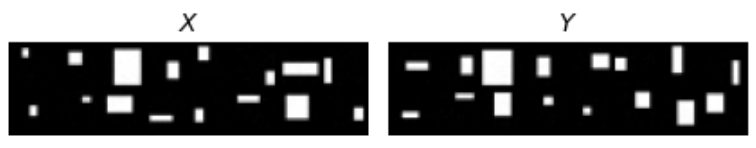}
                    \end{minipage}
            \subcaption{\textbf{Rectangles, reconstructed}}
            \end{subfigure}
            \hspace{0.010\linewidth}
            \begin{subfigure}{.488\textwidth}
                    \centering
                    \begin{minipage}[t]{1.0\linewidth}
                    \centering
                     \includegraphics[width=1.0\linewidth]{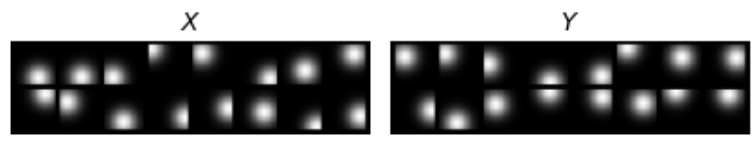}
                    \end{minipage}
            \subcaption{\textbf{Gaussian plates, reconstructed}}
            \end{subfigure}
            
\caption[Quality of autoencoder-based compression.]{Quality of autoencoder-based compression. Upper panel: original data. Lower panel: reconstructed data with 16-dimensionality latent representation. The compression is near-lossless. }
\label{fig:iamge-dataset-recon}
\end{figure*}

\begin{table}[t]
\centering
\hspace{0.00\textwidth}
\begin{subfigure}{0.492\textwidth}
\begin{minipage}{1.0\textwidth}
  \resizebox{\textwidth}{!}{
 \begin{tabular}{lcccc}
    \toprule
    \cmidrule(r){1-5}
	$d_{\text{latent}}$ & 4	& 8	& 16	& 32 \\
\midrule
Relative MSE & 3e-2	& 2e-2	& 8e-3	& 7e-3 	 \\
    \bottomrule \\
  \end{tabular}
  }
\end{minipage}
\subcaption{\textbf{Image - Rectangles}}
\end{subfigure}
\hspace{-0.01\textwidth}
\begin{subfigure}{0.492\textwidth}
\begin{minipage}{1.0\textwidth}
  \resizebox{\textwidth}{!}{
 \begin{tabular}{lcccc}
    \toprule
    \cmidrule(r){1-5}
	$d_{\text{latent}}$ & 8	& 16	& 32	& 64 \\
\midrule
Relative MSE & 1e-3 & 5e-4 & 3e-4 & 2e-4	 \\
    \bottomrule \\
  \end{tabular}
  }
\end{minipage}
\subcaption{\textbf{Bert embeddings}}
\end{subfigure}
\caption[Quality of compression. ]{Quality of autoencoder compression. Relative MSE is defined as $\mathbb{E}[\| h(e(\vecx)) - \vecx \|^2_2 / \|\vecx\|^2_2]$. Here $h: \R^{d_{\text{latent}}} \to \R^{d_{\text{data}}}$ is the decoder. Moderately large $d_{\text{latent}}$ yields near-lossless compression.   }
\label{tab:compression}
\end{table}

Table~\ref{tab:compression} reports the \emph{relative} mean squared error (Relative MSE) of reconstruction, defined as 
\[
\mathbb{E}[| h(e(\vecx)) - \vecx |^2_2 / |\vecx|^2_2]
\]
where $e: \R^{d_{\text{data}}} \to \R^{d_{\text{latent}}}$ is the encoder and $h: \R^{d_{\text{latent}}} \to \R^{d_{\text{data}}}$ is the decoder. The results show that reconstruction is nearly perfect for both datasets under the chosen latent dimensionalities ($d_{\text{latent}} = 16$ for the image dataset and $d_{\text{latent}} = 32$ for the text dataset), indicating that the compression retains almost all the original information: $I(X; Y) \approx I(e(X); e(Y))$, as grounded by Proposition 5 above.

\begin{figure*}[t!]
            \hspace{-0.02\linewidth}
            \begin{subfigure}{.205\textwidth}
                    \centering
                    \begin{minipage}[t]{0.99\linewidth}
                    \centering
                     \includegraphics[width=1.0\linewidth]{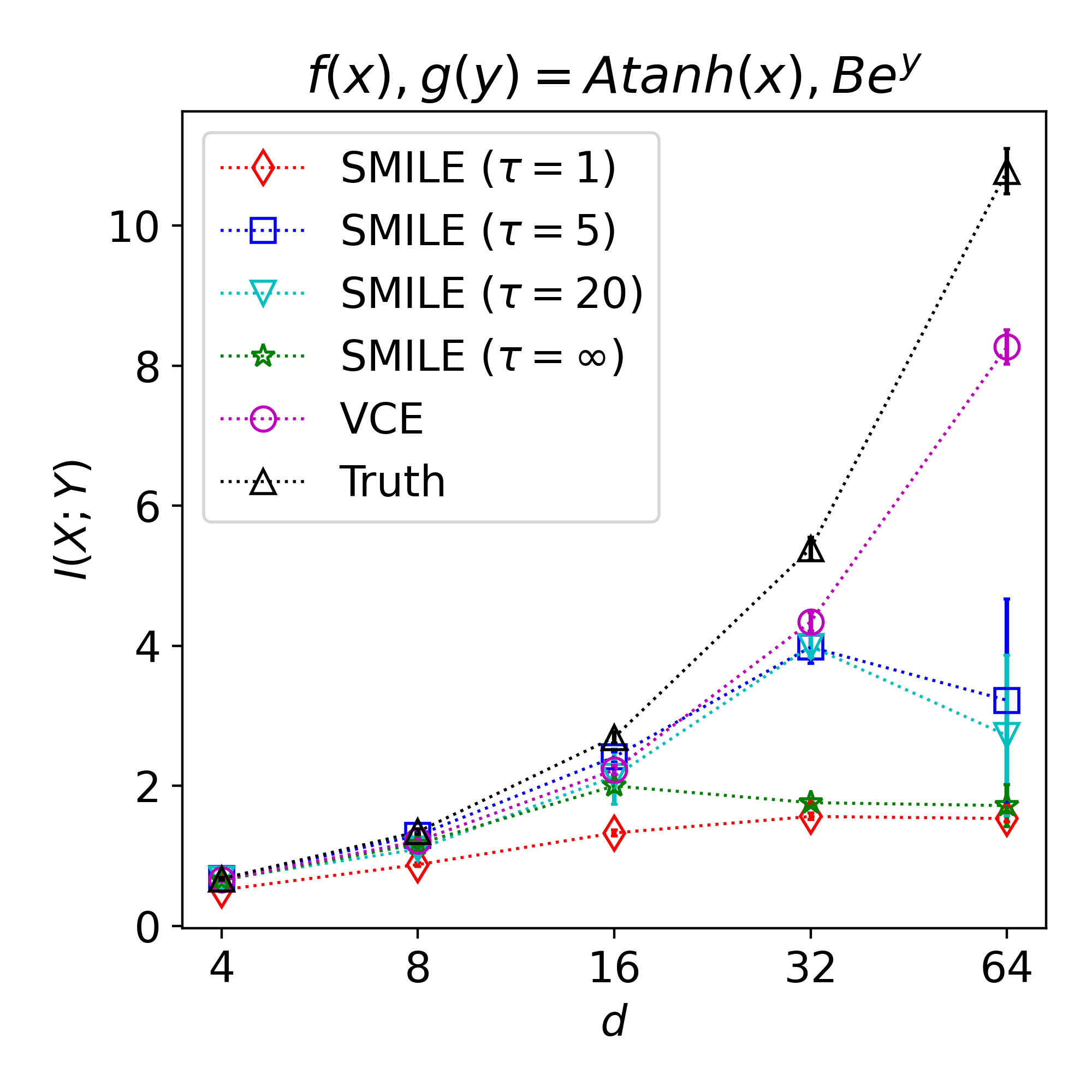}
                    \end{minipage}
            \subcaption{\centering $\textbf{A}\text{tanh}(X), \textbf{B}e^Y$}
            \end{subfigure}
            \hspace{-0.015\linewidth}
            \begin{subfigure}{.205\textwidth}
                    \centering
                    \begin{minipage}[t]{1.0\linewidth}
                    \centering
                     \includegraphics[width=1.0\linewidth]{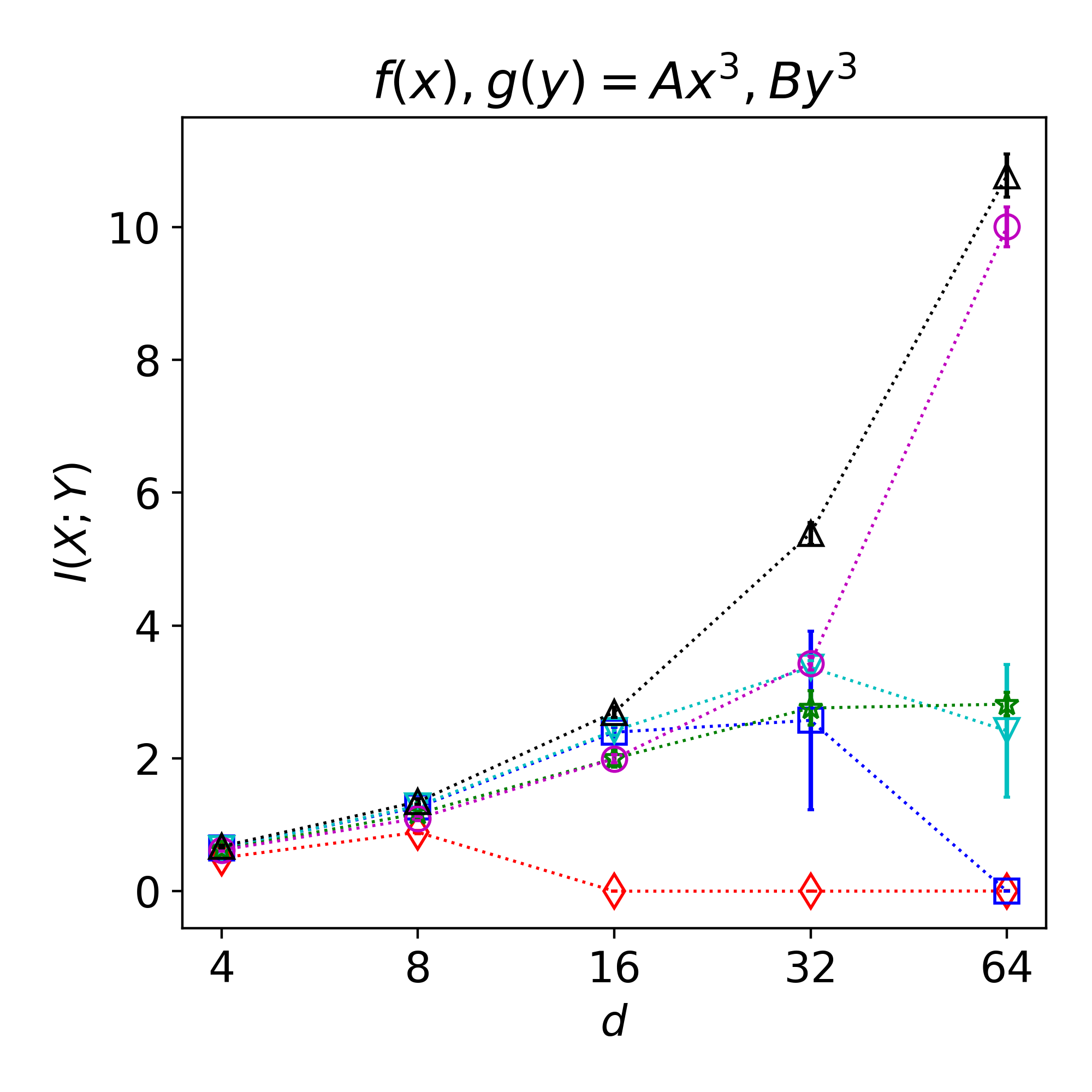}
                    \end{minipage}
            \subcaption{\centering $\textbf{A}X^3, \textbf{B}Y^3$}
            \end{subfigure}
            \hspace{-0.015\linewidth}
            \begin{subfigure}{.205\textwidth}
                    \centering
                    \begin{minipage}[t]{1.0\linewidth}
                    \centering
                     \includegraphics[width=1.0\linewidth]{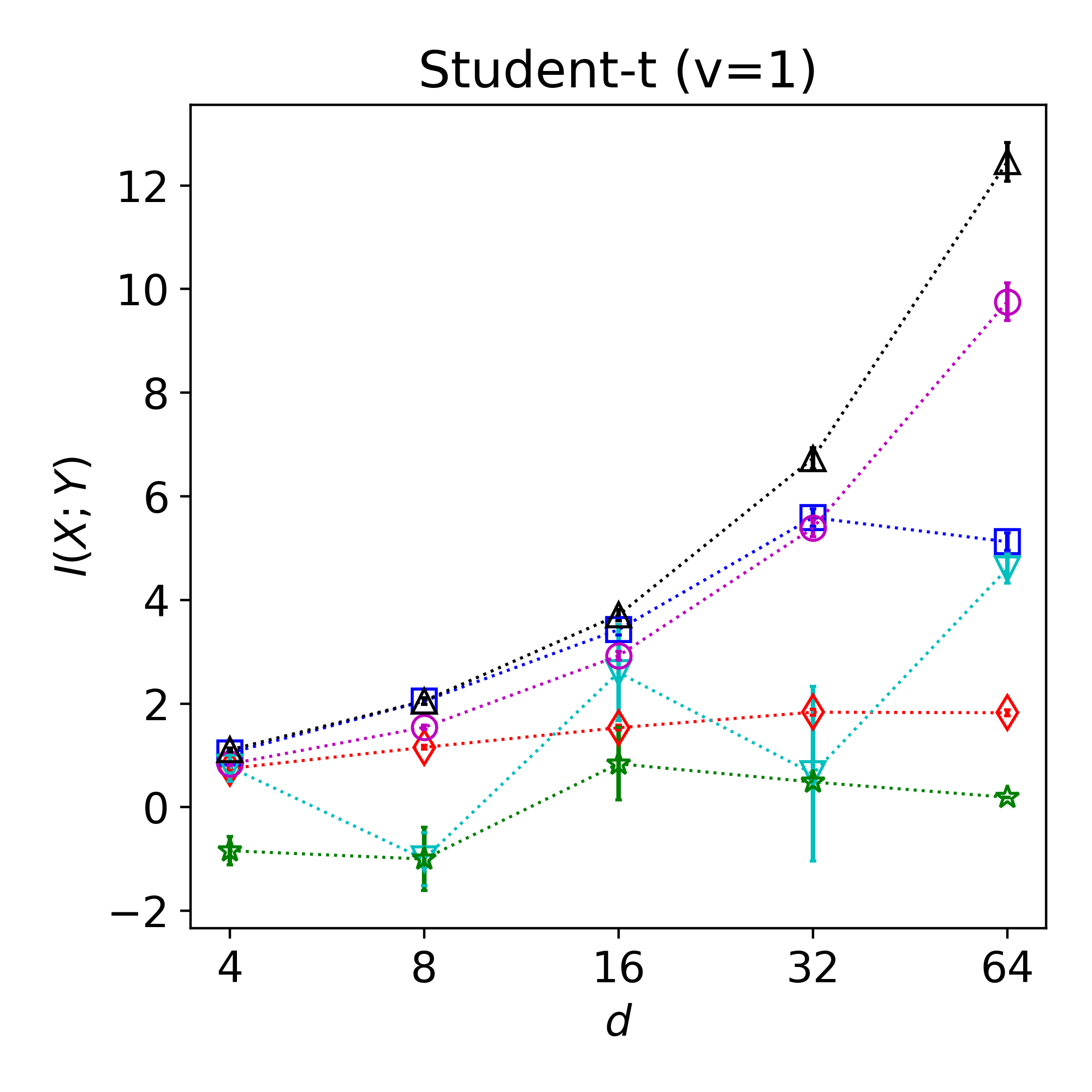}
                    \end{minipage}
            \subcaption{\centering Student-$t$}
            \end{subfigure}
            \hspace{-0.015\linewidth}
            \begin{subfigure}{.205\textwidth}
                    \centering
                    \begin{minipage}[t]{1.0\linewidth}
                    \centering
                     \includegraphics[width=1.0\linewidth]{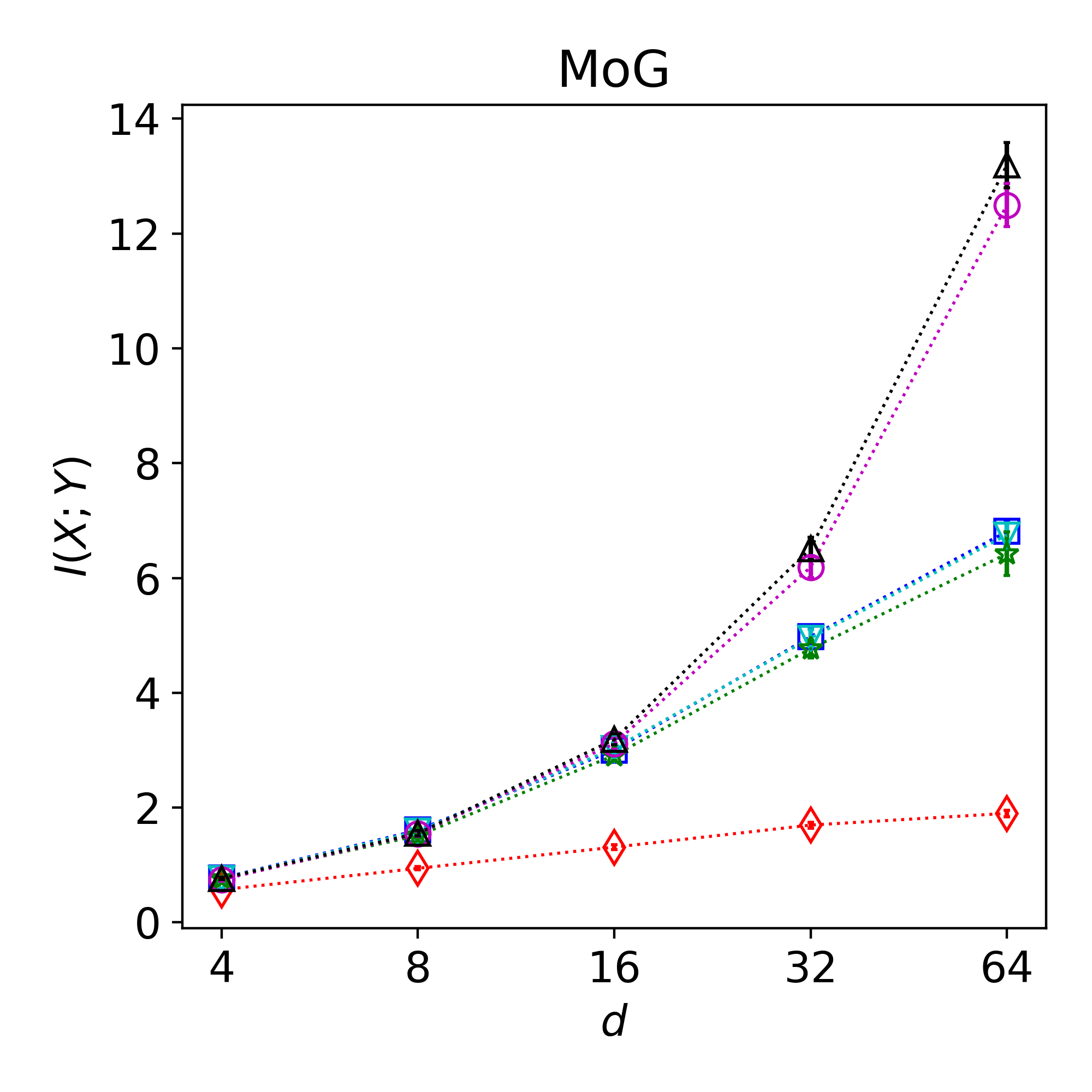}
                    \end{minipage}
            \subcaption{\centering MoG}
            \end{subfigure}
            \hspace{-0.015\linewidth}
            \begin{subfigure}{.205\textwidth}
                    \centering
                    \begin{minipage}[t]{1.0\linewidth}
                    \centering
                     \includegraphics[width=1.0\linewidth]{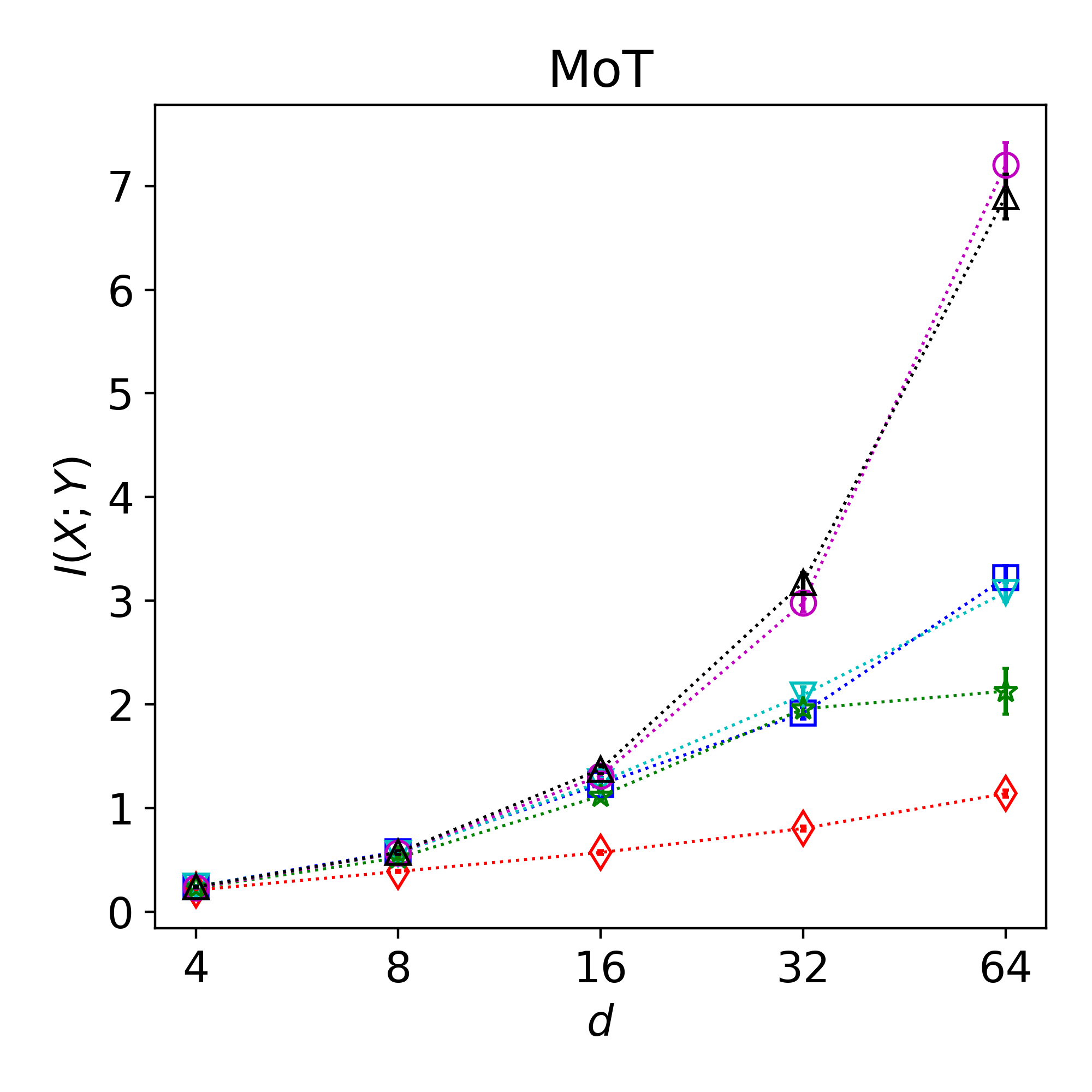}
                    \end{minipage}
            \subcaption{\centering MoT}
            \end{subfigure}
\vspace{-0.10cm}
\caption[Comparing VCE with the SMILE estimator.]{Comparison with the SMILE estimator under different clipping values $\tau$.  }
\label{fig:comparison-to-SMILE}
\end{figure*}

\textbf{Comparison to SMILE}. We additionally compare our method to SMILE~\cite{song2019understanding}, a robust MI estimator that also provides explicit control over the trade-off between model complexity and capacity, akin to our method. This estimator is defined as
\[
    \hat{I}(X; Y)_{\text{SMILE}} \coloneqq \sup_T \, \mathbb{E}_{p(\vecx, \vecy)}[T(\vecx, \vecy)] - \log \mathbb{E}_{p(\vecx)p(\vecy)}[e^{T(\vecx, \vecy)}],
\]
where
\[
    T(\vecx, \vecy) = \mathrm{MLP}(\vecx, \vecy).\mathrm{clip}(-\tau, \tau),
\]
The function $T: \mathcal{X} \times \mathcal{Y} \to \R$ is a neural network (typically an MLP) whose output is clipped to the range $[-\tau, \tau]$. The clipping parameter $\tau$ governs the balance between expressiveness and variance:
\begin{itemize}[leftmargin=*]
    \item A larger $\tau$ allows the model to capture complex dependencies but increases the estimation variance.
    \item A smaller $\tau$ suppresses variance by limiting flexibility, but may reduce the model's expressiveness.
\end{itemize}
Figure~\ref{fig:comparison-to-SMILE} presents the results, highlighting the superior performance of our proposed VCE method. \\

\begin{figure*}[t!]
            \hspace{-0.02\linewidth}
            \begin{subfigure}{.205\textwidth}
                    \centering
                    \begin{minipage}[t]{0.99\linewidth}
                    \centering
                     \includegraphics[width=1.0\linewidth]{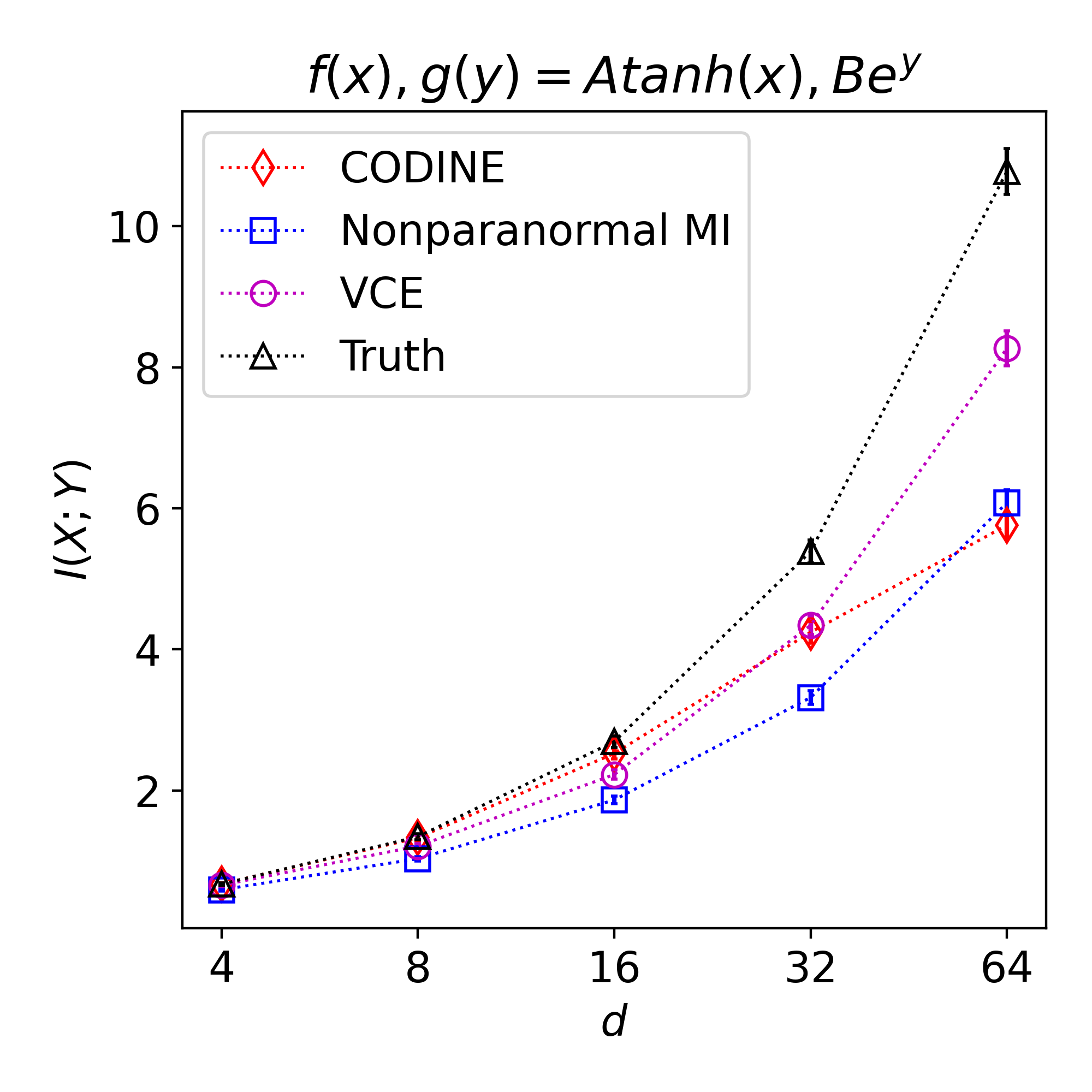}
                    \end{minipage}
            \subcaption{\centering $\textbf{A}\text{tanh}(X), \textbf{B}e^Y$}
            \end{subfigure}
            \hspace{-0.015\linewidth}
            \begin{subfigure}{.205\textwidth}
                    \centering
                    \begin{minipage}[t]{1.0\linewidth}
                    \centering
                     \includegraphics[width=1.0\linewidth]{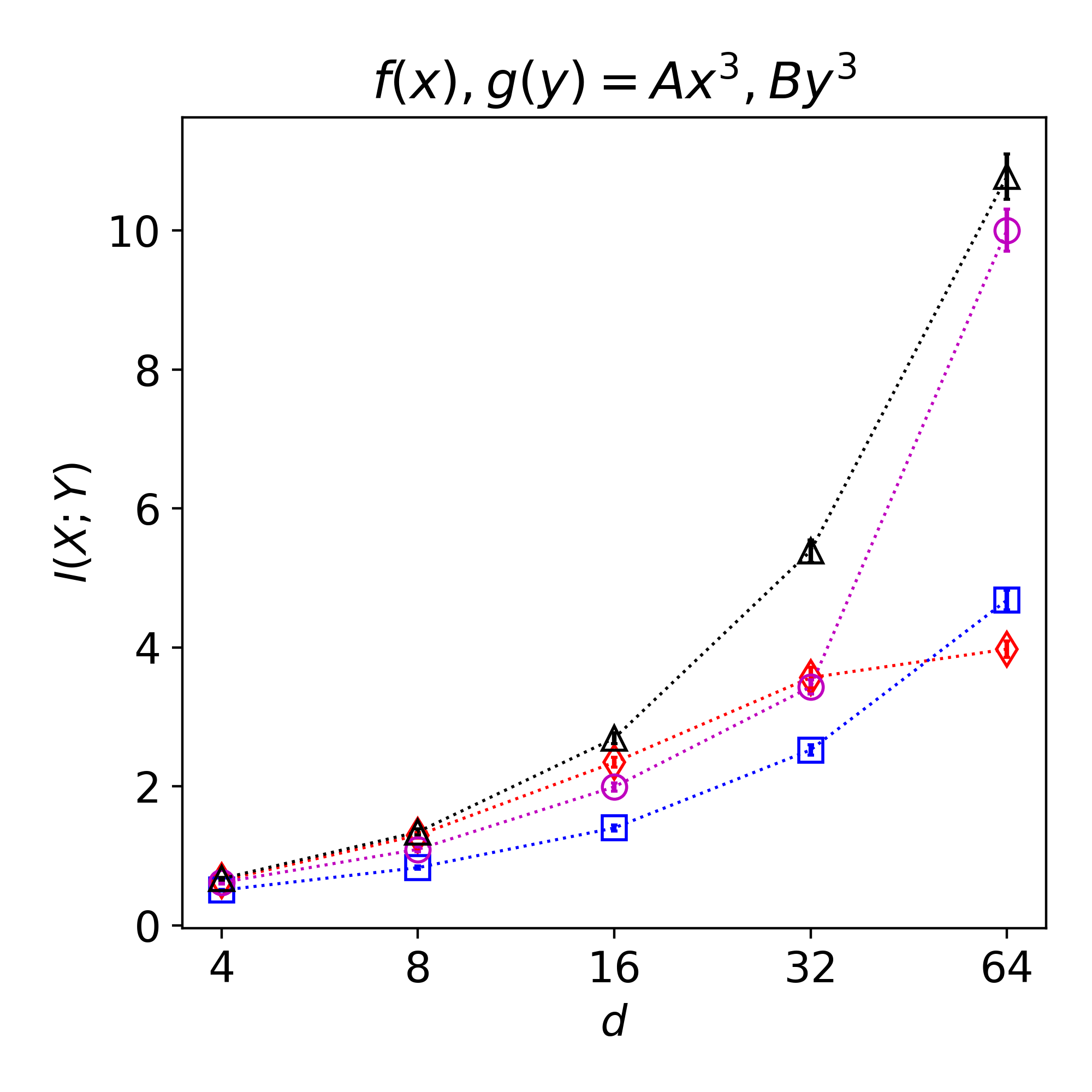}
                    \end{minipage}
            \subcaption{\centering $\textbf{A}X^3, \textbf{B}Y^3$}
            \end{subfigure}
            \hspace{-0.015\linewidth}
            \begin{subfigure}{.205\textwidth}
                    \centering
                    \begin{minipage}[t]{1.0\linewidth}
                    \centering
                     \includegraphics[width=1.0\linewidth]{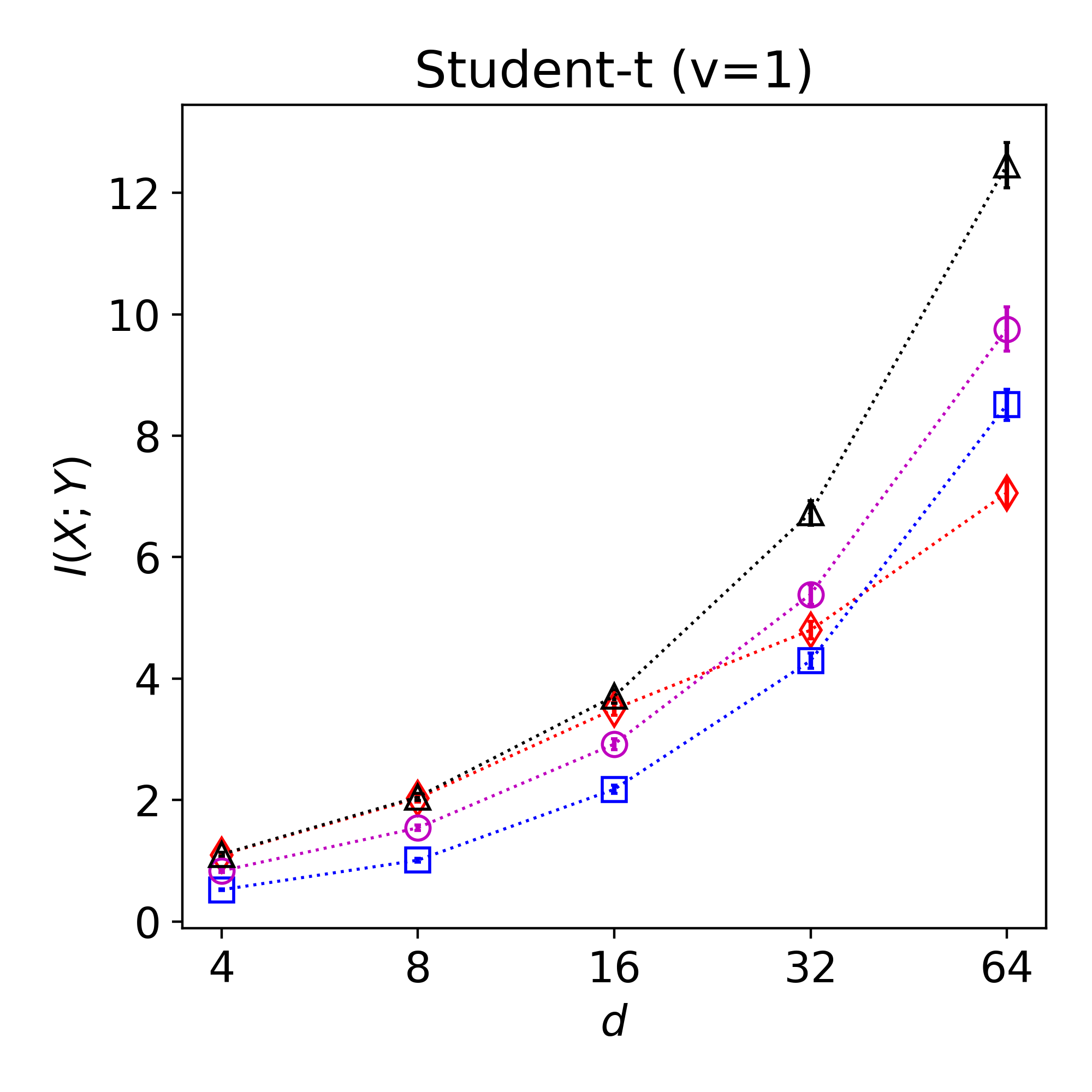}
                    \end{minipage}
            \subcaption{\centering Student-$t$}
            \end{subfigure}
            \hspace{-0.015\linewidth}
            \begin{subfigure}{.205\textwidth}
                    \centering
                    \begin{minipage}[t]{1.0\linewidth}
                    \centering
                     \includegraphics[width=1.0\linewidth]{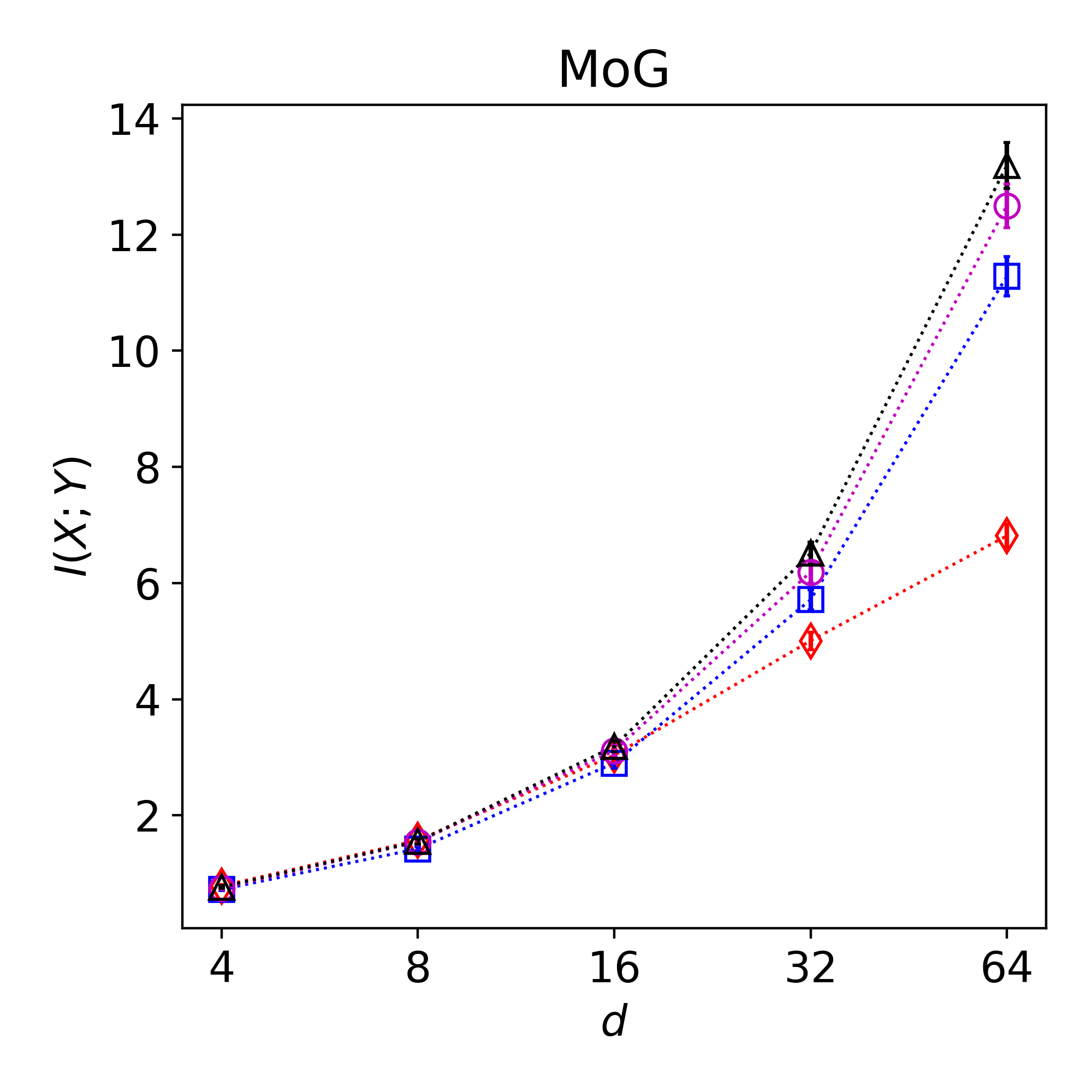}
                    \end{minipage}
            \subcaption{\centering MoG}
            \end{subfigure}
            \hspace{-0.015\linewidth}
            \begin{subfigure}{.205\textwidth}
                    \centering
                    \begin{minipage}[t]{1.0\linewidth}
                    \centering
                     \includegraphics[width=1.0\linewidth]{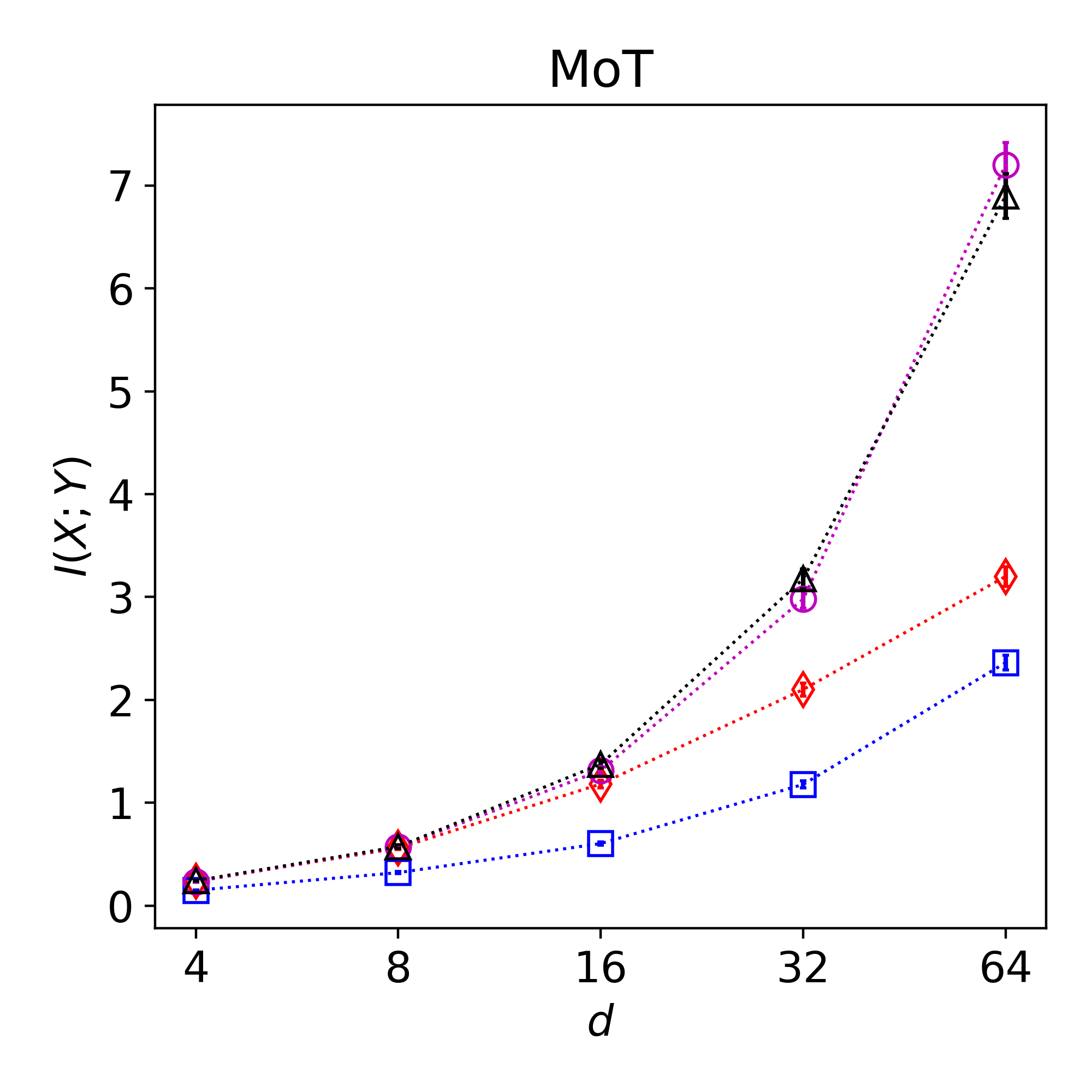}
                    \end{minipage}
            \subcaption{\centering MoT}
            \end{subfigure}
\vspace{-0.10cm}
\caption[Comparing VCE with classic copula-based estimators.]{Comparing VCE with classic copula-based estimators e.g. nonparanormal MI (which uses a Gaussian copula to estimate MI) and CODINE (equivalent to classic copula transformation + MINE). }
\label{fig:comparison-with-classic-copula-methods}
\end{figure*}

\textbf{Comparison to classic copula-based MI estimator}. We further compare our method against two \emph{classic} copula-based approaches, which rely on parametric and neural models for copula modeling, respectively:
\begin{itemize}[leftmargin=*]
    \item Nonparanormal information estimation (\emph{Nonparanormal MI}\cite{singh2017nonparanormal}): This method assumes the data can be approximated by a Gaussian copula model and directly computes MI induced by the corresponding Gaussian copula model.
    \item Copula neural density estimation (\emph{CODINE}~\cite{letizia2025copula}): This method models the copula by a deep neural network and computes MI based on the (classic) copula of the joint distribution and that of the product of marginals.
\end{itemize}
Figure~\ref{fig:comparison-with-classic-copula-methods} reports the results. Our proposed VCE estimator consistently outperforms both methods, underscoring the benefits of leveraging vector copulas over classic copula for information estimation. \\

\begin{figure*}[t!]
            \begin{subfigure}{.50\textwidth}
                    \centering
                    \hspace{-0.075\linewidth}
                    \begin{minipage}[t]{1.0\linewidth}
                    \centering
                     \includegraphics[width=1.0\linewidth]{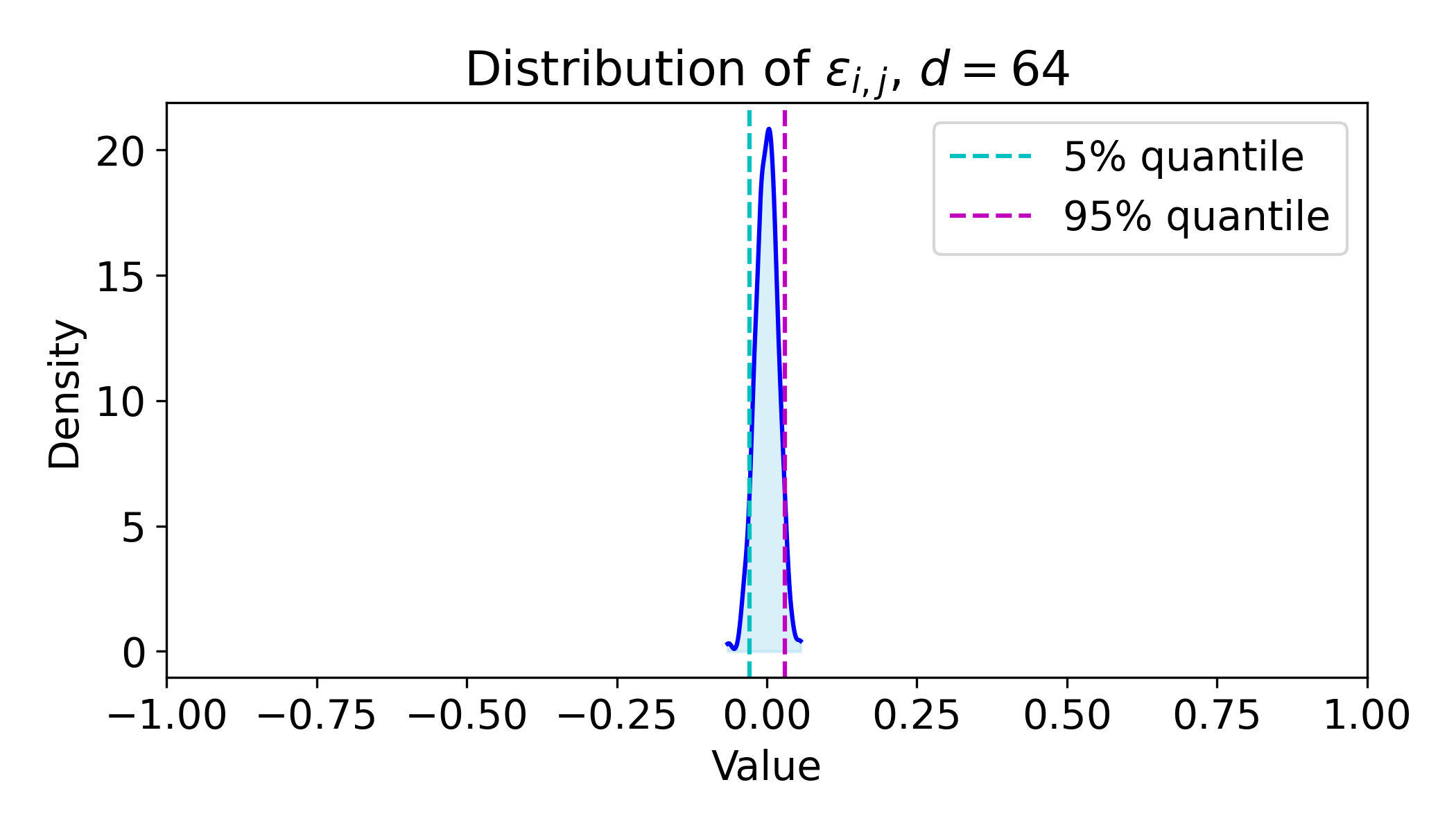}
                    \end{minipage}
            \subcaption{\text{Spiral distribution, 64D}}
            \end{subfigure}
            \begin{subfigure}{.50\textwidth}
                    \centering
                    \begin{minipage}[t]{1.0\linewidth}
                    \centering
                     \includegraphics[width=1.0\linewidth]{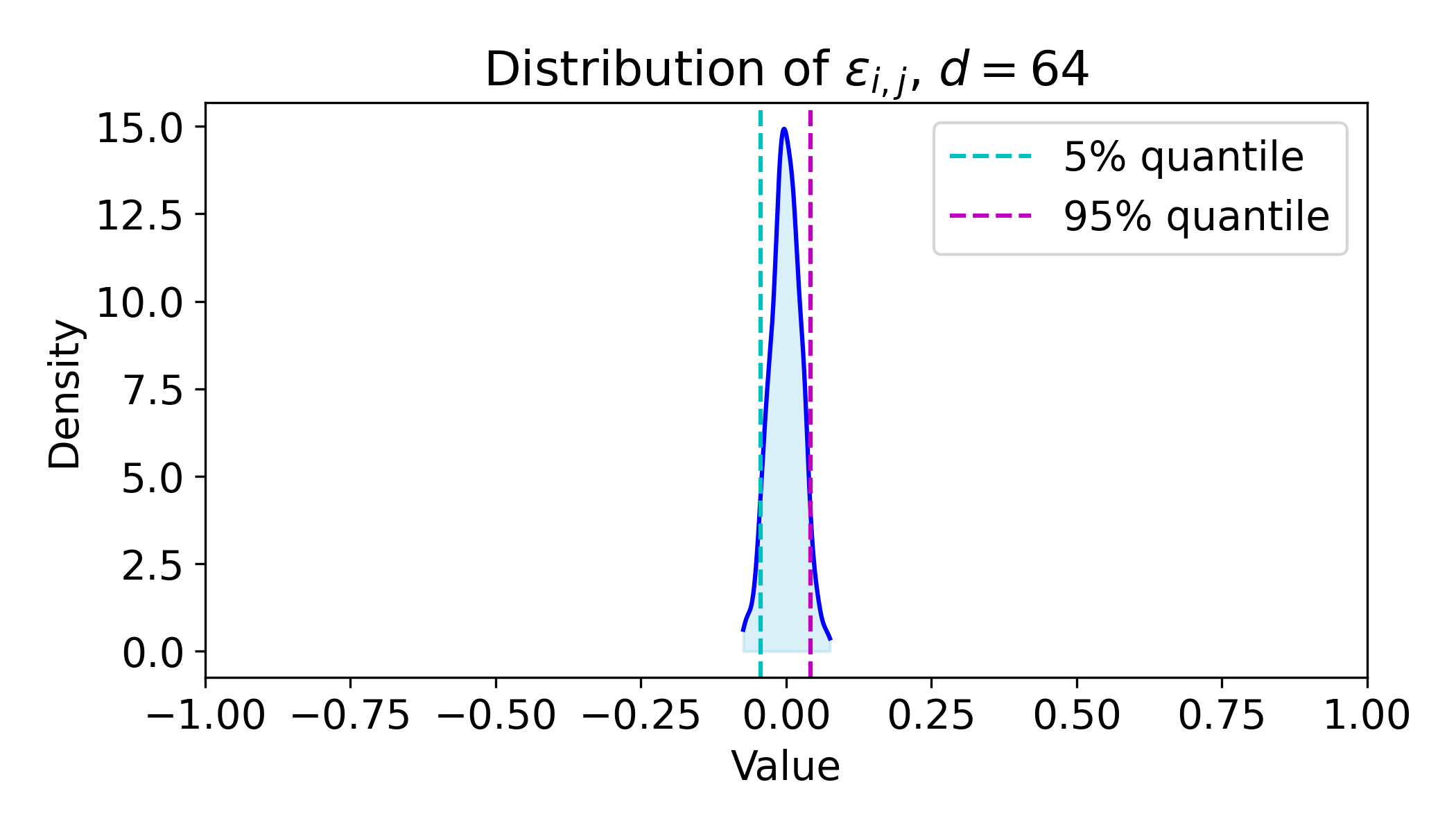}
                    \end{minipage}
            \subcaption{\text{$t$-distribution ($\nu = 1$), 64D}}
            \end{subfigure}
            \begin{subfigure}{.50\textwidth}
                    \centering
                    \hspace{-0.075\linewidth}
                    \begin{minipage}[t]{1.0\linewidth}
                    \centering
                     \includegraphics[width=1.0\linewidth]{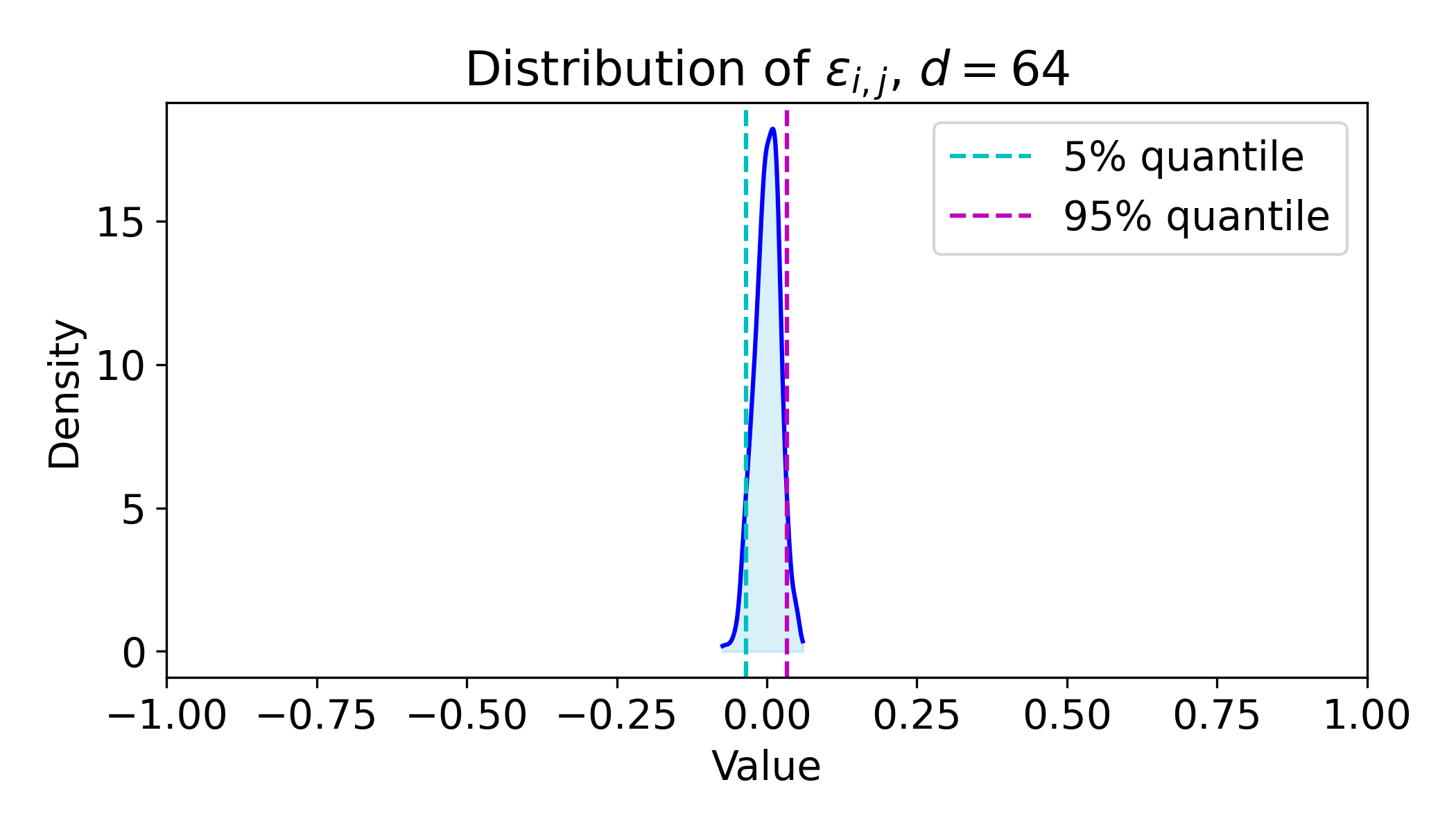}
                    \end{minipage}
            \subcaption{\text{Mixture of Gaussians, 64D}}
            \end{subfigure}
            \begin{subfigure}{.50\textwidth}
                    \centering
                    \begin{minipage}[t]{1.0\linewidth}
                    \centering
                     \includegraphics[width=1.0\linewidth]{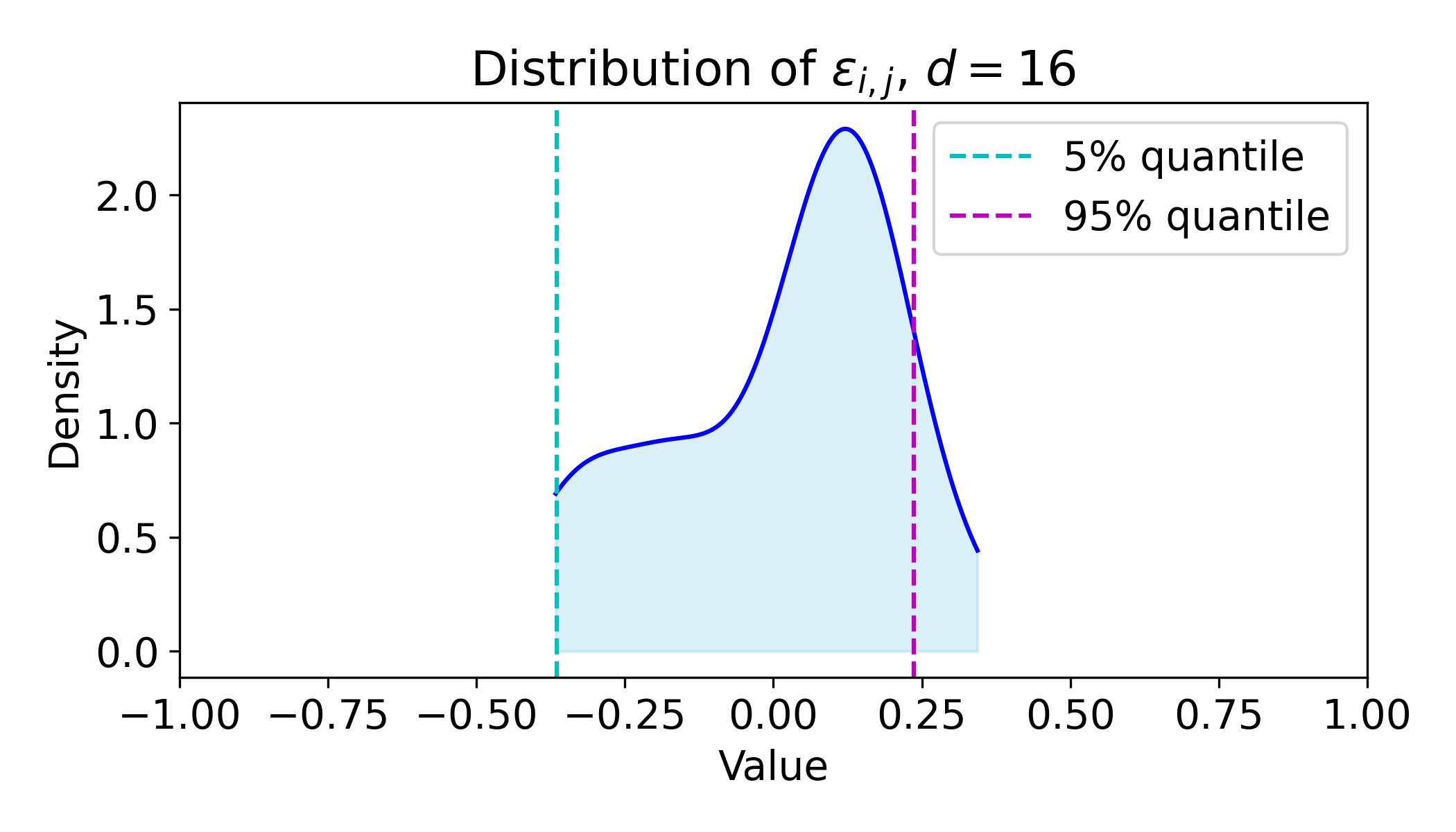}
                    \end{minipage}
            \subcaption{\text{Images (Gaussian plates), 16D latent}}
            \end{subfigure}
            
\caption[Inspecting the independence of the computed vector ranks. ]{Inspecting the quality of the computed vector ranks. Here, we visualize the distributions of the non-diagonal elements $\Sigma_{ij}$ in the \emph{correlation matrix} $\Sigma$ of the estimated vector ranks $\hat{\vecu}$. The results suggest that in most scenarios except case (d), $\hat{\vecu}_i$ and $\hat{\vecu}_j$ are highly independent.    }
\label{fig:vector-rank-quality}
\end{figure*}

\noindent \textbf{Diagnostics on the quality of the estimated vector ranks}. As discussed in the methodology and theory sections, the effectiveness of the proposed VCE method hinges on learning accurate vector ranks. We assess the quality of the estimated ranks $\hat{\vecu}_X, \hat{\vecu}_Y$ from two perspectives:

\begin{itemize}[leftmargin=*]
    \item \emph{Element-wise uniformity}: Each univariate component $\hat{\vecu}_d$ is guaranteed to follow a perfectly uniform distribution in our method, as we employ element-wise empirical ranking when mapping the learned latent in the flow model to $\vecu$.
    \item \emph{Cross-element independence}: We further examine whether different dimensions, $\hat{\vecu}_i$ and $\hat{\vecu}_j$, are statistically independent. Figure~\ref{fig:vector-rank-quality} visualizes the diagnostic results. In most settings, $\hat{\vecu}_i$ and $\hat{\vecu}_j$ appear highly independent, with the exception of case (d).
\end{itemize}

In practice, in addition to the above visual diagnostic, one can also use the following test statistics $t(\Sigma)$ to quantify the quality of the learned vector ranks:
\[
    t(\Sigma) = \max\Big(|\mathbb{Q}_{5\%}(\Sigma_{ij})|, \enskip |\mathbb{Q}_{95\%}(\Sigma_{ij})|\Big)
\]
where $\mathbb{Q}_{a\%}(\Sigma_{ij})$ is the $a\%$ quantile of the non-diagonal elements in the correlation matrix $\Sigma$ of $\hat{\vecu}$. Intuitively, a small $t(\Sigma)$ will indicate that most non-diagonal elements $\Sigma_{ij}$ in the correlation matrix $\Sigma$ is close to zero, reflecting strong independence between $\hat{\vecu}_i$ and $\hat{\vecu}_j$ for different dimensions $i$ and $j$.

Interestingly, we find that even if the vector ranks are not perfectly learned (see e.g. case (d) in Figure \ref{fig:vector-rank-quality}), our estimator still yields a reasonable estimate. This may be due to that all univariate ranks 
$\hat{\vecu}_d$ are perfectly uniform, so even if $\hat{\vecu}_i$, $\hat{\vecu}_j$ occasionally exhibit weak dependence, the overall entropy $|H(\hat{\vecu})|$ remains low, leading to an acceptable bias in Proposition \ref{prop:lossy-compression} and a reasonable final estimate.

\end{document}